%% file: poc_slt.tex
\documentclass{article} %
\usepackage{times}

\RequirePackage{fancyhdr}
\RequirePackage{natbib}

\setcitestyle{authoryear,round,citesep={;},aysep={,},yysep={;}}

\usepackage{fancyhdr}
\usepackage{hyperref}
\usepackage{url}

\input{sections/latex_header}

\title{POC-SLT: Partial Object Completion with SDF Latent Transformers}

\author{Faezeh Zakeri, Raphael Braun, Lukas Ruppert, Henrik P.A. Lensch\\
Department of Computer Science\\
University of T\"ubingen\\
Tübingen, Baden-W\"urttemberg, 72076, Germany \\
\small \texttt{\{faezeh-sadat.zakeri, raphael.braun, lukas.ruppert, hendrik.lensch\}} \\
\small \texttt{@uni.tuebingen.de} \\
}

\begin{document}

\maketitle

\begin{abstract}
\input{sections/abstract}

\end{abstract}

\input{sections/introduction}

\input{sections/related_work}

\input{sections/method}

\input{sections/evaluation}

\input{sections/discussion}
\input{sections/conclusion}

\section*{Acknowledgements}
\input{sections/acks}

\bibliography{MyPaper}
\bibliographystyle{poc_slt}

\appendix
\input{sections/appendix}

\input{sections/appendix_metrics_and_preprocessing}
\input{sections/appendix_architecture_and_training}

\end{document}

%% file: sections/latex_header.tex
\usepackage{amsmath}
\usepackage{amsfonts}
\usepackage{amssymb}
\usepackage{booktabs}
\usepackage{multirow}

\usepackage{svg}
\usepackage{tikz}
\usetikzlibrary{arrows,shapes,positioning}
\usetikzlibrary{calc, decorations.markings}
\usetikzlibrary{patterns.meta}
\usetikzlibrary{3d}

\definecolor{conv3d_color}{HTML}{ffffff}
\definecolor{batchnorm_color}{HTML}{ffffff}
\definecolor{convblock_color}{HTML}{ff7f0e}
\definecolor{maxpool_color}{HTML}{ffffff}
\definecolor{upsample_color}{HTML}{ffffff}
\definecolor{encoderlayer_color}{HTML}{98df8a}
\definecolor{decoderlayer_color}{HTML}{d62728}
\definecolor{decoder_color}{HTML}{ff9896}
\definecolor{encoder_color}{HTML}{e377c2}
\definecolor{vae_color}{HTML}{c5b0d5}
\definecolor{up_color}{HTML}{8c564b}
\definecolor{final_color}{HTML}{c49c94}
\definecolor{linear_color}{HTML}{ffffff}
\definecolor{local_patch_color}{HTML}{f7b6d2}
\definecolor{sdf_transformer_color}{HTML}{1f77b4}
\definecolor{sdf_value}{HTML}{73f6ff}

%% file: sections/abstract.tex
3D geometric shape completion hinges on representation learning and a deep understanding of geometric data.
Without profound insights into the three-dimensional nature of the data, this task remains unattainable.
Our work addresses this challenge of 3D shape completion given partial observations
by proposing a transformer operating on the latent space representing Signed Distance Fields (SDFs).
Instead of a monolithic volume, the SDF of an object is partitioned into smaller high-resolution patches leading to a sequence of latent codes.
The approach relies on a smooth latent space encoding learned via a variational autoencoder (VAE), trained on millions of 3D patches.
We employ an efficient masked autoencoder transformer to complete partial sequences into comprehensive shapes in latent space.
Our approach is extensively evaluated on partial observations from ShapeNet and the ABC dataset where only fractions of the objects are given.
The proposed POC-SLT architecture compares favorably with several baseline state-of-the-art methods,
demonstrating a significant improvement in 3D shape completion, both qualitatively and quantitatively.

%% file: sections/introduction.tex
\section{Introduction}

The 3D measurements from scanners or sensors can be used to assemble virtual objects or entire worlds, either for visualizations, but also for reasoning about the world for example in self-driving cars.
In general, with optical sensors, it is only possible to measure 3D surfaces that are visible to the sensor from its current point of view.
3D scans are therefore always composed of many partial scans from different angles.
The more unique views are combined, the more complete the reconstructed object surface will become.
This creates a trade-off between speed and completeness.
Shape completion takes an incomplete shape and tries to estimate the complete geometry based on some learned shape prior. 

In this work, we present a new method for shape completion using a transformer in the style of Masked Auto Encoders (MAE) \citep{he2022masked} operating in latent SDF space. %
For an unknown or masked region, it is supposed to fill in a plausible shape that connects well with the given geometry. 
The method consumes and produces shapes represented as Signed Distance Fields (SDF) on volumetric tiles.
Similar to Stable Diffusion~\citep{rombach2022high}, the incomplete shapes are first encoded into a smooth, compressed latent space representation using our dedicated \emph{Patch Variational Autoencoder (P-VAE)}. For high-resolution models, the entire shape is represented as a sequence of latent codes generated by the P-VAE on volumetric patches. 
Completion is performed in latent space and the completed shape is decoded back to an SDF using the decoder of the P-VAE on each patch. 
Note that our \emph{SDF-Latent-Transformer} is a Masked Auto Encoder.
It therefore solves the task directly in a single inference step without a transformer decoder.

Shape completion requires a strong prior for real-world shapes in order to estimate the most likely shapes for incomplete inputs.
We trained our P-VAE and SDF-Latent-Transformer on ShapeNetCoreV1~\citep{ShapeNet} with all 55 categories.
We further fine-tuned the transformer on a subset of the ABC dataset \citep{ABC_Koch_2019_CVPR} with 100K meshes.

By thoroughly evaluating our methods on shape completion tasks,
we demonstrate superior quality compared to state of the art.
Our key contributions to the field are:
i) A comprehensive Patch Variational Autoencoder (P-VAE) to compress SDF shapes into sequences of codes in a smooth latent space.        
ii) A SDF-Latent-Transformer trained as Masked Auto Encoder completes input shapes within milliseconds in a single inference step.
iii) Upon acceptance, we will release code and model checkpoints and make the training data sets available.
For the P-VAE, it consists of 2.6 million $32^3$ patches and for the SDF-Latent-Transformer of $128^3$ SDFs for all objects in ShapeNetCoreV1 which were severely cleaned.

Ours is the first SDF-based method trained and evaluated
on all of ShapeNet (around 50k meshes) and a 100k subset of ABC.
This is an order of magnitude more than previous SDF-based approaches and 
at higher resolution, while using consumer-grade GPUs.
While MAEs are used in many domains by now, they usually are a means to pretrain a powerful encoder,
whose strong representation can be used for downstream tasks.
We however directly use the training objective as a shortcut to get instant completion.
This only works because of the robustness and strong geometric prior in our latent space.

%% file: sections/related_work.tex
\section{Related Work}
We review related point-based, occupancy-based, and SDF-based 3D completion methods %
and methods inspiring our deep-learning architecture. %

\paragraph{Point-based Shape Completion.}
3D Shape completion tackles the task of reconstructing 3D objects from partial observations in the form of 2D images,
sometimes with depth information, or 3D point clouds.
Areas occluded during capture need to be filled in to form a complete object.

PointNet and its variations \citep{PointNet_Qi_2017_CVPR, PointNet++_NIPS2017_d8bf84be} have pioneered the direct processing of 3D coordinates with neural networks, catalyzing research in numerous downstream tasks, including point cloud completion, as for example PCN~\citep{PCN_8491026}.
PCN uses PointNet in an encoder-decoder framework and integrates a FoldingNet~\citep{FoldingNet_Yang_2018_CVPR} to transpose 2D points onto a 3D surface by simulating the deformation of a 2D plane.
Following PCN, a plethora of other methods have emerged \citep{TopNet_8953650, PF-Net_Huang_2020_CVPR, GRNet_10.1007/978-3-030-58545-7_21, minghua_2019_morphing}, aiming to enhance point cloud completion with higher resolution and increased robustness.

PointTR~\citep{PoinTr_Yu_2021_ICCV} was one of the first approaches
which used transformers~\citep{vaswani2017attention} for point cloud completion.
They derive the point proxies from a fixed count of furthest point sampled representatives,
which are turned into a feature vector using a DGCNN~\citep{wang2019dynamic}.
Completion is done by a geometry-aware transformer, which is queried for missing point proxies.
The missing points are extracted from the predicted proxies using a
FoldingNet~\citep{FoldingNet_Yang_2018_CVPR}. The completed points are finally
merged with the input point cloud.
AdaPointTR~\citep{AdaPoinTr} addresses discontinuity issues and is more robust to noisy input.
SeedFormer~\citep{SeedFormer_10.1007/978-3-031-20062-5_24} completes a sparse set of patch seeds
from an incomplete input and upsamples the seeds to a complete point cloud.
The seeds are sparse 3D positions enriched by a transformer with semantic information about the local neighborhood.
SnowFlakeNet~\citep{SnowflakeNet_Xiang_2021_ICCV} solves this decoding task with a SkipTransformer-based architecture, which repeatedly splits and refines low-resolution points.
LAKe-Net~\citep{LAKe_Net_Tang_2022_CVPR} first localizes aligned keypoints to generate a surface skeleton mesh which aids in producing a complete point cloud.
AnchorFormer~\citep{Chen2023AnchorFormer} extracts global features
and predicts key "anchor" points which are combined with a subset of input points and then upsampled into a dense mesh.
VRPCN~\citep{VRPCN_Pan_2021_CVPR} models the shape from the partial input as
probability distributions, which are sampled and then refined by a hierarchical
encoder-decoder network. This is related to our approach as we
also use a Variational Auto Encoder (VAE) to compress the shape in latent space.
LION~\citep{LION_NEURIPS2022_40e56dab} introduced a flexible latent diffusion model for point clouds also using a VAE to map point cloud features into latent space. The features and diffusion model are constructed from Point-Voxel CNNs.

\paragraph{Occupancy-based Shape Completion.}
Occupancy Networks~\citep{Mescheder2019OccupancyNets} introduced a functional, implicit representation of 3D shapes via their occupancy.
These methods can be queried for arbitrary 3D locations and will return whether the queried position is inside or outside of the represented object.
Meshes can quickly be recovered from this representation using the MISE algorithm~\citep{Mescheder2019OccupancyNets}.
\citet{Peng2020ConvolutionalOccNets} extend this to a convolutional method and represent entire scenes by storing and interpolating latent representations in 2D/3D grids.
\citet{Chibane2020IFNets} combine local and global information using multi-scale feature grids in IF-Nets.

These methods focus on identifying and representing 3D objects seen in images
and adding detail to uniformly low-resolution point cloud and voxel input.

ShapeFormer~\citep{ShapeFormer_Yan_2022_CVPR} compactly encodes shapes into a sparse grid of vector-quantized deep implicit functions (VQDIF).
Partial inputs are completed by an autoregessive transformer operating on a sequence of location and patch index pairs.
3DILG~\citep{Zhang2022_3DILG} uses an irregular grid of latent codes for shape representation and completes partial inputs with an autoregressive transformer.
3DShape2VecSet~\citep{Zhang2023_3DShape2VecSet} derives multiple global vectors for shape representation using cross-attention between input points and sampled query points.
The (variationally) auto-encoded shape is then completed using latent-space diffusion.
The output occupancy is queried by cross-attention between a dense point cloud and the global vectors.

The input to these methods is typically a point cloud generated from a single camera view.
Occluded parts need to be inferred from the partial point cloud.
In many cases, the input still covers large parts of the object,
exposing global information about the shape,
while our method completes objects from a small subset of complete patches.

\paragraph{SDF-based Shape Completion.}
Signed Distance Fields (SDFs) represent 3D shapes implicitly as continuous 3D scalar functions, which describe the distance of any 3D point to the closest surface.
The surface is given by the zero-level-set while in- or outside points are identified by the sign.
This ensures that objects represented by SDFs are always watertight.
In contrast, meshes are often non-manifold since they are commonly created to "look right" on screen,
rather than to represent the volume of solid geometry.
While SDFs could be modeled in any functional basis, we use regular 3D grids.

There are only few approaches related to shape completion on SDFs in the literature.
For SDF representation and class-related shape generation, DeepSDF~\citep{DeepSDF_Park_2019_CVPR} proposes to
use an
AutoDecoder~\citep{AutoDecoder_osti_80034} to jointly optimize a compressed
latent representation of an SDF and the decoder to extract the full SDF from the latent representations.
Due to the fully connected layer, the approach has limited resolution. Our approach instead splits the SDF into patches, which are
encoded with a rigorously trained VAE~\citep{kingma2013auto}.
Those patches can be assembled to very highly detailed shapes.
We experimented with using a similar AutoDecoder optimization scheme to refine latent codes for our frozen decoder
but found that the optimized latent codes complicate subsequent shape completion tasks (Appendix~\ref{sec:appendix:slt_ad}).

AutoSDF and SDFusion~\citep{AutoSDF_Mittal_2022_CVPR, SDFusion_Cheng_2023_CVPR} focus on multi-modal shape generation and completion in latent space, e.g.\ from an input image, point clouds, or depth maps.
Both approaches use Vector Quantized Variational Auto Encoders (VQ-VAEs)~\citep{VQ_NIPS2017_7a98af17,
HVQ_NEURIPS2019_5f8e2fa1} to patchwise compress SDF shapes to a compact latent space. SDFusion then uses a latent diffusion model, while AutoSDF employs an autoregressive transformer to generate or complete shapes sequentially.
AutoSDF~\citep{AutoSDF_Mittal_2022_CVPR} is in many regards similar to our
approach. The main differences are: i) the choice of a VAE over a VQ-VAE to avoid
quantization errors.
ii) Larger patches with $32^3$ grid points over the $8^3$
grid points used by AutoSDF.
iii) Our transformer is trained as Masked Auto Encoder
(MAE)~\citep{he2022masked} while the AutoSDF Transformer is trained
autoregressively. This drastically changes how the model can be used as our approach does not require a sequential generation.
iv) Our approach is trained on SDF grids with a resolution of $128^3$ and trained on the full ShapeNetCoreV1~\citep{ShapeNet} and refined on the ABC datasets~\citep{ABC_Koch_2019_CVPR} instead of the SDF resolution of $64^3$ on a 13-category subset of SapeNetCoreV1 for AutoSDF.

PatchComplete~\citep{Rao2022PatchComplete} and DiffComplete~\citep{Chu2023DiffComplete} both complete $32^3$ Truncated SDF input using multi-scale features.
While PatchComplete uses an attention mechanism, DiffComplete uses diffusion
and also allows combination of multiple inputs.
Both methods learn strong priors that generalize well to new inputs, albeit at comparably low resolution.
While our method works on a subset of clean SDF patches to extend partial geometry,
these methods work on the 3D-EPN~\citep{TR1_Dai_2017_CVPR} dataset, containing noisy, low-resolution partial SDFs generated from a single camera view.

\paragraph{Variational Autoencoders (VAE)} \citep{kingma2013auto} are networks that learn latent representations of input data.
Unlike standard autoencoders, they learn a probability distribution over the latent space.
This generates a smooth latent space which allows manipulation and computations in latent space.
Inspired by Stable Diffusion~\citep{rombach2022high}, we train a VAE to encode the $32^3$ SDF patches into a smooth latent space.
We call this network Patch Variational Autoencoder (P-VAE).
A full $128^3$ SDF grid is therefore represented by only 64 latent tokens. 
This makes any further processing very efficient and scalable.

\paragraph{Transformers}~\citep{vaswani2017attention} are a network architecture for sequence-to-sequence tasks.
It uses attention as the core mechanism to identify pairwise relationships between elements in a sequence.
They have revolutionized Natural Language Processing and are the foundation of all current Large Language Models.
Transformers can however process any sequence of data.
In \citet{dosovitskiy2020image}, images are split into a sequence of patches, which is then processed by transformers. The approach in
\citet{he2022masked} masks out some of those patches and train to fill in the masked-out patches.
Similarly, we split the SDF grid into patches and encode them into a sequence of latent codes.
We mask out parts of the latent codes that contain unknown geometry and use our SDF-Latent-Transformer to complete the missing patches.

%% file: sections/method.tex
\begin{figure*}[htb]
    \centering
    \input{figures/overview_arch}
    \vspace{-4mm}
    \caption{Architecture Overview. The SDF is partitioned into tiles of $32^3$
    samples. For each tile, a latent code is generated by a variational
    autoencoder (P-VAE) resulting in the potentially partial input stream to our
    SDF-Latent-Transformer. It is trained as a Masked Autoencoder
    to generate a completed series of tokens which are finally translated back
    to SDF tiles using the P-VAE decoder.} 
    \label{fig:overview}
\end{figure*}
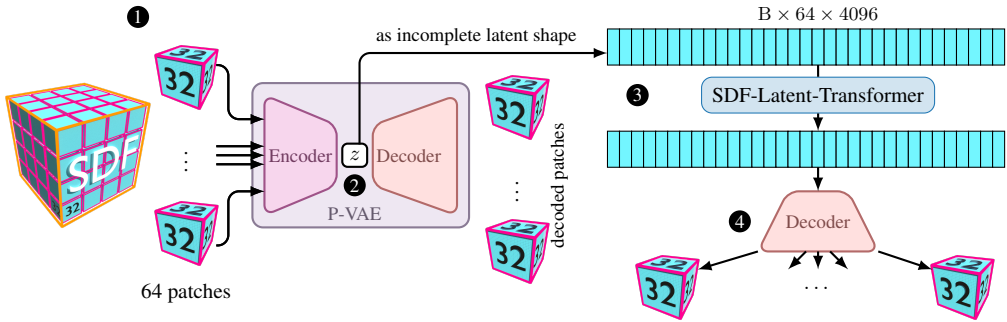
\section{Method}
Our POC-SLT shape completion pipeline consists of four main steps which are depicted in Figure~\ref{fig:overview}:
First, the input SDF grid is split into patches of $32^3$ \tikz{\node[circle, fill=black, text=white, inner sep=1pt]{\tiny 1}}.
Each of those patches is encoded into a latent vector using the P-VAE \tikz{\node[circle, fill=black, text=white, inner sep=1pt]{\tiny 2}}.
The latent codes $z$ that form the input shape are then assembled in a sequence,
which is masked and processed by the SDF-Latent-Transformer (SLT) \tikz{\node[circle, fill=black, text=white, inner sep=1pt]{\tiny 3}}.
Finally, the resulting sequence is decoded patch-by-patch with the decoder from the P-VAE into a completed SDF \tikz{\node[circle, fill=black, text=white, inner sep=1pt]{\tiny 4}}.
We will now elaborate on these four steps.

\subsection{P-VAE} \label{sec:pvae}
The task for the P-VAE is to encode $32^3$-SDF grid patches $(p_i)_{i=0,\dots,n}$ into latent code vectors $(z_i)_{i=0,...,n}$.
For details about the architecture and implementation of our P-VAE, refer to Appendix~\ref{appendix:architecture}.
The P-VAE consists of an encoder $E_\mathrm{VAE}$ and a decoder $D_\mathrm{VAE}$, which are both based on 3D-convolutional layers.
A patch $p_{i}$ is encoded by the encoder into mean $\mu_i$ and variance $\sigma^2_i$ of a Gaussian distribution.
During training, this distribution is sampled to obtain the latent code $z_i \sim \mathcal{N}(\mu_i, \sigma_i^2)$,
which is then decoded by the Decoder $D_\mathrm{VAE}$ into an SDF patch $\tilde{p}_i = D_\mathrm{VAE}(z_i)$.
We use the mean absolute error between $p_i$ and $\tilde{p}_i$ as reconstruction loss. 
To regularize the latent space of the VAE, we follow the \emph{KL-regularized} VAE implementation from Stable Diffusion~\citep{rombach2022high}.
During inference, we directly use the means $z_i=\mu_i$ without additional sampling.

The P-VAE is trained on SDF patches extracted randomly from parts of the meshes of ShapeNetCoreV1~\citep{ShapeNet}.
Each patch has a fixed SDF resolution of $32^3$.
We however strongly vary the side length of each extracted patch while ensuring that they remain close to a surface. %
Specifically, we uniformly sample surface points and side lengths $d \sim \mathcal{U}_{[0.1,1]}$ as well as offsets $x,y,z \sim \mathcal{N}(0, d/3)$.
The P-VAE is therefore exposed to a large variety of detail levels, scales, and surface types and has to learn the full variety of how patches of natural shapes can potentially look like.
This yields a strong, generalizable shape prior which allows the P-VAE to generalize even to patches of shapes from completely different datasets.
 \begin{figure}[tb]
     \centering
     \raisebox{4em}{\input{figures/auto_decoder}}
     \input{figures/transformer}
     \vspace{-5mm}
     \caption{
     \textbf{P-VAE (left)}: On millions of patches, we train a smooth embedding space for SDFs
     with a variational autoencoder (P-VAE) that samples the noise according to
     the predicted variance before passing the estimated latent to the decoder. %
     \textbf{SDF-Latent-Transformer (SLT) (right)}: Performs shape completion on input
     sequences consisting of SDF-Patches (\textcolor{sdf_value!70!black}{cyan}) in latent space.
     During training, some of the input patches
     are masked and substituted with a trainable shared vector (\textcolor{blue}{blue}).
     The utilized masking schemes \emph{Random}, \emph{Half}, \emph{Octant} and
     \emph{Slice} are visualized on the right. 3D positional encoding is added
     before a TransformerEncoder propagates the information from the remaining
     patches to all the masked patches completing the 3D shape. 
     }
     \label{fig:autodecoder}
     \label{fig:transformer}
 \end{figure}
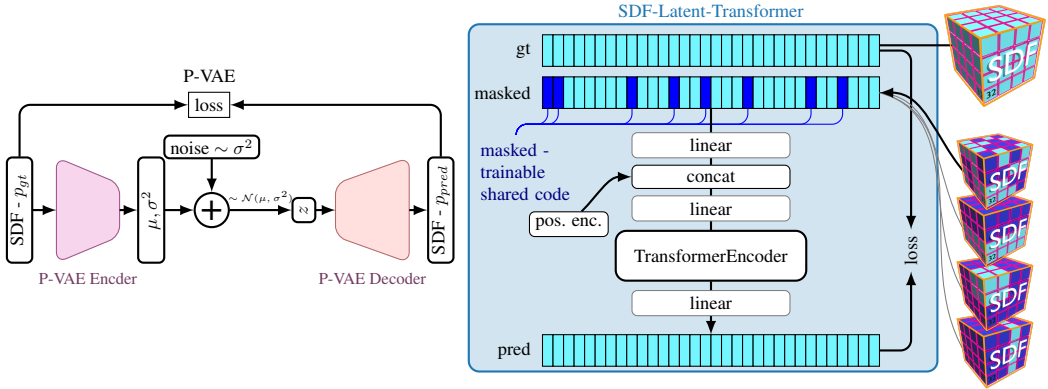
 
\subsection{SDF-Latent-Transformer} \label{sec:slt}
The actual shape completion happens completely in latent space.
Our approach is visualized on the right in Figure~\ref{fig:transformer}.
The incomplete input shape is provided as a sequence of latent space vectors. 
We employ a standard transformer encoder as a Masked Auto Encoder to fill in missing latent vectors.
The missing input parts are marked by a special trainable \emph{mask token} (\textcolor{blue}{blue} in the figure).
The positions of all vectors, including masked ones, are provided to the transformer
as 3D positional encoding~\citep{Siren_NEURIPS2020_53c04118}. 
Linear layers before and after concatenating the positional embedding have
empirically shown to improve convergence. Finally, the transformer completes the
sequence.

\paragraph{Outside SDF Patches.}
For many, especially anisotropic shapes, some of the encoded SDF-Patches will not contain any geometry either because they are completely inside or outside an object.
Note however, that those patches still contain valid distance values towards surfaces that lie outside
those patches. We therefore do not differentiate between patches that explicitly contain surfaces and those that lie outside.

\paragraph{Masking during Training.}\label{para:method:masking}
The training strategy of the SDF-Latent-Transformer is visualized in Figure~\ref{fig:transformer}.
For creating the mask for a given training example, we randomly choose from one of the following masking strategies:
1. \emph{Random Masking}, 2. \emph{Octant}, 3. \emph{Half}, 4. \emph{Slice}.
With \emph{Random Masking}, each patch is masked out with a $40\%$ probability. 
With \emph{Octant}, all but one of the octants of the patch-grid are masked out.
With \emph{Half}, one half of the patch-grid is masked out.
With {Slice}, everything but one axis-aligned slice through the patch-grid is masked out.
The four different masking schemes are visualized in the same order from top to bottom on the right side of Figure~\ref{fig:transformer}.
We use \emph{Random Masking} with a probability of $35\%$ and all the other strategies with a probability of $21.7\%$ each.

\paragraph{Loss Functions.}
We supervise our training with ground truth patch encodings $z = E_\mathrm{VAE}$ for the complete object. 
The entire training of the SDF-Latent-Transformer happens in latent space.
The full loss function $\mathcal{L}_\mathrm{comp}$ consists of two terms:
\begin{equation}
  \mathcal{L}_{\text{comp}} = \alpha\mathcal{L}_{\text{masked}} + \beta\mathcal{L}_{\text{non-masked}}
\end{equation}
Both $\mathcal{L}_\mathrm{masked}$ and $\mathcal{L}_\mathrm{non-masked}$ are simple $L1$-losses on the latent codes
evaluated on results with masked and non-masked inputs respectively.
The weights $\alpha$ and $\beta$ are chosen, such that the contribution of 
every patch to the total loss is equal, no matter how many patches were masked.
With $N$ patches in total from which $M$ are masked out we get:
\begin{align}
    \alpha &= \frac{M}{N} &
    \beta &= \frac{N-M}{N}
\end{align}

\paragraph{Shape Completion.}
For shape completion, one can use the SDF-Latent-Transformer in the same mode as during training. 
We assume the partial shape is given in the form of a high-resolution SDF volume. 
For each given patch, a latent code is generated with the P-VAE encoder. 
The known SDF patches (including known empty patches)
are handed over as input to the transformer while marking all others as masked. 
The transformer will then predict proper latent codes for all masked tokens, which includes predicting latent codes for empty patches.

The resulting grid of latent codes can then be decoded back into an $128^3$ SDF
which can be converted into a mesh, e.g.\ using Marching Cubes~\citep{marching_cube_article}.

%% file: figures/overview_arch.tex
\begin{tikzpicture}[scale=0.8, transform shape]
    \node at (0,0) {\includegraphics[width=2.5cm]{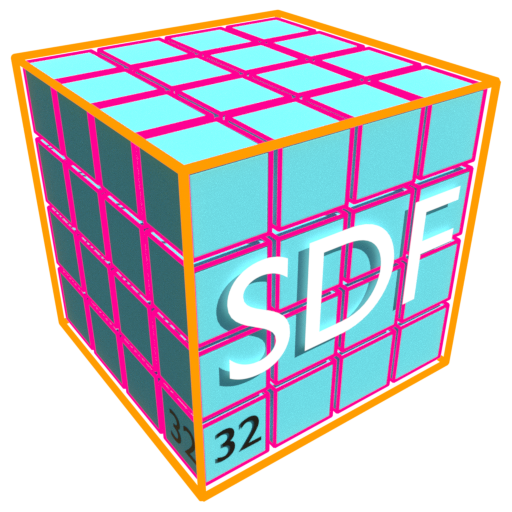}};
    \node at (1.8,1.3) {\includegraphics[width=1.1cm]{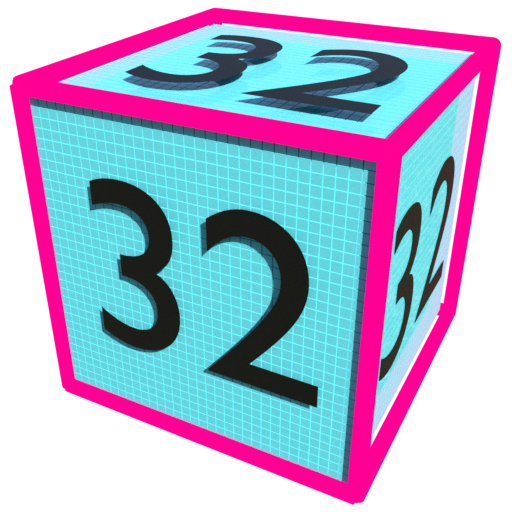}};
    \node at (1.8, 0) {\vdots};
    \node at (1.8,-1.3 ) {\includegraphics[width=1.1cm]{figures/patch.png}};
    \node at (1.8, -2.3) {64 patches};
    \draw[thick, vae_color!70!black, fill=vae_color!30, rounded corners] (2.9, -1.2) rectangle (6.50, 1.2);
    \node[vae_color!50!black, anchor=south] at (4.6, -1.2) {\small P-VAE};
    \draw[thick, encoder_color!70!black, fill=encoder_color!30, rounded corners, xshift=5cm] 
    (-0.7, 0.5) -- (-1.9, 1) -- (-1.9, -1) -- (-0.7, -0.5) -- cycle;
    
    \node[thick, encoder_color!50!black] at (3.7, 0) {\small Encoder};
    \draw[thick, -latex, rounded corners] (2.3, 1.5) -- (2.5, 1.5) |- (3.1, 0.6); 
    \draw[thick, -latex, rounded corners] (2.3, 0.15) -- (3.1, 0.15); 
    \draw[thick, -latex, rounded corners] (2.3, 0.0) -- (3.1, 0.0); 
    \draw[thick, -latex, rounded corners] (2.3, -0.15) -- (3.1, -0.15); 
    \draw[thick, -latex, rounded corners] (2.3, -1.5) -- (2.5, -1.5) |- (3.1, -0.6); 
    \draw[thick, decoder_color!70!black, fill=decoder_color!30, rounded corners, xshift=2cm] (2.9, -0.5) -- (2.9, 0.5) -- (4.3, 1) -- (4.3, -1) -- cycle;
    \node[thick, decoder_color!50!black] at (5.5, 0) {\small Decoder};
    \node[thick, draw, rounded corners=2pt, fill=white, minimum height = 0.29] (z) at (4.6, 0) {$z$};
    \node at (7.3,0.8) {\includegraphics[width=1.1cm]{figures/patch.png}};
    \node at (7.3, -0.5) {\vdots};
    \node at (7.3,-1.6 ) {\includegraphics[width=1.1cm]{figures/patch.png}};
    \node[rotate=90] at (8.0, -0.5) {\small decoded patches};
    \draw[thick, -latex, rounded corners] (2.3, 1.5)   -- (2.5, 1.5) |- (3.1, 0.6); 
    \draw[thick, -latex, rounded corners] (2.3, 0.15)  -- (3.1, 0.15); 
    \draw[thick, -latex, rounded corners] (2.3, 0.0)   -- (3.1, 0.0); 
    \draw[thick, -latex, rounded corners] (2.3, -0.15) -- (3.1, -0.15); 
    \draw[thick, -latex, rounded corners] (2.3, -1.5)  -- (2.5, -1.5) |- (3.1, -0.6); 

    \node[anchor=south] at (12.3, 2.1) {\small $\mathrm{B} \times 64 \times 4096$};
    \foreach \i in {0,...,32}{
        \draw[fill=sdf_value] (8.8+\i * 0.2, 1.5) rectangle (9.0+\i*0.2, 2.1);
    }
    \draw[thick, -latex, rounded corners] (4.68, 0.18) to (4.68, 1.7) to node[anchor=south] {\small \hspace*{-1em} as incomplete latent shape} (8.8, 1.7);

    \node[draw, rounded corners, inner sep = 5pt, sdf_transformer_color, fill=sdf_transformer_color!20, text=black] (transformer) at (12.3, 1.0) {SDF-Latent-Transformer};
    \foreach \i in {0,...,32}{
        \draw[fill=sdf_value] (8.8+\i * 0.2, 0.4) rectangle (9.0+\i*0.2, -0.2);
    }
    \draw[thick, decoder_color!70!black, fill=decoder_color!30, rounded corners] (12.3, -0.6 ) -- (12.8, -0.6) -- (13.3, -1.6) -- (11.3, -1.6) -- (11.8, -0.6) -- cycle;
    \node[thick, decoder_color!50!black] at (12.3, -1.1) {\small Decoder};
    
    \node at (9.8, -2.2) {\includegraphics[width=1.1cm]{figures/patch.png}};
    \node at (12.3, -2.2) {$\dots$};
    \node at (14.7, -2.2) {\includegraphics[width=1.1cm]{figures/patch.png}};

    \draw[-latex, thick] (12.3, 1.5) to (transformer) to (12.3, 0.4);
    \draw[-latex, thick] (12.3, -0.2) to (12.3, -0.6);
    
    \draw[-latex, thick] (11.3, -1.6) to (10.3, -1.9);
    \draw[-latex, thick] (12.1, -1.6) to (11.8, -1.9);
    \draw[-latex, thick] (12.3, -1.6) to (12.3, -1.9);
    \draw[-latex, thick] (12.5, -1.6) to (12.8, -1.9);
    \draw[-latex, thick] (13.3 , -1.6) to (14.2, -1.9);

    \node[circle, fill=black, text=white, inner sep = 1pt] at (1, 2.3) {\small{1}};
    \node[circle, fill=black, text=white, inner sep = 1pt] at (4.6, -0.5) {\small{2}};
    \node[circle, fill=black, text=white, inner sep = 1pt] at (9.3, 1.0) {\small{3}};
    \node[circle, fill=black, text=white, inner sep = 1pt] at (11, -1.1) {\small{4}};
\end{tikzpicture}

%% file: figures/auto_decoder.tex
\begin{tikzpicture}[scale=0.7, transform shape]
    \node[anchor=south] at (1.01,  2.3) {P-VAE};

    \node[draw, thick, fill=white, rounded corners = 2pt, inner sep=2pt, text=black, minimum width=2cm, minimum height = 3mm, rotate=90] (sdfin) at (-2.6,  0) {SDF - $p_{gt}$};
    \draw[encoder_color!70!black, fill=encoder_color!30, rounded corners] (-1.9, 1) -- (-0.7, 0.5) -- (-0.7, -0.5) -- (-1.9, -1) -- cycle;
    \node[encoder_color!50!black] at (-1.3, -1.3) {\small P-VAE Encder};
    
    \node[draw, thick, fill=white, rounded corners = 2pt, inner sep=2pt, text=black, minimum width=2cm, minimum height = 2.5, rotate=90] (z) at (-0.10,  0) {$\mu, \sigma^2$};
    
    \node[draw, thick, fill=white, rounded corners = 2pt, inner sep=2pt, text=black, minimum width=1.8cm, minimum height = 3mm] (noise) at (1.05,  1.2) {noise $\sim \sigma^2$};
    \node[draw, thick, circle, inner sep=1pt, fill=white, minimum width = 3mm] (add1) at (1.05, 0) {{\huge +}};
    \node[draw, thick, fill=white, rounded corners = 2pt, inner sep=4pt, text=black, minimum height = 3mm, rotate=90] (z_sampled) at (2.8,  0.0) {$z$};

    \draw[decoder_color!70!black, fill=decoder_color!30, rounded corners] (3.4, -0.5) -- (3.4, 0.5) -- (4.8, 1) -- (4.8, -1) -- cycle;
    \node[encoder_color!50!black] at (4.1, -1.3) {\small P-VAE Decoder};
    
    \node[draw, thick, fill=white, rounded corners = 2pt, inner sep=2pt, text=black, minimum width=2cm, minimum height = 3mm, rotate=90] (sdfout) at (5.4,  0) {SDF - $p_{pred}$};

    \node[draw] (loss_vae) at (1.0,2.0) {loss};

    \draw[thick, -latex] (sdfin) to (-1.9,0);
    \draw[thick, -latex] (-0.7,0) to (z);
    \draw[thick, -latex] (z) to (add1);
    \draw[thick, -latex] (noise) to (add1);
    \draw[thick, -latex] (add1) to  node[anchor=south] {\tiny{$\sim \mathcal{N}(\mu, \sigma^2)$}}(z_sampled);
    \draw[thick, -latex] (z_sampled) to (3.4, 0);
    \draw[thick, -latex] (4.8, 0) to (sdfout);
    \draw[thick, -latex, rounded corners] (sdfin) |- (loss_vae);
    \draw[thick, -latex, rounded corners] (sdfout) |- (loss_vae);
    
\end{tikzpicture}

%% file: figures/transformer.tex
\begin{tikzpicture}[scale=0.7, transform shape]
    \draw[thick, sdf_transformer_color, rounded corners, fill=sdf_transformer_color!20] (-1.4, 1.3) rectangle (7.5, -5.3);
    \node[sdf_transformer_color, anchor=south] at (3.2, 1.3) {SDF-Latent-Transformer};
    \foreach \i in {0,...,31}{
        \draw[fill=sdf_value] (\i * 0.2, 0.5) rectangle (0.2+\i*0.2, 1.1);
    }
    \node[anchor=east] at (-0.1, 0.8) {gt};
    \node[anchor=west, text width=1.8cm, align=left, blue!50!black](masked) at (-1.3, -1.5) {masked - trainable shared code};
    \foreach \i in {0,1,8,12,15,19,25,28}{
        \draw[fill=blue] (\i * 0.2, -0.3) rectangle (0.2+\i*0.2, 0.3);
        \draw[blue, rounded corners] (-0.5, -0.8) to (-0.5, -0.6) -| (0.1+\i*0.2, -0.3);
    }
    
    \foreach \i in {2,3,4,5,6,7,9,10,11,13,14,16,17,18,20,21,22,23,24,26,27,29,30,31}{
        \draw[fill=sdf_value] (\i * 0.2, -0.3) rectangle (0.2+\i*0.2, 0.3);
    }
    \node[anchor=east] at (-0.1, 0.0) {masked};
    \node[draw, rounded corners=2pt, linear_color!50!black, fill=linear_color!50, text=black, minimum width=3cm] (redundant) at (3.2, -1.0) {linear};
    \node[draw, rounded corners=2pt, fill=white, minimum width=3cm] (cat) at (3.2, -1.6) {concat};
    \node[draw, rounded corners=2pt, linear_color!50!black, fill=linear_color!50, text=black, minimum width=3cm] (map) at (3.2, -2.2) {linear};
    \node[draw, rounded corners, thick, inner sep=10pt, fill=white] (transformer) at (3.2, -3.1) {TransformerEncoder};
    \node[draw, rounded corners=2pt, fill=white] (pos) at (0.5, -2.5) {pos. enc.};
    \node[draw, rounded corners=2pt, linear_color!50!black, fill=linear_color!50, text=black, minimum width=3cm] (out) at (3.2, -4.0) {linear};
    
    \foreach \i in {0,...,31}{
        \draw[fill=sdf_value] (\i * 0.2, -5.2) rectangle (0.2+\i*0.2, -4.6);
    }
    \node[anchor=east] at (-0.1, -4.9) {pred};
    
    \node[rotate=90] (loss) at (7.0, -3.0) {loss};

    \draw[thick, -latex] (pos) to[in=180] (cat);
    \draw[thick, -latex] (3.2, -0.3) to (redundant) to (cat) to (map) to (transformer) to (out) to (3.2, -4.6);

    \draw[thick, -latex, rounded corners] (6.4, 0.8) -| (loss);
    \draw[thick, -latex, rounded corners] (6.4, -4.9) -| (loss);

    \draw[thick] (6.4, 0.9) to (8, 0.9);
    \draw[gray] (6.4, 0.0) to[out=-5, in=120] (8, -2.6);
    \draw[gray] (6.4, 0.0) to[out=-15, in=120] (8, -3.8);
    \draw[gray] (6.4, 0.0) to[out=-15, in=120] (8, -4.9);
    \draw[thick, latex-] (6.4, 0.0) to[out=0, in=130] (8, -1.5);
    \node at (8.6, 0.7) {\includegraphics[width=2.0cm]{figures/sdf_nonmasked.png}};
    \node at (8.6, -4.9) {\includegraphics[width=1.5cm]{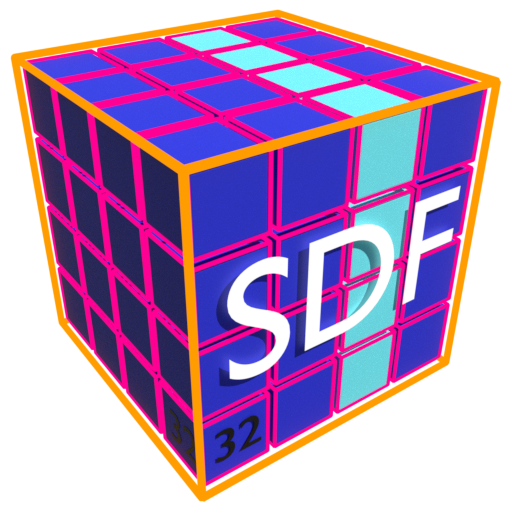}};
    \node at (8.6, -3.8) {\includegraphics[width=1.5cm]{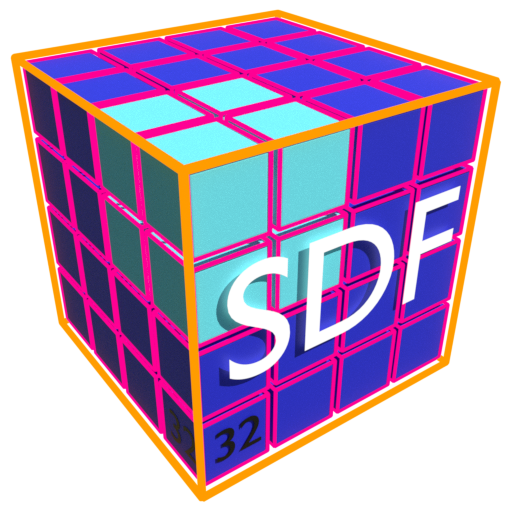}};
    \node at (8.6, -2.6) {\includegraphics[width=1.5cm]{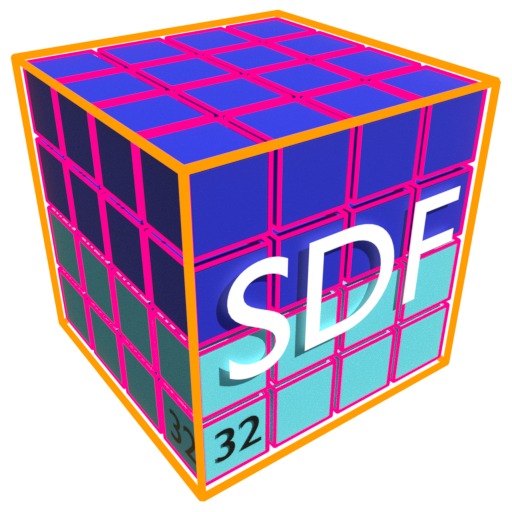}};
    \node at (8.6, -1.5) {\includegraphics[width=1.5cm]{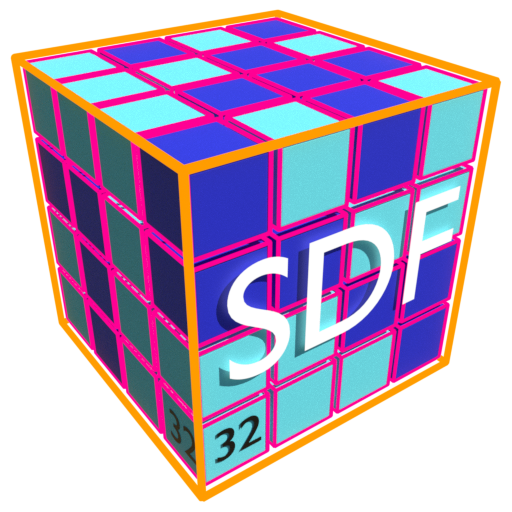}};
\end{tikzpicture}

%% file: sections/evaluation.tex
\section{Evaluation} \label{sec:evaluation}
POC-SLT is an efficient and fast solution for patch-wise SDF
completion in latent space. It consists of the P-VAE which encodes SDF patches
into latent space and the SDF-Latent-Transformer which fills in missing patches in latent space.
To demonstrate the effectiveness of our method, we measure its performance on various completion
tasks and compare it to related work below. 
Additional tasks are demonstrated in Appendix~\ref{sec:appendix:additional_eval}.
All metrics used in the experiments are defined in Appendix~\ref{appendix:metrics}.
Data preparation for turning meshes into SDFs is described in Appendix~\ref{appendix:data_prep}.

\paragraph{Completion:}
First, we provide quantitative and qualitative results for three SDF shape
completion tasks on the full ShapeNetCoreV1~\citep{ShapeNet}
dataset and ABC~\citep{ABC_Koch_2019_CVPR} test sets.
The tasks are to complete an SDF based on different types of partial inputs.
1. Only the bottom half of the SDF is given as input (Half).
2. Only the bottom right octant is given as input (Oct).
3. Patches are removed randomly
with 25\%, 50\%, and 75\% of the SDF remaining (R25, R50, R75).

A simpler version of those tasks on a subset of ShapeNet has been suggested
by~\citet{MPC} and used by~\citep{MPC, PoinTr_Yu_2021_ICCV,
AutoSDF_Mittal_2022_CVPR, SDFusion_Cheng_2023_CVPR} for completion comparisons.
Here, only 13 categories of
ShapeNet are used for training and evaluation and the Unidirectional Haushorff
Distance (UHD) is reported. Note that this only measures how faithfully the
geometry given in the input was preserved during the completion.

In addition, we also compare to AnchorFormer~\citep{Chen2023AnchorFormer} on the (Half) task, by 
using point cloud inputs with only points in the bottom half of the bounding box (Half).

\paragraph{P-VAE:}
The P-VAE is evaluated by measuring the deviation between
the input of the encoder and the output of the decoder.
We furthermore compare the reconstructions
from our P-VAE latent space to the reconstructions of an expensively optimized latent space
using~\citet{DeepSDF_Park_2019_CVPR, AutoDecoder_osti_80034}.

\paragraph{Timing:}
Our latent space completion consists of a single forward step of the SLT.
This makes the shape completion inference very fast. We demonstrate this
advantage over previous latent shape completion
works~\citep{AutoSDF_Mittal_2022_CVPR, ShapeFormer_Yan_2022_CVPR} in
Table~\ref{tab:runtime}.

\begin{table}[htb]
    \caption{Comparing shape completion inference time.}
    \label{tab:runtime}
    \centering~
    \begin{tabular}{lcc}
        \toprule
        Method & Inference Time & Output Resolution \\
        \midrule
        POC-SLT (Ours) & 8.6 milliseconds & $128^3$ \\
        AutoSDF~\citep{AutoSDF_Mittal_2022_CVPR} & 4.3 seconds    & $64^3$ \\
        ShapeFormer~\citep{ShapeFormer_Yan_2022_CVPR} & 20 seconds & $128^3$ \\
        \bottomrule
    \end{tabular}~
\end{table}

\paragraph{Hardware:}
The P-VAE and SDF-Latent-Transformer variations were trained on a machine with 3 Nvidia RTX 4090 GPUs and 512GB of main memory.
The P-VAE was trained for approximately two weeks, while the transformers were trained for about two days each.
Additional computation was required for preprocessing (Appendix~\ref{appendix:data_prep}) and evaluation,
which was carried out over several machines in our cluster, using mostly Nvidia RTX 2080 Ti GPUs and several CPUs.

\subsection{Completion}
The SDF-Latent-Transformer is the core component of our completion pipeline. 
It receives an incomplete sequence of SDF patches in latent space and completes %
it to a full sequence of latent codes in a single forward step.
The latent codes of the completed sequence are independently decoded into SDF patches with the P-VAE decoder. 
The patches are trivially assembled into a high-resolution shape volume from
which we extract surface meshes using Marching Cubes~\citep{marching_cube_article}.
We trained two slightly different variants of the SDF-Latent-Transformer:
{\bf SLT:} trained on ShapeNetCoreV1
with latent codes $z$ from P-VAE as ground truth
and {\bf SLT ABC:} SLT fine-tuned on a subset of the ABC dataset.
The quality of our shape completion approach is evaluated on unseen objects from
ShapeNetCoreV1~\citep{ShapeNet} and ABC~\citep{ABC_Koch_2019_CVPR}.

\begin{figure}[p]
    \centering
    \vspace*{-1em}
    \begin{tabular}{c@{}c@{}c@{}c@{}c@{}c@{}c@{}c}
         \rotatebox{90}{\parbox{0.11\linewidth}{\centering Input}}
         & \includegraphics[width=0.11\linewidth]{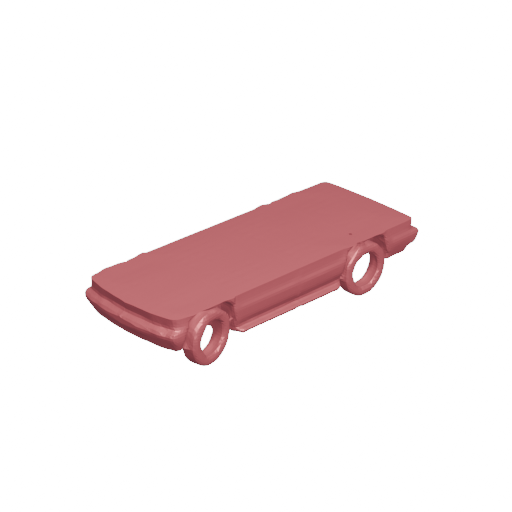}
         & \includegraphics[width=0.11\linewidth]{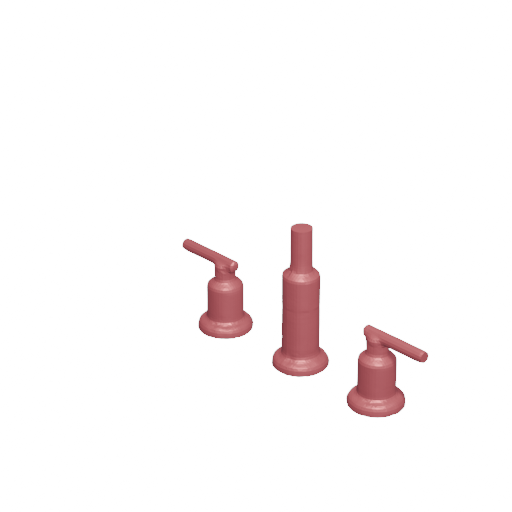}
         & \includegraphics[width=0.11\linewidth]{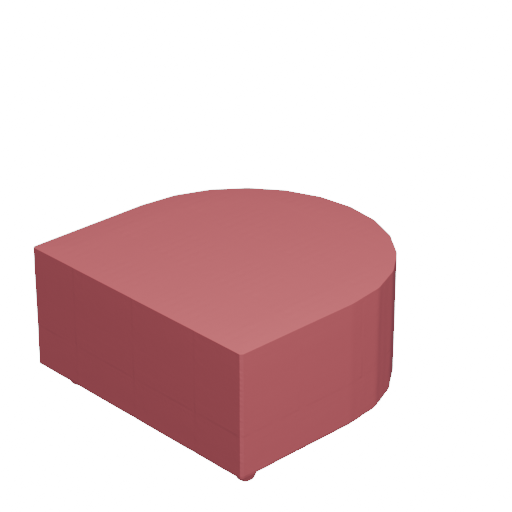}
         & \includegraphics[width=0.11\linewidth]{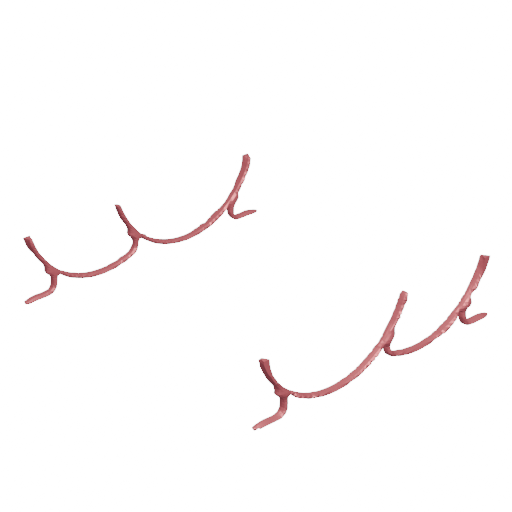}
         & \includegraphics[width=0.11\linewidth]{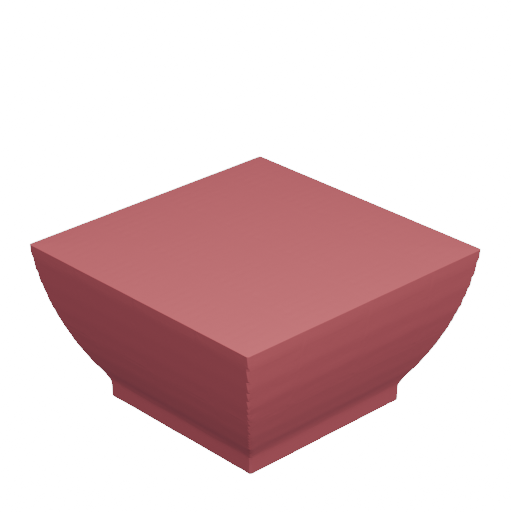}
         & \includegraphics[width=0.11\linewidth]{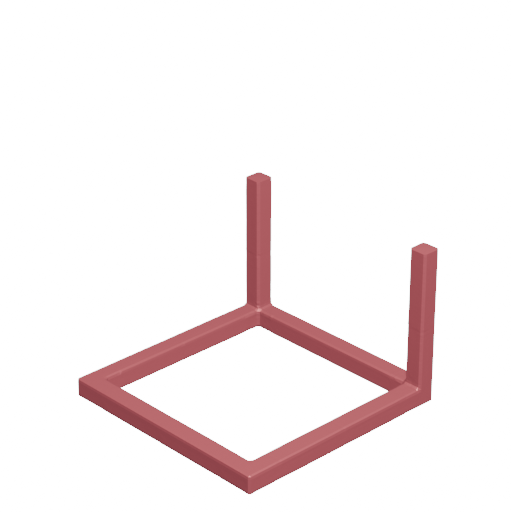}
         & \includegraphics[width=0.11\linewidth]{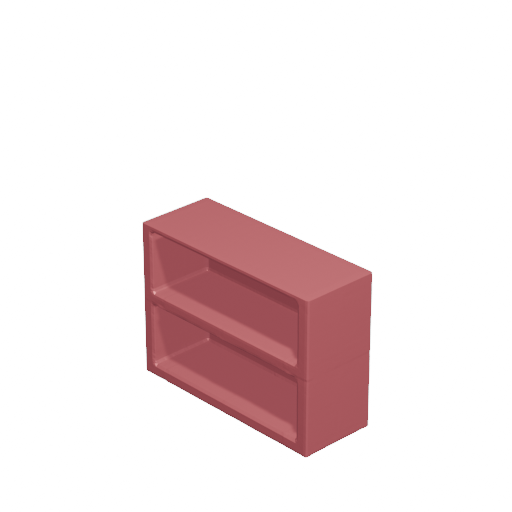}
         \\
         \rotatebox{90}{\parbox{0.11\linewidth}{\centering AutoSDF \citep{AutoSDF_Mittal_2022_CVPR}}}
         & \includegraphics[width=0.11\linewidth]{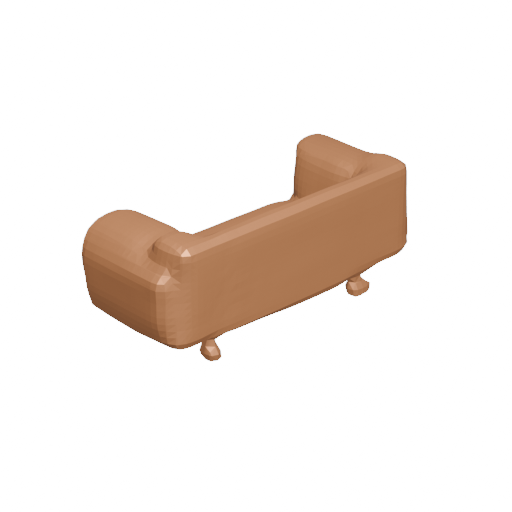}
         & \includegraphics[width=0.11\linewidth]{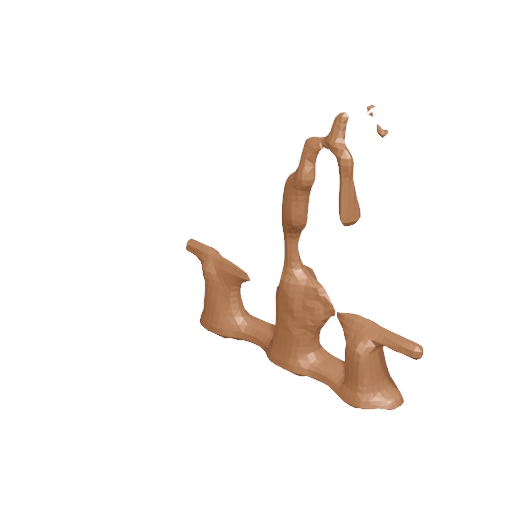}
         & \includegraphics[width=0.11\linewidth]{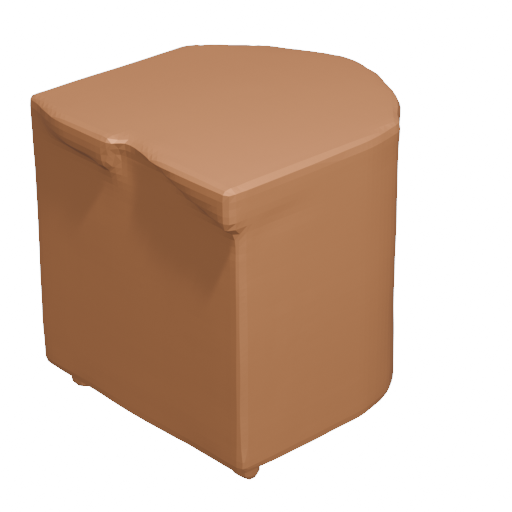}
         & \includegraphics[width=0.11\linewidth]{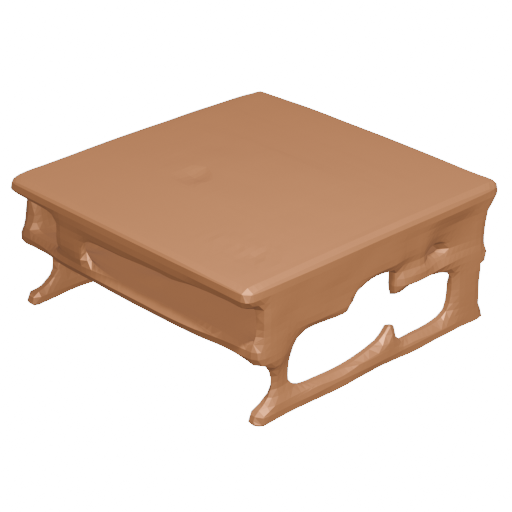}
         & \includegraphics[width=0.11\linewidth]{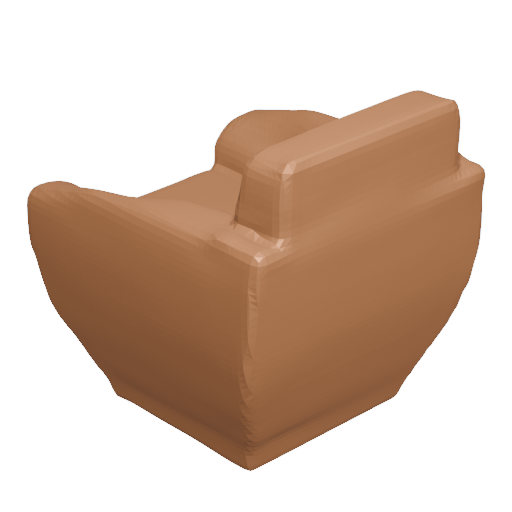}
         & \includegraphics[width=0.11\linewidth]{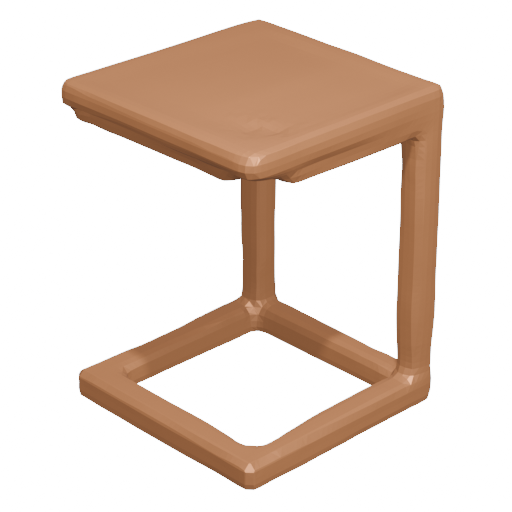}
         & \includegraphics[width=0.11\linewidth]{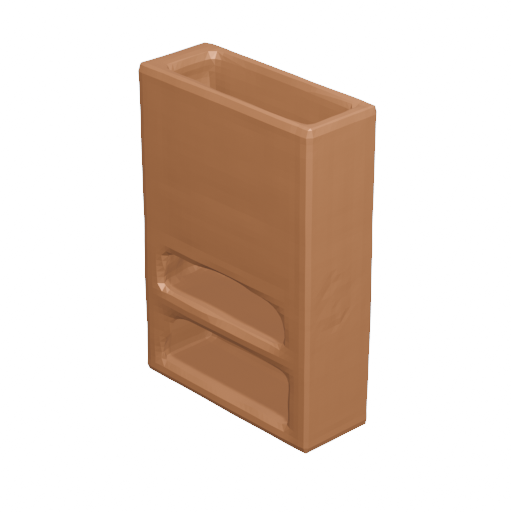}
         \\
         \rotatebox{90}{\parbox{0.11\linewidth}{\centering Anchor\-Former \citep{Chen2023AnchorFormer}}}
         & \includegraphics[width=0.11\linewidth]{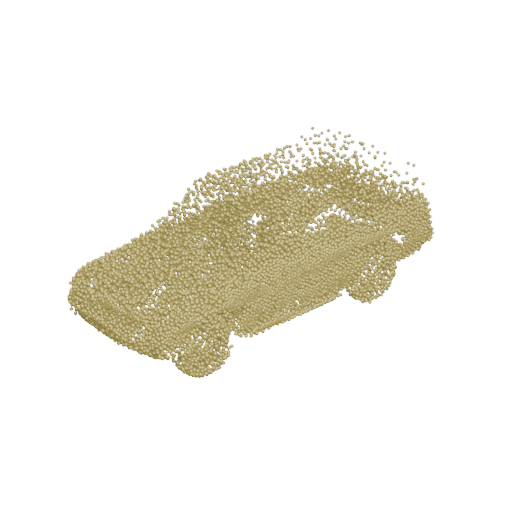}
         & \includegraphics[width=0.11\linewidth]{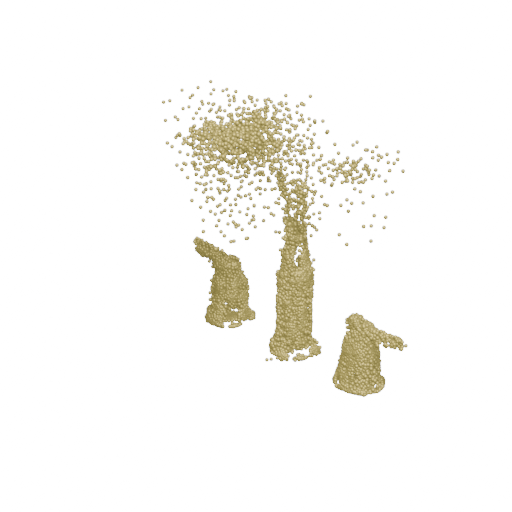}
         & \includegraphics[width=0.11\linewidth]{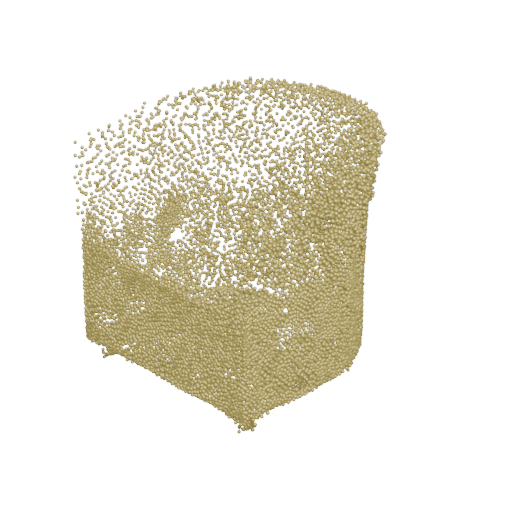}
         & \includegraphics[width=0.11\linewidth]{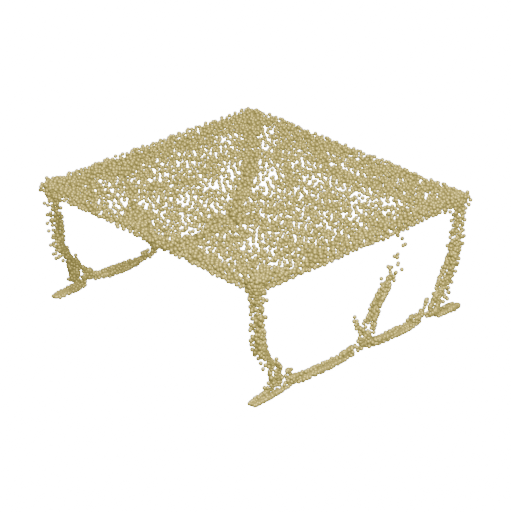}
         & \includegraphics[width=0.11\linewidth]{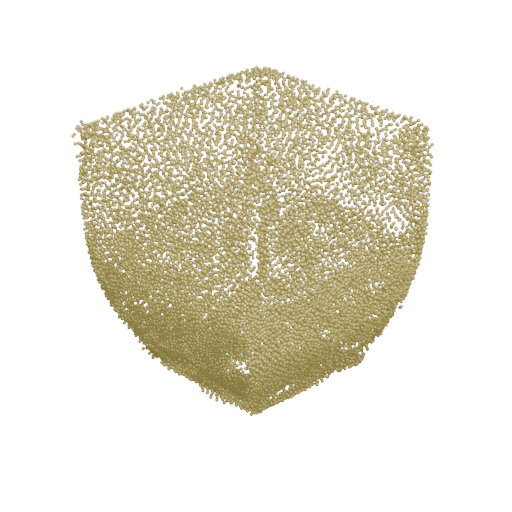}
         & \includegraphics[width=0.11\linewidth]{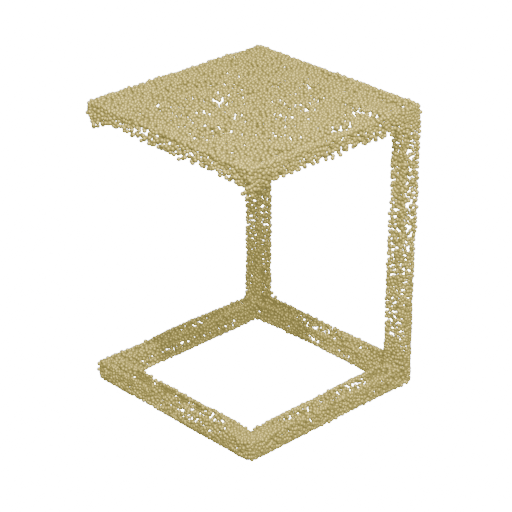}
         & \includegraphics[width=0.11\linewidth]{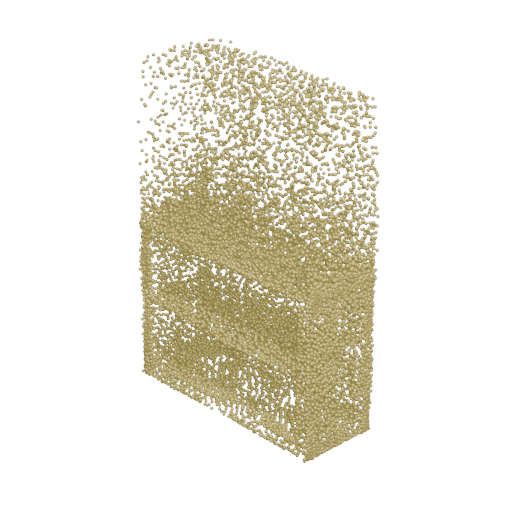}
         \\
         \rotatebox{90}{\parbox{0.11\linewidth}{\centering SLT (Ours)}} %
         & \includegraphics[width=0.11\linewidth]{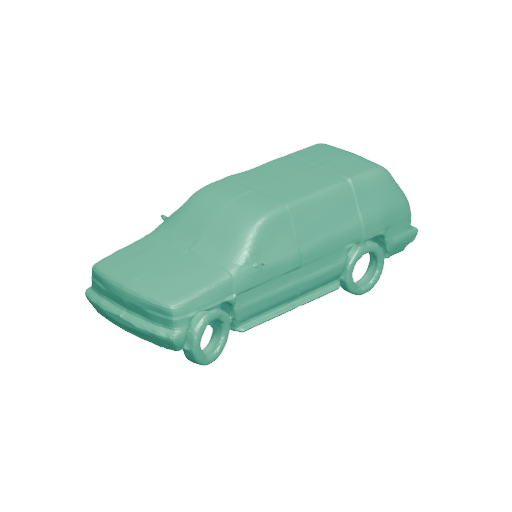}
         & \includegraphics[width=0.11\linewidth]{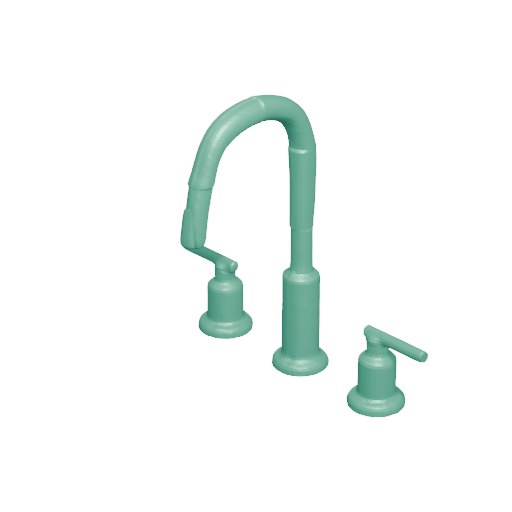}
         & \includegraphics[width=0.11\linewidth]{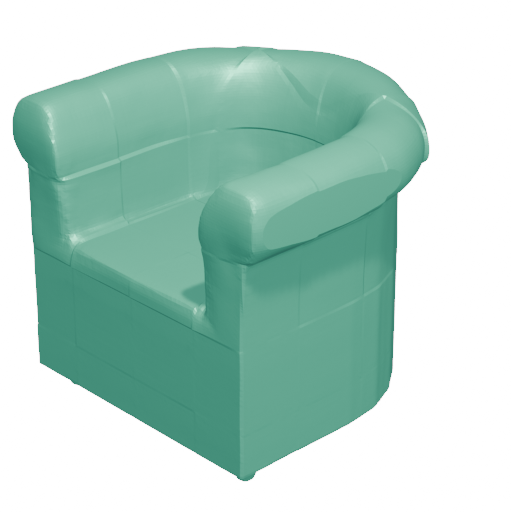}
         & \includegraphics[width=0.11\linewidth]{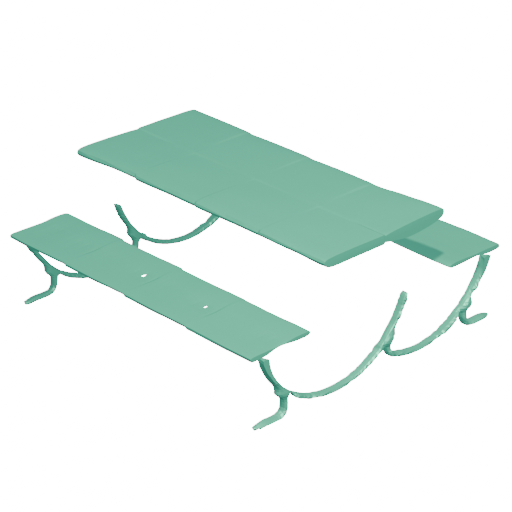}
         & \includegraphics[width=0.11\linewidth]{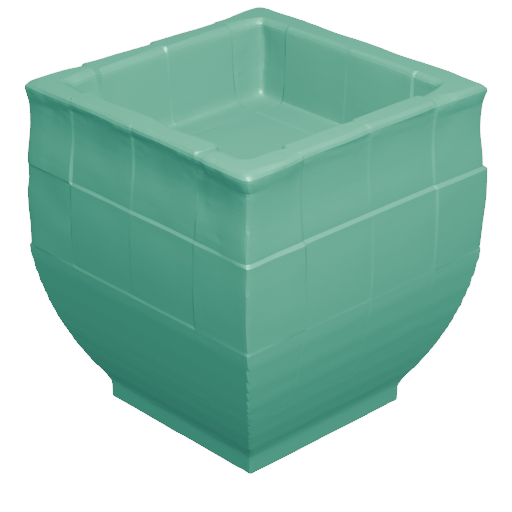}
         & \includegraphics[width=0.11\linewidth]{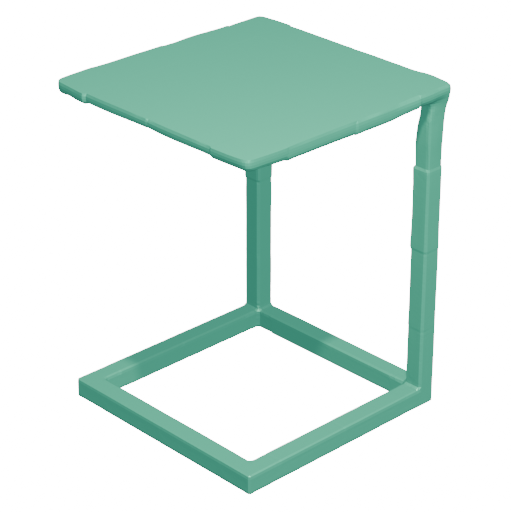}
         & \includegraphics[width=0.11\linewidth]{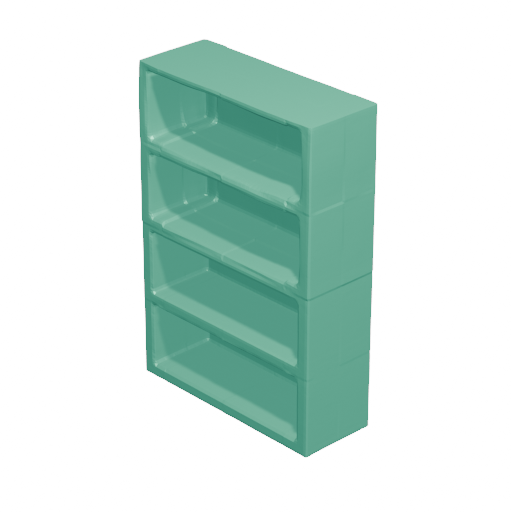}
         \\
         \rotatebox{90}{\parbox{0.11\linewidth}{\centering GT}}
         & \includegraphics[width=0.11\linewidth]{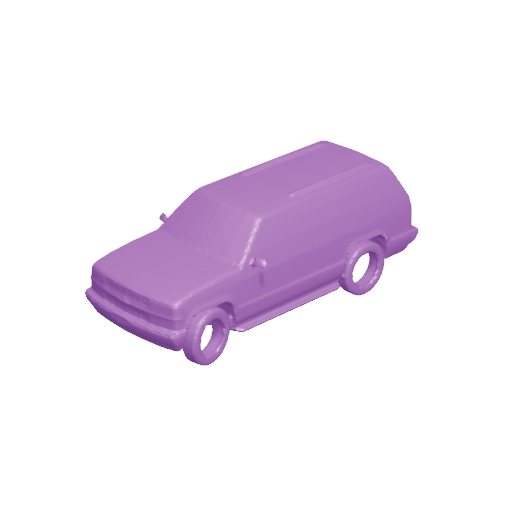}
         & \includegraphics[width=0.11\linewidth]{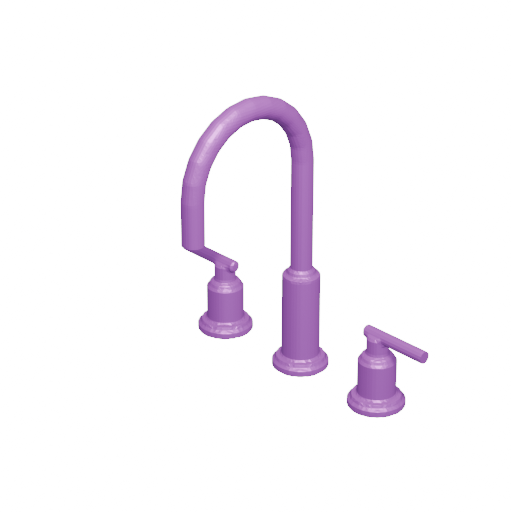}
         & \includegraphics[width=0.11\linewidth]{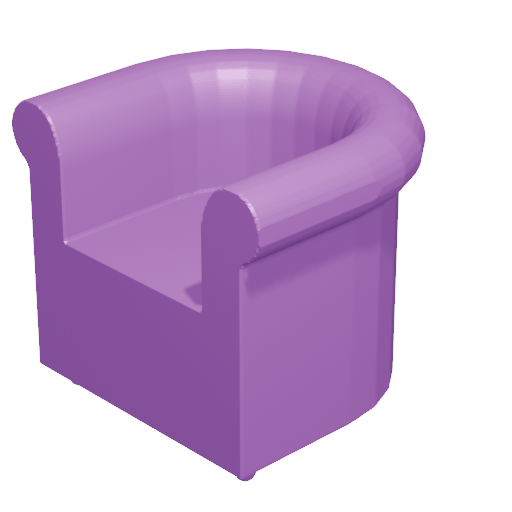}
         & \includegraphics[width=0.11\linewidth]{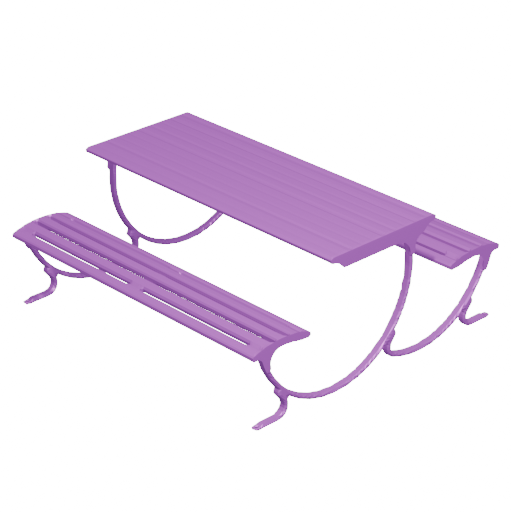}
         & \includegraphics[width=0.11\linewidth]{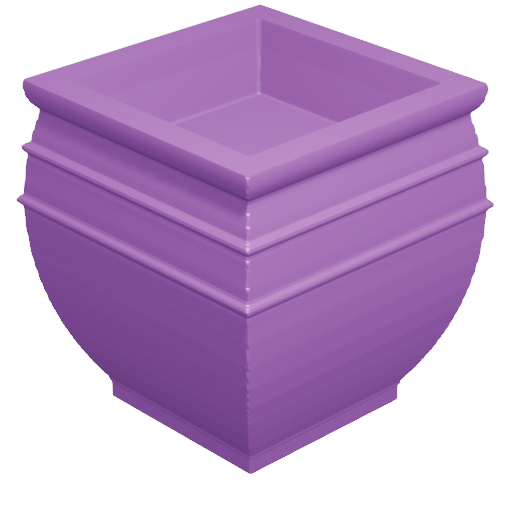}
         & \includegraphics[width=0.11\linewidth]{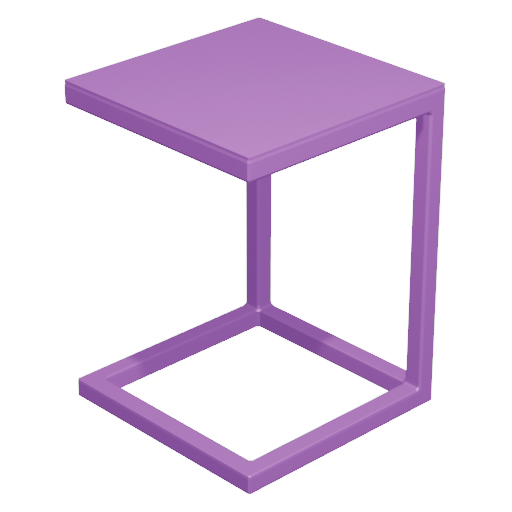}
         & \includegraphics[width=0.11\linewidth]{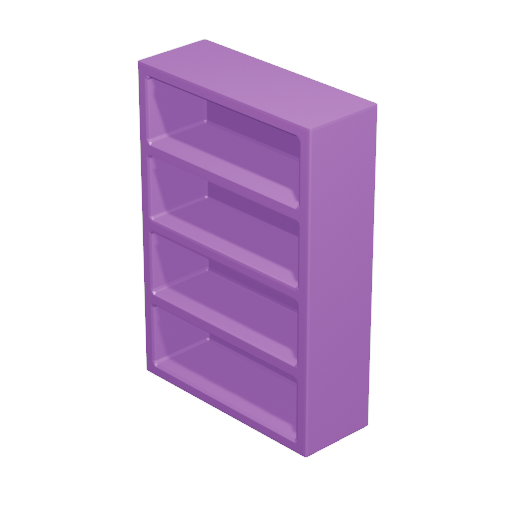}
         \\
    \end{tabular}
    \caption{Completion of ShapeNet~\citep{ShapeNet} objects from bottom half.
    Comparison to AutoSDF~\citep{AutoSDF_Mittal_2022_CVPR} and AnchorFormer~\cite{Chen2023AnchorFormer}.
    Our SLT completes these objects more plausibly than AutoSDF.
    The density of completed points by AnchorFromer drastically varies in the completed regions.}
    \label{fig:eval:completion_shape_net}
\end{figure}

\paragraph{Qualitative Results (ShapeNet).} Examples for the previously
described bottom half (Half) completion task on ShapeNetCoreV1~\citep{ShapeNet}
are shown in Figure~\ref{fig:eval:completion_shape_net}. Here, we compare our
SLT against AutoSDF~\citep{AutoSDF_Mittal_2022_CVPR} and
AnchorFormer~\citep{Chen2023AnchorFormer} as recent state-of-the-art methods.

Our POC-SLT pipeline generates highly plausible shapes.
Both AutoSDF~\citep{AutoSDF_Mittal_2022_CVPR} and AnchorFormer~\citep{Chen2023AnchorFormer} struggle to fill in the missing parts. 
Compared to AutoSDF, our method works at a significantly higher resolution and can handle all ShapeNet classes.
The higher resolution is for example important for the fine structures on the bench table in column four or the faucet in column two.
AutoSDF~\citep{AutoSDF_Mittal_2022_CVPR} completely fails to reconstruct the car in column one or the armchair in
column three, while our SLT on the same input produces correct results.
AnchorFormer~\citep{Chen2023AnchorFormer} often produces sparse and simplistic completions for inputs with unknown top halves which is especially
visible for the faucet in column two.
However, the sparseness and therefore lack of detail are also visible in the car in the first column,
the armchair in the second, and the cupboard in the last column.

\begin{figure}[p]
    \centering
    \vspace*{-1em}
    \begin{tabular}{c@{}c@{}c@{}c@{}c c@{}c@{}c@{}c}
        \rotatebox{90}{\parbox{0.11\linewidth}{\centering Input}}
        & \includegraphics[width=0.11\linewidth]{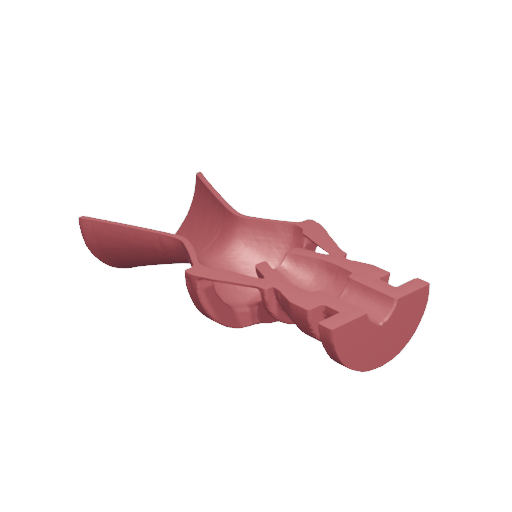}
        & \includegraphics[width=0.11\linewidth]{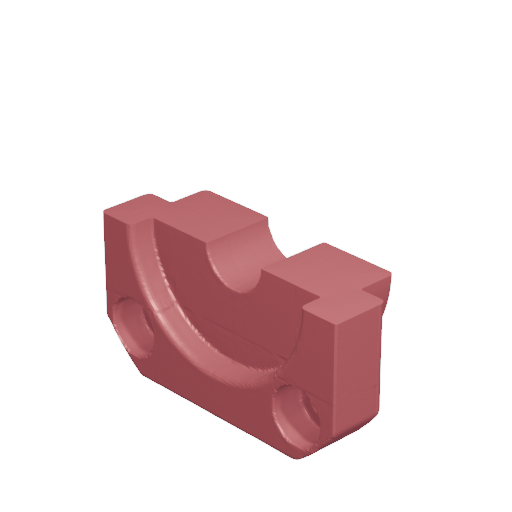}
        & \includegraphics[width=0.11\linewidth]{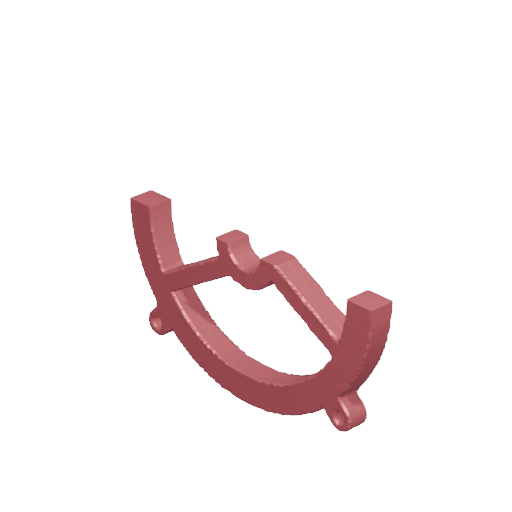}
        & \includegraphics[width=0.11\linewidth]{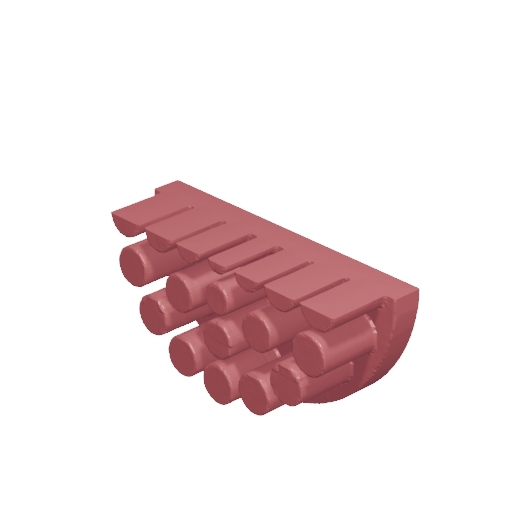}
        & \includegraphics[width=0.11\linewidth]{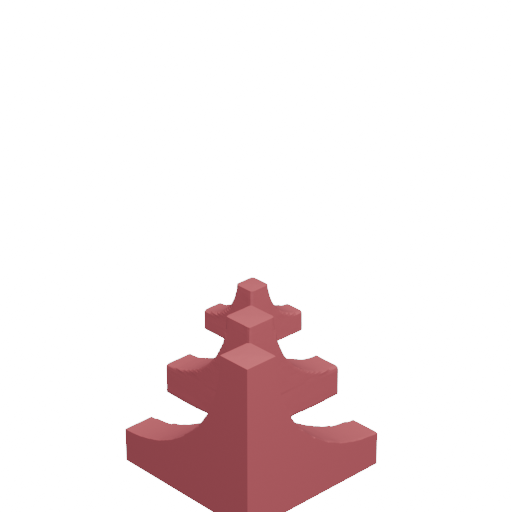}
        & \includegraphics[width=0.11\linewidth]{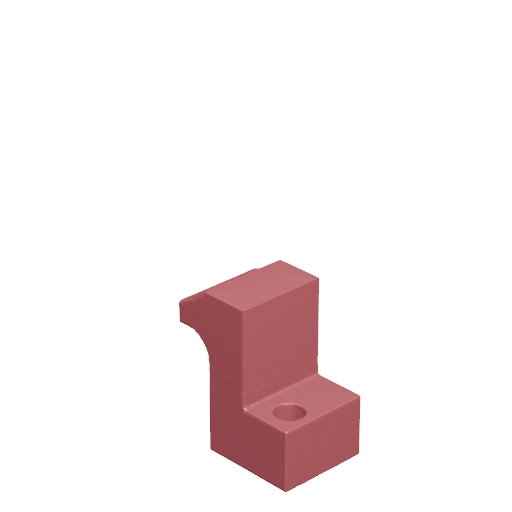}
        & \includegraphics[width=0.11\linewidth]{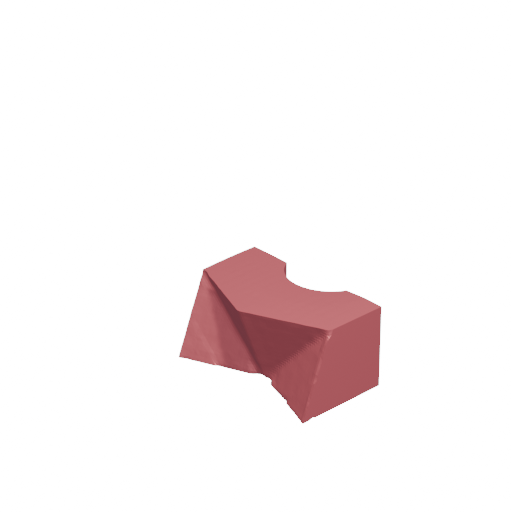}
        & \includegraphics[width=0.11\linewidth]{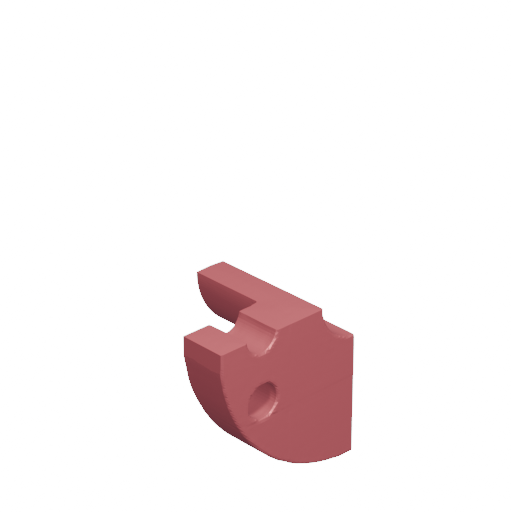}
        \\
        \rotatebox{90}{\parbox{0.11\linewidth}{\centering SLT ABC}}
        & \includegraphics[width=0.11\linewidth]{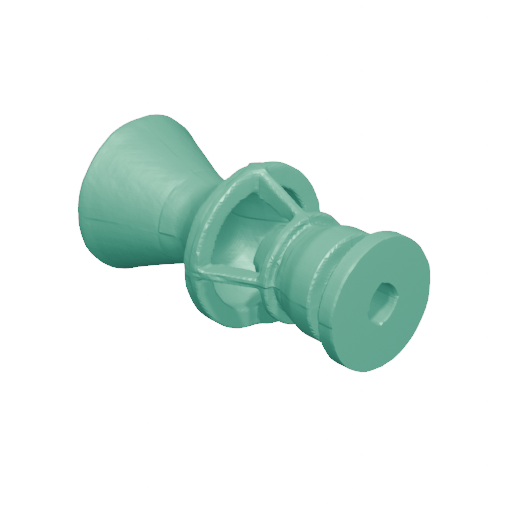}
        & \includegraphics[width=0.11\linewidth]{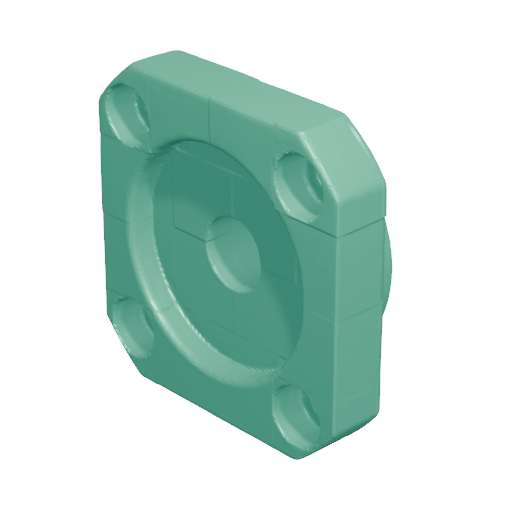}
        & \includegraphics[width=0.11\linewidth]{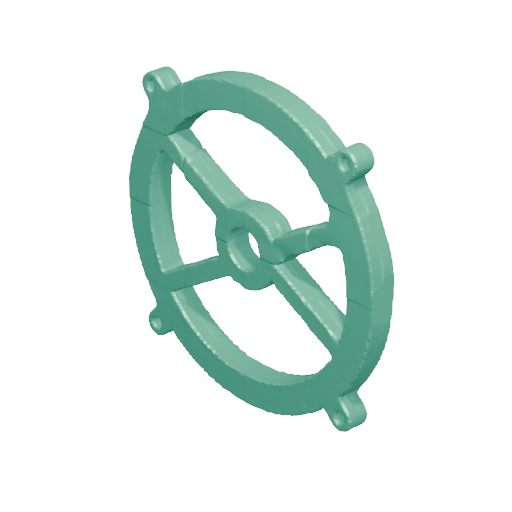}
        & \includegraphics[width=0.11\linewidth]{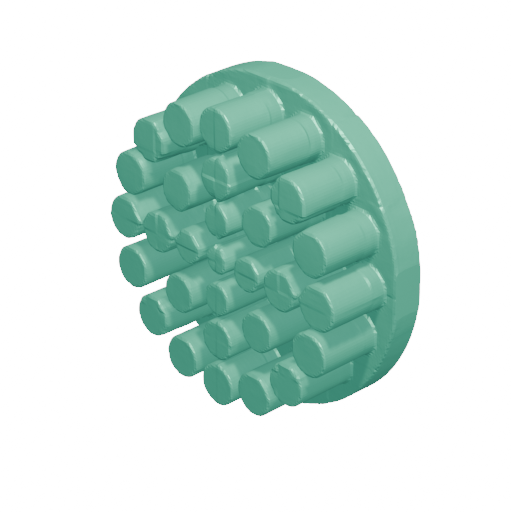}
        & \includegraphics[width=0.11\linewidth]{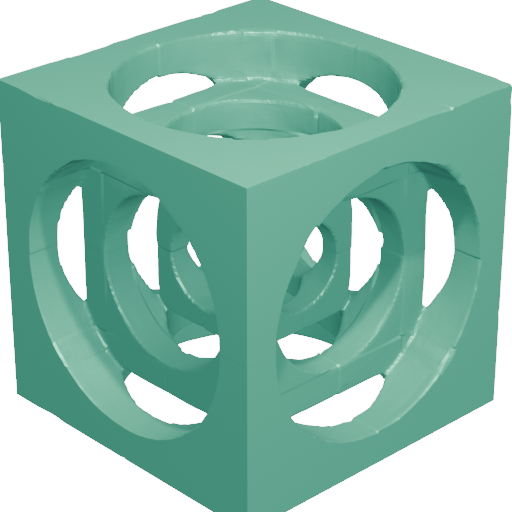}
        & \includegraphics[width=0.11\linewidth]{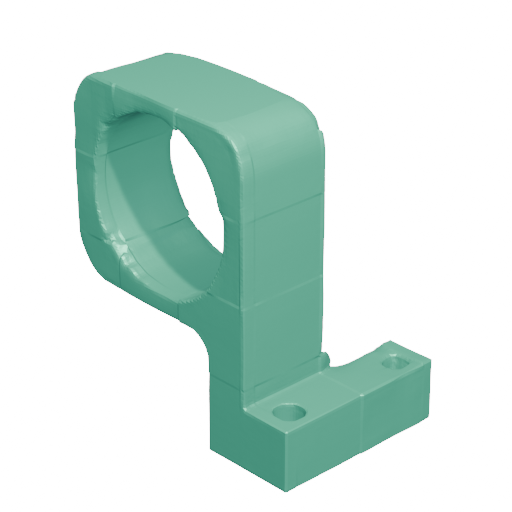}
        & \includegraphics[width=0.11\linewidth]{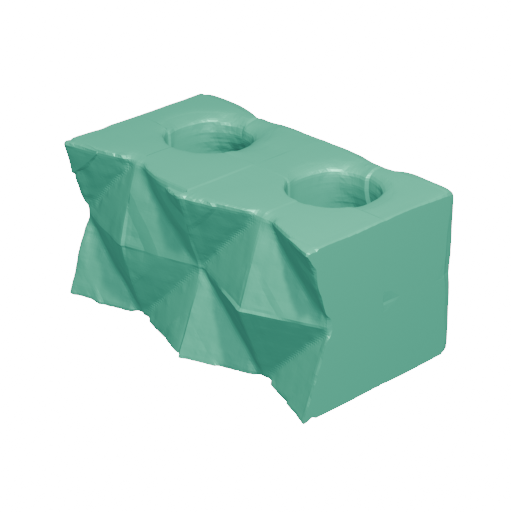}
        & \includegraphics[width=0.11\linewidth]{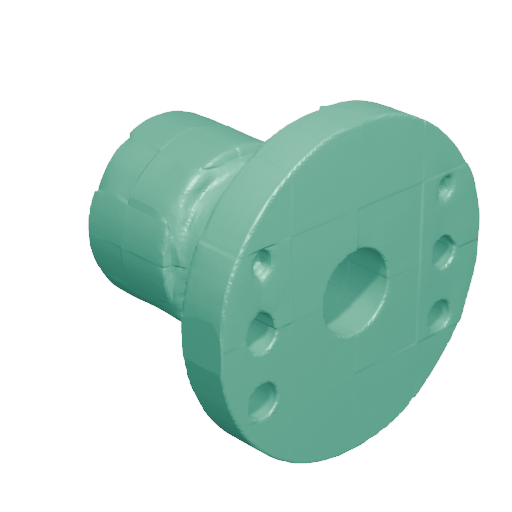}
        \\
        \rotatebox{90}{\parbox{0.11\linewidth}{\centering GT}}
        & \includegraphics[width=0.11\linewidth]{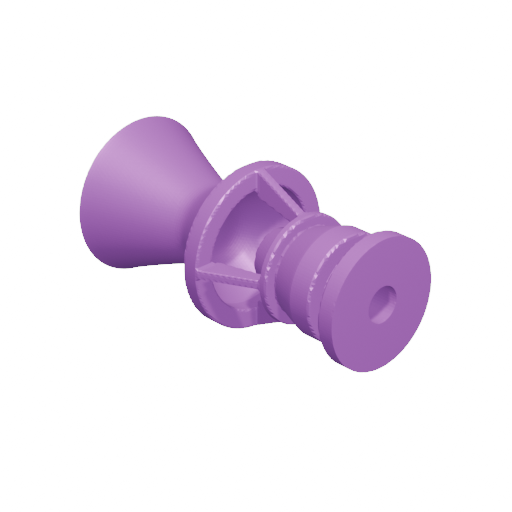}
        & \includegraphics[width=0.11\linewidth]{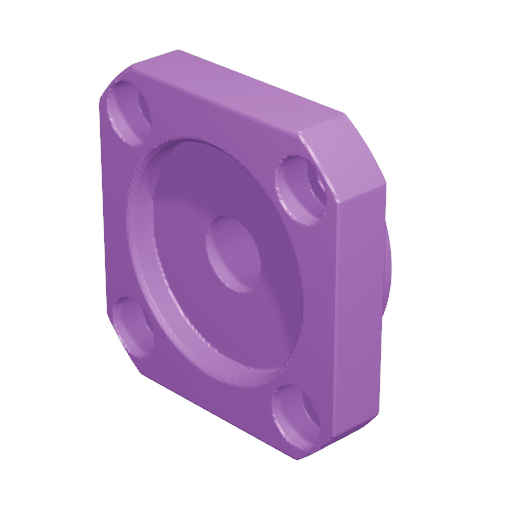}
        & \includegraphics[width=0.11\linewidth]{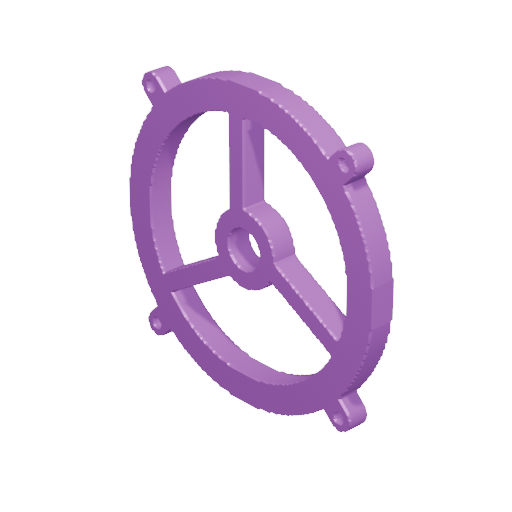}
        & \includegraphics[width=0.11\linewidth]{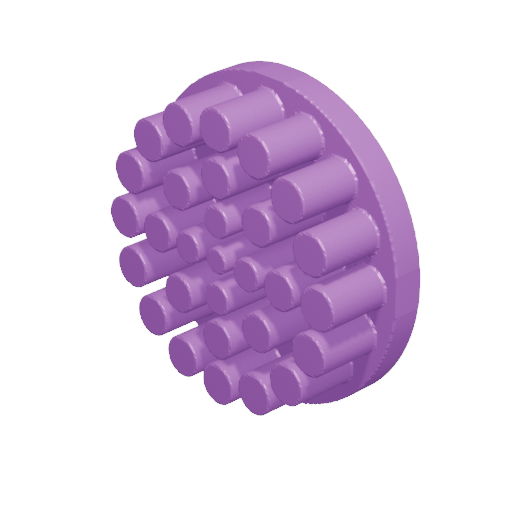}
        & \includegraphics[width=0.11\linewidth]{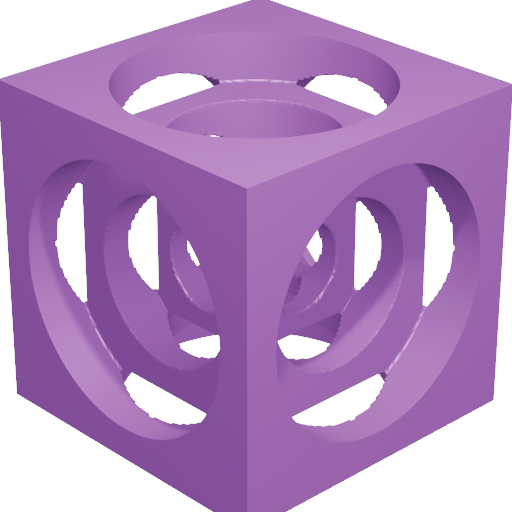}
        & \includegraphics[width=0.11\linewidth]{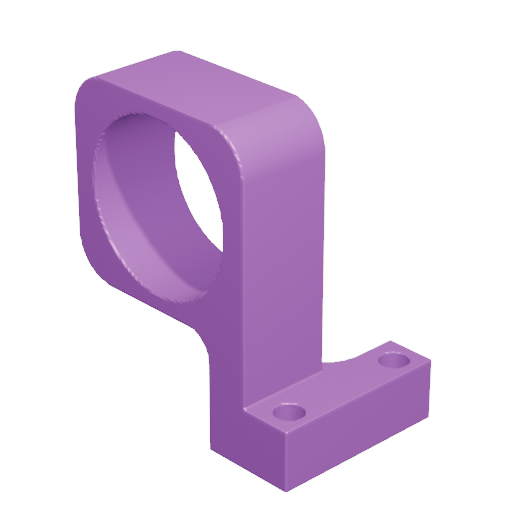}
        & \includegraphics[width=0.11\linewidth]{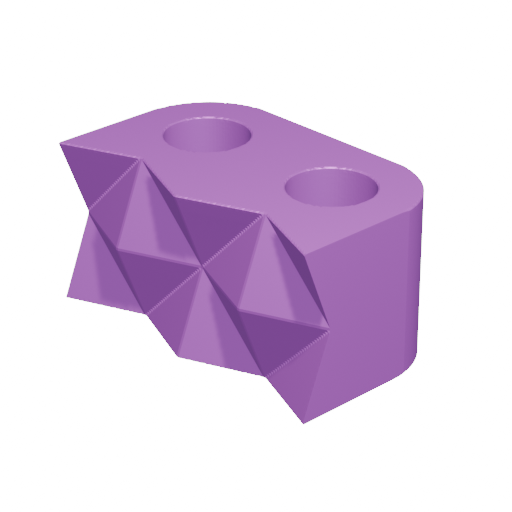}
        & \includegraphics[width=0.11\linewidth]{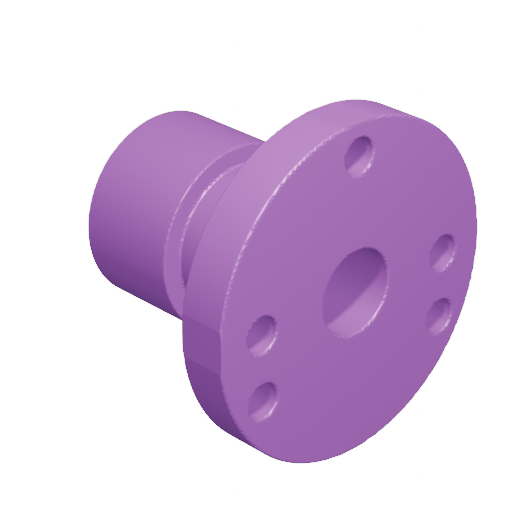}
        \\
    \end{tabular}
    \caption{Completion of ABC~\citep{ABC_Koch_2019_CVPR} objects from the bottom half (left) and octant (right).
    The SLT learned to complete partially symmetric objects quite successfully.}
    \label{fig:eval:completion_abc}
\end{figure}

\paragraph{Qualitative Results (ABC).}
Shape completion results from halves (Half) or octants (Oct) on
ABC~\citep{ABC_Koch_2019_CVPR} are visualized in Figure~\ref{fig:eval:completion_abc}.
For this task, we use SLT ABC, which was fine-tuned on the ABC dataset.
The ABC dataset contains many planar and rotationally symmetric objects. These symmetries are picked up by the SDF-Latent-Transformer to complete missing parts with high detail and high plausibility, such as the object in column one or column four.

On the other hand, the provided input does not always constrain the output sufficiently. This can lead to deviations when compared to ground truth while still generating plausible objects, such as the upper spokes in the third column, the unsymmetric object in the sixth column, or the missing hole in
the last column of the figure. 

\begin{figure}[p]
    \centering
    \vspace*{-1em}
    \begin{tabular}{c@{}c@{\!}c@{\!}c@{\!}c@{\!}c@{\!}c@{}c@{\!}c@{\!}c}
          \rotatebox{90}{\parbox{0.11\linewidth}{\centering Input}}
        & \includegraphics[width=0.11\linewidth]{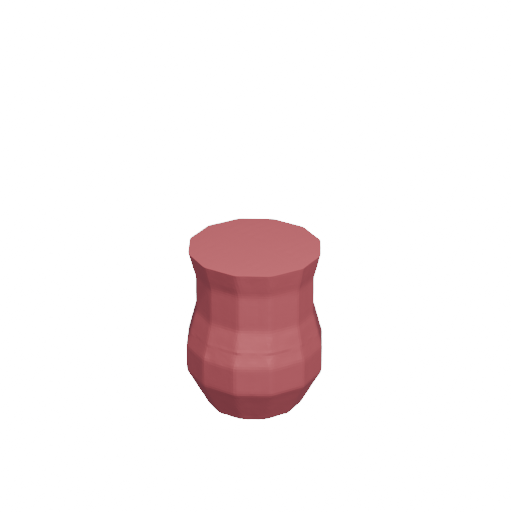}
        & \includegraphics[width=0.11\linewidth]{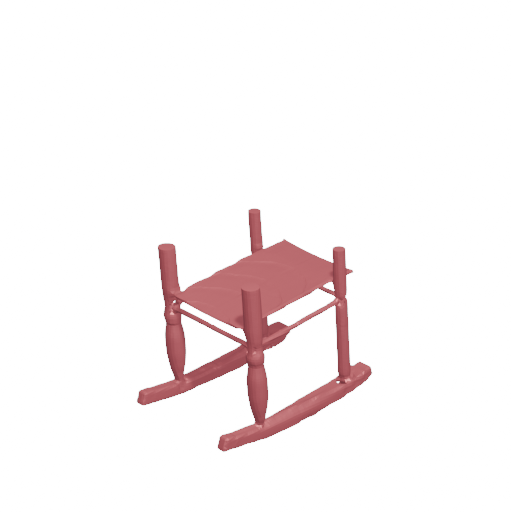}
        & \includegraphics[width=0.11\linewidth]{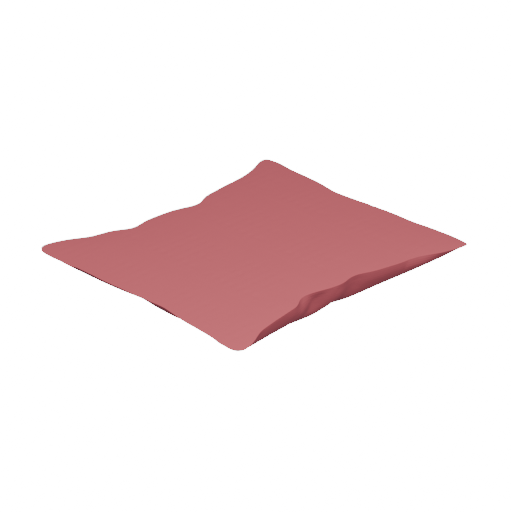}
        & \includegraphics[width=0.11\linewidth]{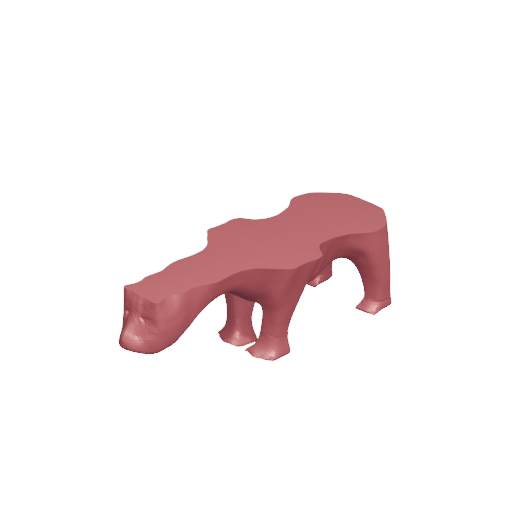}
        & \includegraphics[width=0.11\linewidth]{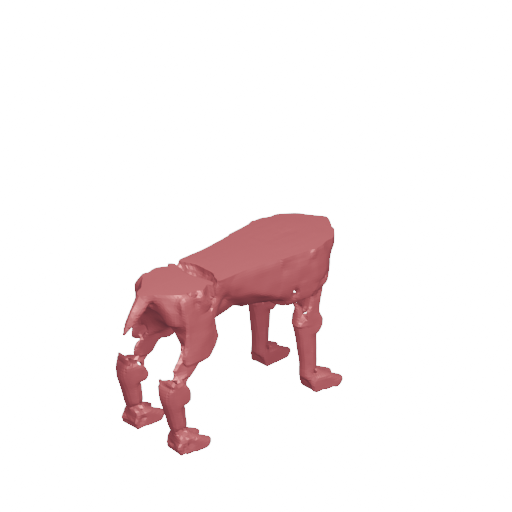}
        & \includegraphics[width=0.11\linewidth]{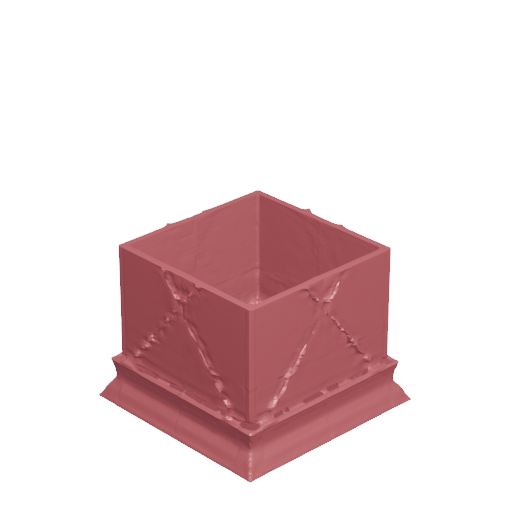}
        & \includegraphics[width=0.11\linewidth]{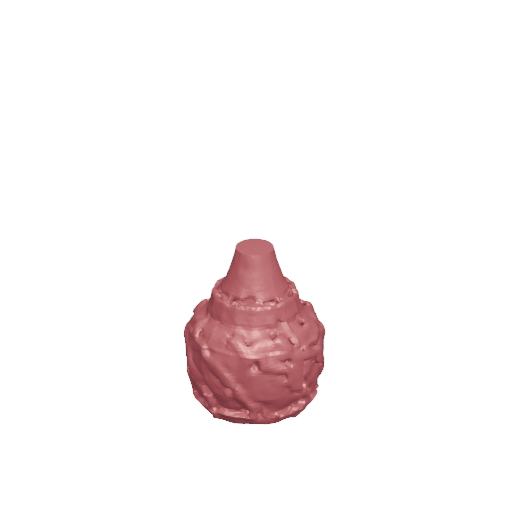}
        & \includegraphics[width=0.11\linewidth]{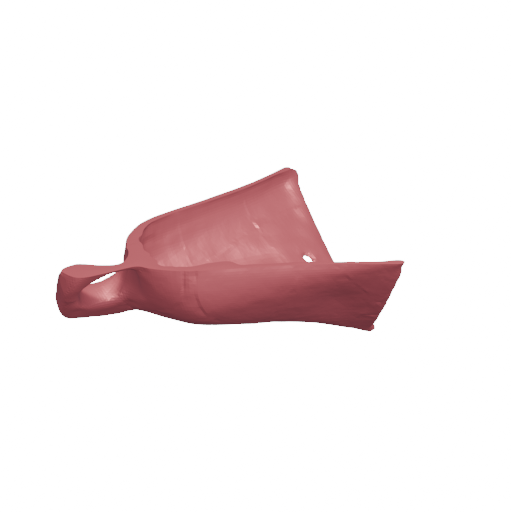}
        & \includegraphics[width=0.11\linewidth]{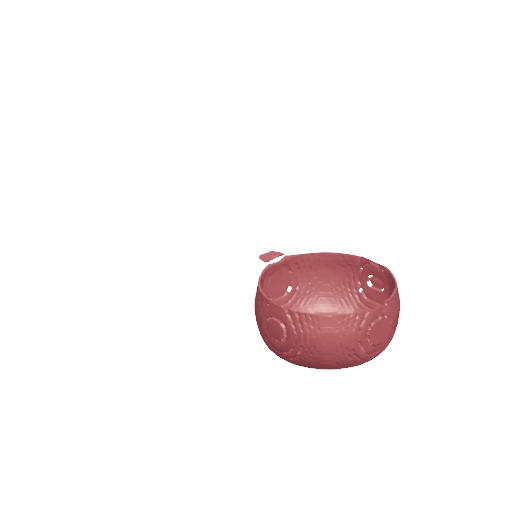}
        \\
          \rotatebox{90}{\parbox{0.11\linewidth}{\centering SLT}}
        & \includegraphics[width=0.11\linewidth]{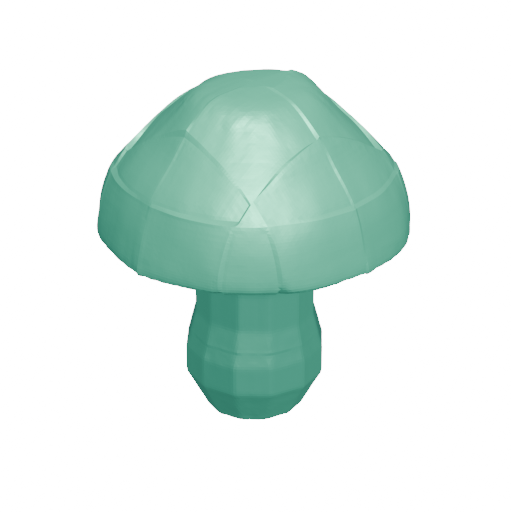}
        & \includegraphics[width=0.11\linewidth]{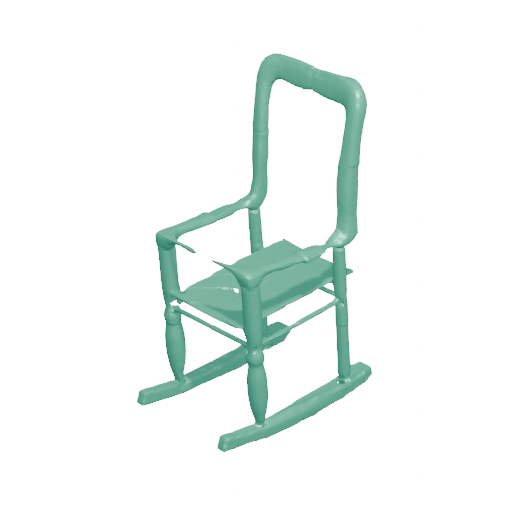}
        & \includegraphics[width=0.11\linewidth]{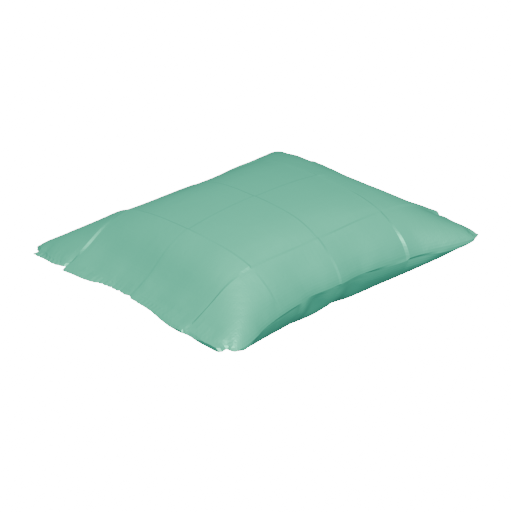}
        & \includegraphics[width=0.11\linewidth]{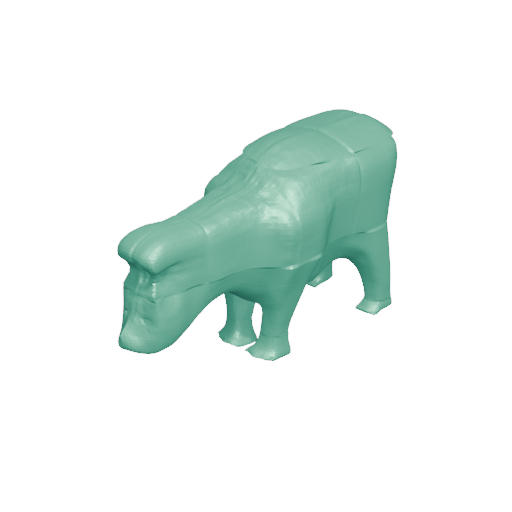}
        & \includegraphics[width=0.11\linewidth]{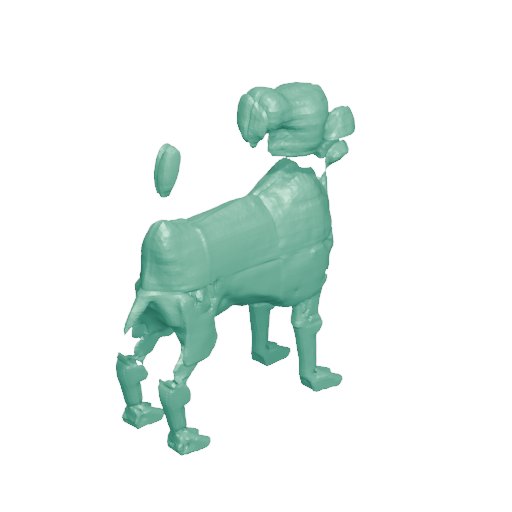}
        & \includegraphics[width=0.11\linewidth]{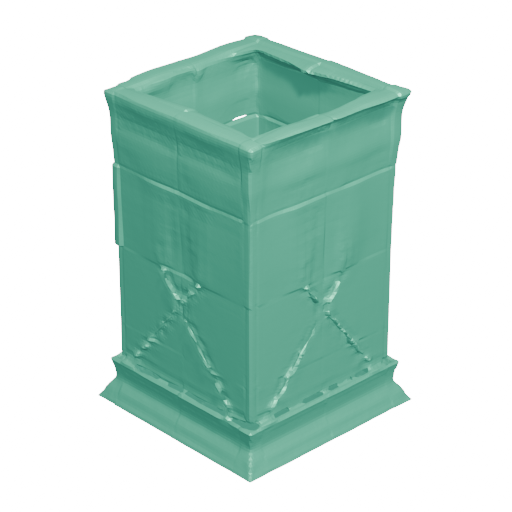}
        & \includegraphics[width=0.11\linewidth]{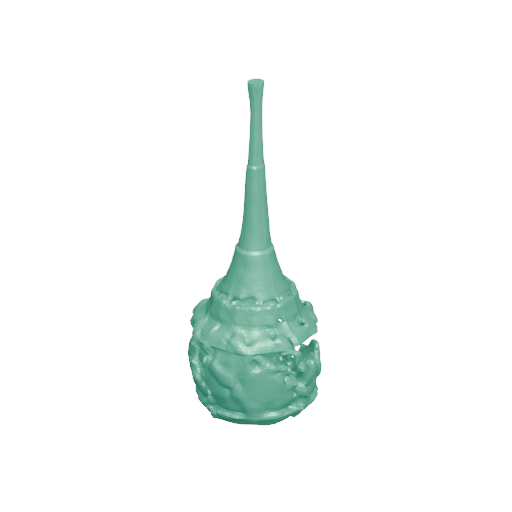}
        & \includegraphics[width=0.11\linewidth]{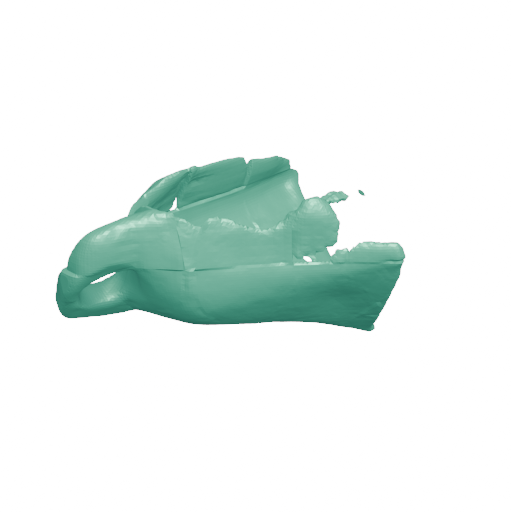}
        & \includegraphics[width=0.11\linewidth]{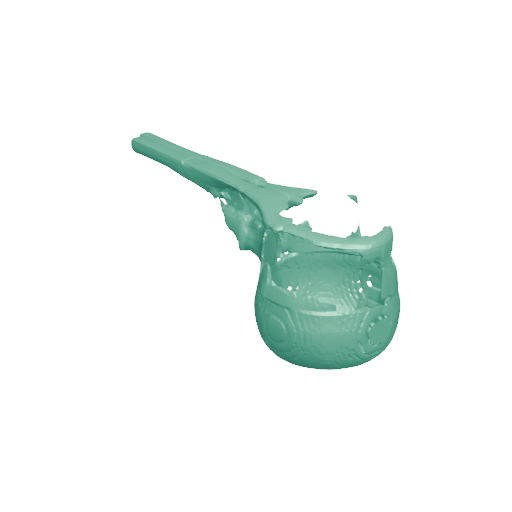}
        \\
          \rotatebox{90}{\parbox{0.11\linewidth}{\centering GT}}
        & \includegraphics[width=0.11\linewidth]{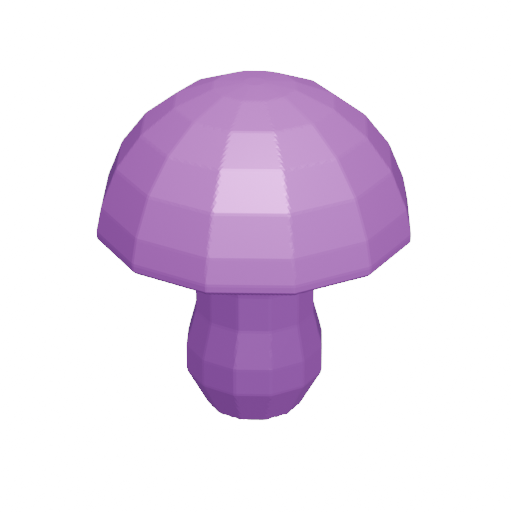}
        & \includegraphics[width=0.11\linewidth]{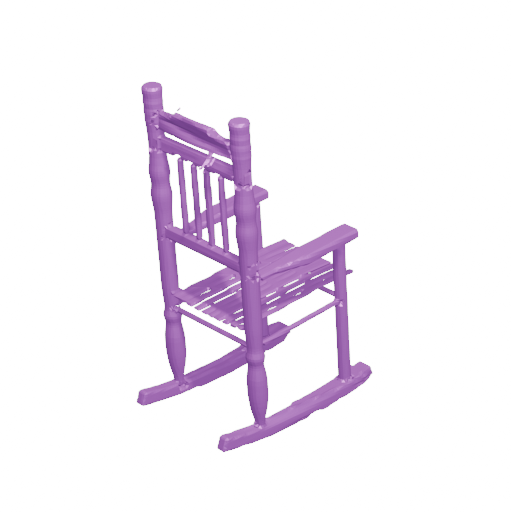}
        & \includegraphics[width=0.11\linewidth]{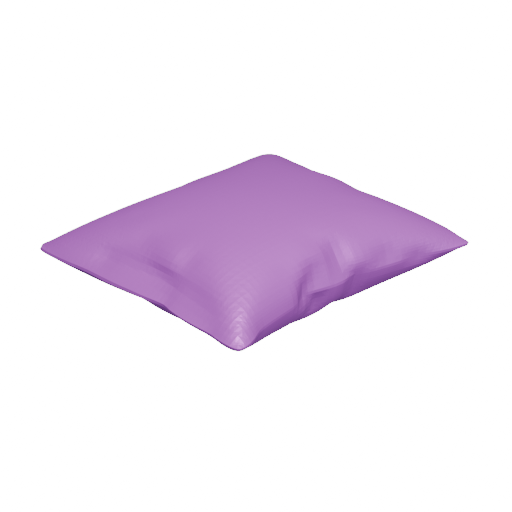}
        & \includegraphics[width=0.11\linewidth]{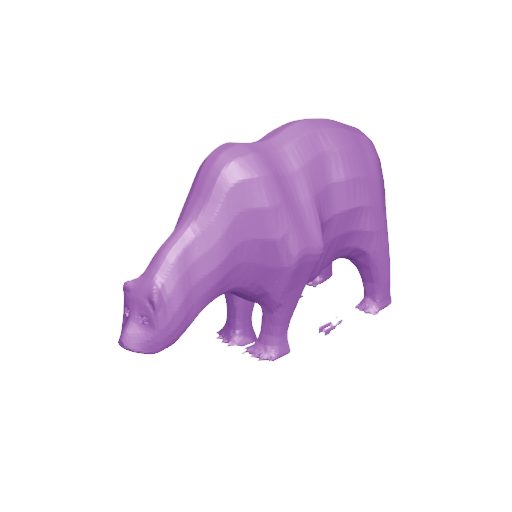}
        & \includegraphics[width=0.11\linewidth]{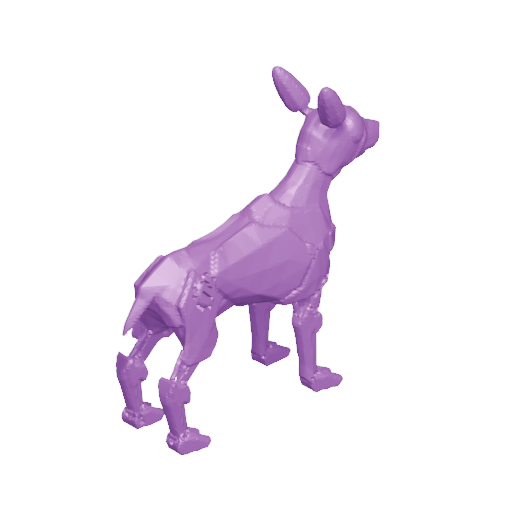}
        & \includegraphics[width=0.11\linewidth]{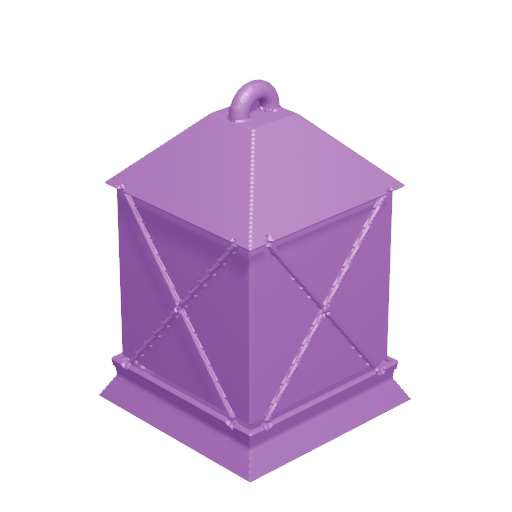}
        & \includegraphics[width=0.11\linewidth]{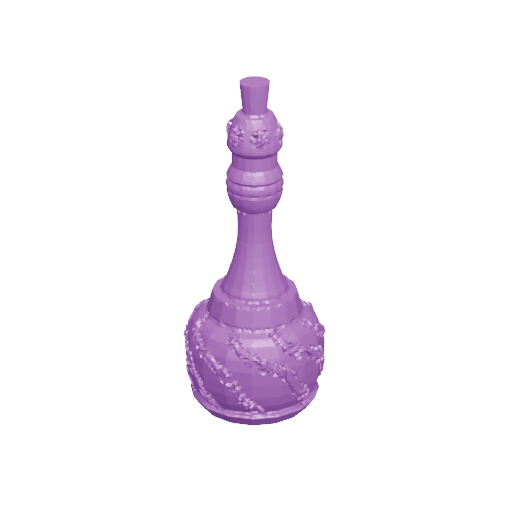}
        & \includegraphics[width=0.11\linewidth]{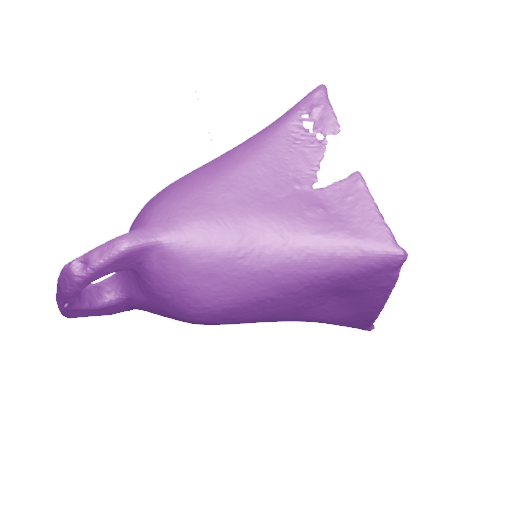}
        & \includegraphics[width=0.11\linewidth]{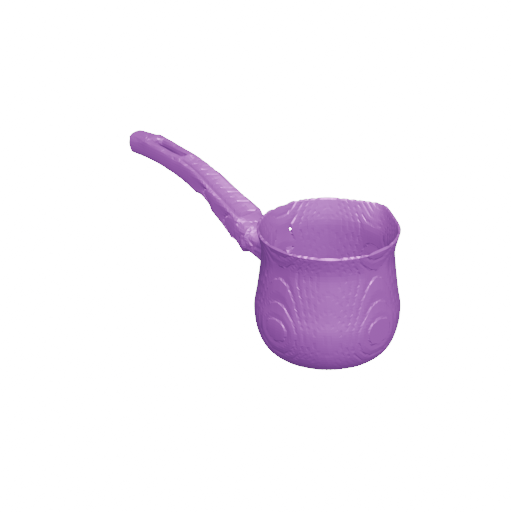}
        \\
    \end{tabular}
    \caption{Completion of out-of-distribution objects from Objaverse~\citep{Deitke2023Objaverse} using the SLT trained on ShapeNet~\citep{ShapeNet}.
    The last three columns show completions of scanned real-world objects.}
    \label{fig:eval:objaverse}
\end{figure}

\paragraph{Generalization (Objaverse).}
To demonstrate the generalizability of our approach we evaluate the SLT, trained only on ShapeNet~\citep{ShapeNet},
on the (Half) task with objects from Objaverse~\citep{Deitke2023Objaverse}.
The type of objects shown in Figure~\ref{fig:eval:objaverse} have never been seen by the SLT. 
As expected, the results show clear deviations from the ground truth,
especially for objects which are far away from the kind of objects that are found in ShapeNet,
such as the animals in column four and five.
For simpler objects, such as the pillow in column three or the mushroom in column one,
our approach produces good completions.
There is a chair category in ShapeNet, which allows our model to plausibly complete the rocking chair in column two.

\paragraph{Quantitative Results.}
We numerically evaluate the previously described tasks in Table~\ref{tab:eval:completion}.

\begin{table*}[htb]
  \caption{Shape Completion.
  Evaluation on completion tasks (Half), (Oct), (R75), (R50) and (R25) with the
  ShapeNetCoreV1~\citep{ShapeNet} dataset with all 55 categories and on (Half)
  and (Oct) with the ABC~\citep{ABC_Koch_2019_CVPR} dataset and .
  The results report the mean over all categories.
  Details on the metrics can be found in Appendix~\ref{appendix:metrics}.}
  \label{tab:eval:completion}
  \centering
  \begin{tabular}{llccccccccccccccc}
    \toprule
    Model
    & Dataset
    & Task 
    & IoU$\uparrow$ &$F_1$$\uparrow$ &CD$\downarrow$ &HD$\downarrow$  &NC$\uparrow$ &IN$\downarrow$& CMP$\uparrow$
    \\

    \midrule                  %
    SLT & SN & Half & 0.7466 & 0.8468 & 1.0221 & 0.0765 & 0.9200 & 0.4196 & 0.9067 \\
    SLT & SN & Oct  & 0.5884 & 0.7336 & 1.2467 & 0.0966 & 0.8589 & 0.6034 & 0.8404 \\
    SLT & SN & R75  & 0.9153 & 0.9792 & 0.2258 & 0.0452 & 0.9677 & 0.2862 & 0.9905 \\
    SLT & SN & R50  & 0.8650 & 0.9495 & 0.4829 & 0.0595 & 0.9504 & 0.3512 & 0.9751 \\
    SLT & SN & R25  & 0.7645 & 0.8677 & 0.8567 & 0.0789 & 0.9183 & 0.4512 & 0.9329 \\
    
    \midrule                 %
    SLT ABC   & ABC & Half & 0.8617 & 0.9159 & 0.8703 & 0.0575 & 0.9435 & 0.2551 & 0.9466 \\
    SLT ABC   & ABC & Oct  & 0.7144 & 0.7744 & 2.9247 & 0.1077 & 0.8779 & 0.3986 & 0.8391 \\

        \bottomrule
  \end{tabular}
\end{table*}

The results on the aforementioned benchmark from
\citet{MPC} are reported in Table~\ref{tab:eval:Shapenet_13_UHD}.
Numbers from previous works are taken from~\citet{AutoSDF_Mittal_2022_CVPR}.
The numbers for the \emph{chair} category
are given explicitly to compare to previous work. Our SLT outperforms all
other methods. Even when evaluated on all 13 categories, there remains a
significant improvement. 

\begin{table}[htb]
  \caption{Quantitative comparison on shape completion on the \emph{chair} category of
  the ShapeNetCoreV1-subset with 13 categories from~\citet{MPC}. Completion
  tasks are (Half) and (Oct). In compliance with previous work, we report the
  Unidirectional Hausdorff Distance from input to completed result.
}
  \label{tab:eval:Shapenet_13_UHD}
  \centering
  \begin{tabular}{ccc}
    \toprule
    Method/UHD$\downarrow$ &Half &Oct\\
    \midrule
    MPC~\citep{MPC} & 0.0627 & 0.0579\\
    PoinTR~\citep{PoinTr_Yu_2021_ICCV} & 0.0572 & 0.0536 \\
    AutoSDF~\citep{AutoSDF_Mittal_2022_CVPR} & 0.0567 & 0.0599 \\
    SDFusion~\citep{SDFusion_Cheng_2023_CVPR} & 0.0557 & - \\
    SLT (ours)& \textbf{0.0445} &  \textbf{0.0467}\\
    \midrule
    SLT (ours) all 13 categories  & \textbf{0.0379} & \textbf{0.0399}\\
  \bottomrule
\end{tabular}
\end{table}

\subsection{P-VAE \& AutoDecoder}
\label{sec:Latent-Code Refinment}
The P-VAE accepts SDF patches of size $32^3$ as input, encodes them into compressed latent
representations and decodes latent representations back to SDF patches.
It was trained and evaluated on a training split of ShapeNet. We evaluate the P-VAE
on unseen data in Table~\ref{tab:eval:pvae}. The numbers demonstrate the high
quality of our latent space. Note that the P-VAE was not fine-tuned for the
following experiments on ABC~\citep{ABC_Koch_2019_CVPR} or
Objaverse~\citep{Deitke2023Objaverse} which shows the strong generalizability
resulting from our training scheme.
\begin{table*}[htb]
    \caption{P-VAE encode-decode performance evaluated on ShapeNet~\citep{ShapeNet}.}
    \label{tab:eval:pvae}
    \centering
      \begin{tabular}{crccccccc}
        \toprule
        IoU$\uparrow$ &$F_1$$\uparrow$ &CD$\downarrow$ &HD$\downarrow$
        \\
        \midrule
        0.9482 & 0.9931 & 0.0001 & 0.0264 \\ 
        \midrule
      \end{tabular}
\end{table*}

\citet{DeepSDF_Park_2019_CVPR} suggested to use an AutoDecoder
(AD)~\citep{AutoDecoder_osti_80034} to optimize the latent codes for a known
SDF such that the frozen decoder $D_\mathrm{VAE}$ produces the best possible
result. The optimized latent codes should yield more accurate surface
reconstruction than the codes directly produced by the encoder $E_\mathrm{VAE}$
in a single forward pass. However, the time spent on refinement is orders of
magnitude longer, e.g.\ around $0.04$ sec for the P-VAE encoder vs.\ 25 sec for
the AutoDecoder per object. 

We compare our P-VAE to the \emph{AutoDecoder} approach by encoding a  
given SDF patch $p_{gt}$ to a latent code $z = E_\mathrm{VAE}(p_{gt})$. 
We then add random noise to $z$ to get an initial version $z'$, which 
is then decoded into a SDF $p_{pred} = D_\mathrm{VAE}(z')$.
We compute a loss $|p_{gt} - p_{pred}|$ to update $z'$ in a loop till convergence.

\begin{figure}[ht]
    \centering
    \begin{tabular}{c@{}c@{}c@{~~}c@{}c@{~~}c@{}c}
         & \multicolumn{2}{c}{P-VAE} & \multicolumn{2}{c}{AutoDecoder} & \multicolumn{2}{c}{Ground Truth}
         \\
         & \includegraphics[width=0.11\linewidth,trim={2cm 2cm 2cm 2cm},clip]{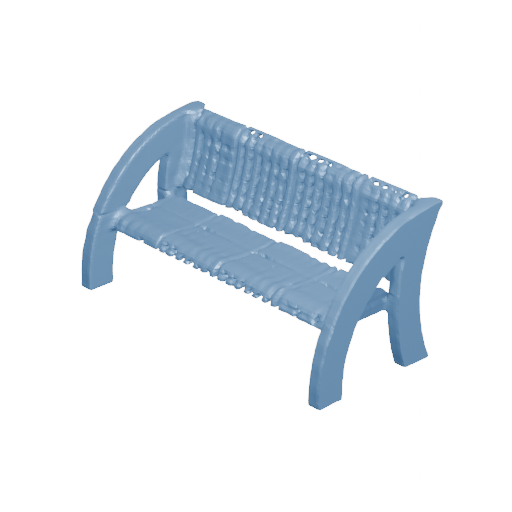}
         & \includegraphics[width=0.11\linewidth,trim={1.5cm 1.5cm 1.5cm 1.5cm},clip]{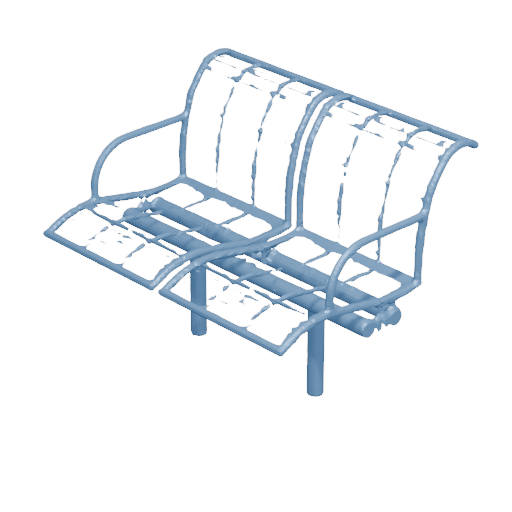}
         & \includegraphics[width=0.11\linewidth,trim={2cm 2cm 2cm 2cm},clip]{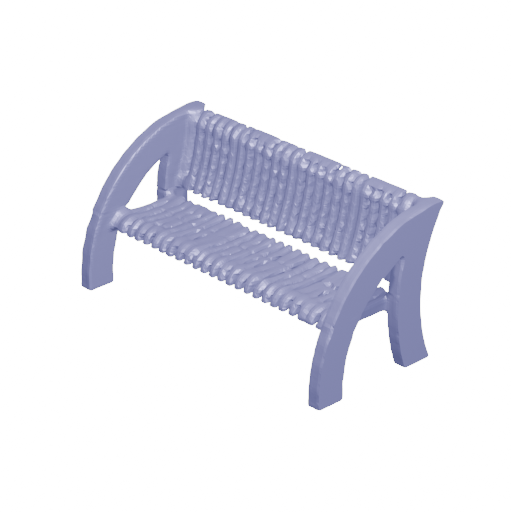}
         & \includegraphics[width=0.11\linewidth,trim={1.5cm 1.5cm 1.5cm 1.5cm},clip]{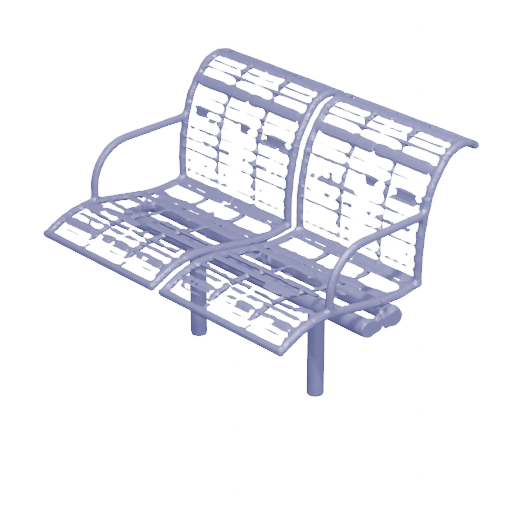}
         & \includegraphics[width=0.11\linewidth,trim={2cm 2cm 2cm 2cm},clip]{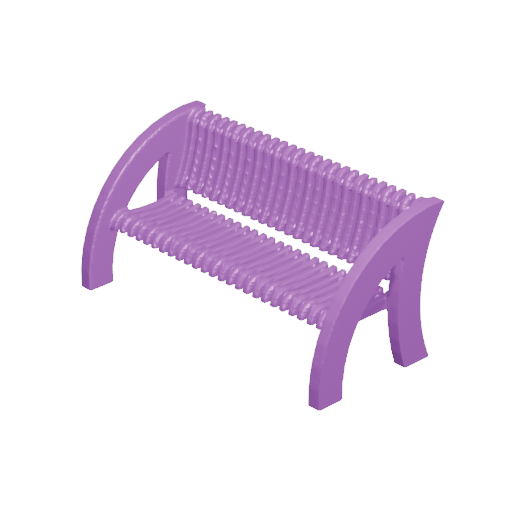}
         & \includegraphics[width=0.11\linewidth,trim={1.5cm 1.5cm 1.5cm 1.5cm},clip]{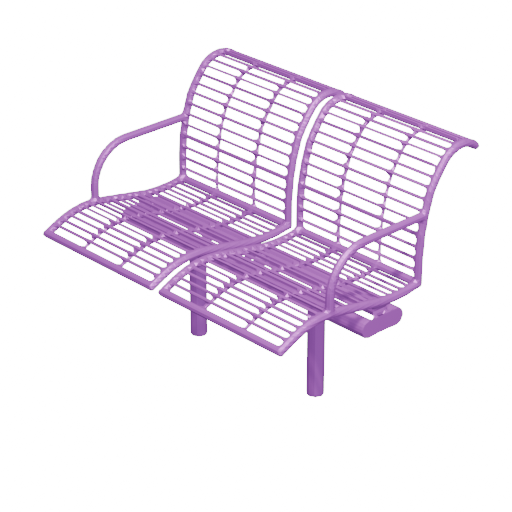}
         \\[-2ex]
         \small HD
         & \small  $0.01688$
         & \small  $0.05261$
         & \small  $0.01490$
         & \small  $0.02845$
         & \small -
         & \small -
         \\
         \small CD
         & \small  $0.01059$
         & \small  $0.14678$
         & \small  $0.00541$
         & \small  $0.01709$
         & \small -
         & \small -
         \\[-1ex]
    \end{tabular}
    \caption{Comparison of latent codes generated by P-VAE encoding 
    and refined via the AutoDecoder.
    While overall subtle, these extreme examples demonstrate the benefit of using the AutoDecoder to refine the patch embeddings.}
    \label{fig:eval:pvae}
\end{figure}

Both P-VAE and AD produce high-quality results with only minor differences to ground truth.
Across the dataset, we get a tiny improvement over the Hausdorff Distance from 0.026 to 0.024 from the optimization.
These improvements are mostly visible for objects with fine detail as shown in Figure~\ref{fig:eval:pvae}.
One could use these optimized latent codes $z'$ to train the SLT towards producing such output tokens at no extra cost.
We experimented on this idea and show results in Appendix~\ref{sec:appendix:slt_ad}.

%% file: sections/discussion.tex
\section{Limitations}\label{sec:limitations}

While the presented results are of high quality, some downstream tasks for the POC-SLT pipeline might need additional work. 
At the moment, it is trained assuming a fixed bounding box. This is a valid assumption for most objects but might be too limiting for open scenes. 
In principle, the transformer should be able to cope with sequences of basically arbitrary length. However, it still needs to be investigated if the current model can deal with higher-resolution spatial encodings to make use of additional tokens.

The completion is performed for masked patches. Implicitly, the approach always assumes either completely given SDF patches or completely unknown patches. 
In completion applications for images, depth maps, or partial 3D scans,
one might additionally need to indicate that the patch information itself might be incomplete, e.g.\ a missing occluded surface in the same cell.

We do not consider the single-view 3D reconstruction task.
Previous work \citep{What3D_Tatarchenko_2019_CVPR} has shown that predicting object-centered results heavily relies on
identifying (and augmenting) similar objects in the training set and can even be outperformed by object retrieval.
Lacking a single global object representation, our method is not well-suited for such a classification task.
Instead, one could first estimate a canonical pose, monocular depth, and then compute 3D SDF patches from the then recovered partial geometry
in order to meaningfully perform geometric 3D reconstruction with our method.
Evaluation of such an approach would highly depend on the models employed for pose estimation and depth estimation in addition to our model,
preventing meaningful comparisons.

%% file: sections/conclusion.tex
\section{Conclusion}

With POC-SLT, we proposed an accurate and efficient new method for SDF shape
completion in latent space. POC-SLT processes SDFs in patches of fixed size.
Two main components are used to refine and fill in missing patches in a shape.
Firstly, an extensively trained Patch Variational Autoencoder (P-VAE) for
accurately compressing the patches into a sequence of latent codes and back.
Secondly, an SDF-Latent-Transformer (SLT) which completes and refines the latent
sequence of an incomplete shape in a single inference step. We demonstrate that
our approach produces highly accurate and plausible 3D shape completions,
outperforming prior works. POC-SLT is trained on ShapeNetCoreV1~\citep{ShapeNet}, is class agnostic,
and can easily be adapted to new datasets, which we demonstrated with the ABC~\citep{ABC_Koch_2019_CVPR}
dataset. We show that the P-VAE works across different datasets, even without
additional training, and hope that it will be a helpful tool for future
research. Extending the approach to deal with partial information within patches
and developing applications like real-time scan completions are left open for
future work.

%% file: sections/acks.tex
This work has been supported by the Deutsche Forschungsgemeinschaft (DFG) – EXC number 2064/1 – Project number 390727645 and SFB 1233, TP 2, Project number 276693517, and by the IMPRS-IS.

%% file: sections/appendix.tex
\section{Additional Experiments} \label{sec:appendix:additional_eval}
In this section we show several additional experiments which did not fit into
the main paper. We evaluate training the SLT on auto-decoded latent codes
(Section~\ref{sec:appendix:slt_ad}), inspect how the SLT trained on
ShapeNet~\citep{ShapeNet} can generalize to the ABC dataset
(Section~\ref{appendix:slt_sn_on_abc}), show visual results of shape completion
on the completion tasks (R25), (R50) and (R75) with randomly masked inputs
(Section~\ref{appendix:random_masking}), compare our latent space representation
with DeepSDF~\citep{DeepSDF_Park_2019_CVPR} and
3DShape2VecSet~\citep{Zhang2023_3DShape2VecSet}
(Section~\ref{sec:appendix:latent_quality_comparison}), provide evaluation on
3D-EPN~\cite{TR1_Dai_2017_CVPR} (Section~\ref{sec:appendix:3d_epn_comparison}),
and, finally, present further comparisons with AnchorFormer~\citep{Chen2023AnchorFormer} in
Section~\ref{sec:appendix:comparison_anchorformer}.

\subsection{SDF-Latent-Transformer on Refined latent codes} \label{sec:appendix:slt_ad}
The comparison in Figure~\ref{fig:eval:pvae} suggests that the latent codes for
some objects might be improved by explicitly optimizing the latent codes of the
patches using the AutoDecoder technique~\citep{DeepSDF_Park_2019_CVPR,
AutoDecoder_osti_80034}. To test if the SLT might improve performance when
trained on regular partial latent codes $z$ as input but with optimized $z'$ as
ground truth, we optimized the latent codes for all objects in the training
dataset of ShapeNet~\citep{ShapeNet}. We trained a separate SDF Latent Transformer on this dataset called SLT-AD, expecting the SLT-AD to also learn to
do the expensive AutoDecoder optimization for free in its forward pass. We also
tested to run the regular SLT and, on the resulting SDF, the SLT-AD without
masking just for refinement. This model is called SLT + SLT-AD. Finally, we tried
running the SLT-AD twice: Once on the incomplete input and then again on the
completed output, which is called SLT-AD + SLT-AD.

\begin{figure}[htp]
    \centering
    \begin{tabular}{c@{}c@{}c@{}c@{}c@{}c@{}c@{}c}
        \rotatebox{90}{\parbox{0.11\linewidth}{\centering SN (Oct)}}
      & \includegraphics[width=0.11\linewidth]{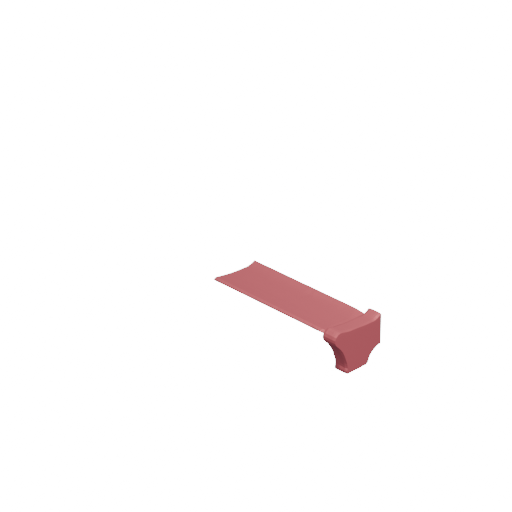}
      & \includegraphics[width=0.11\linewidth]{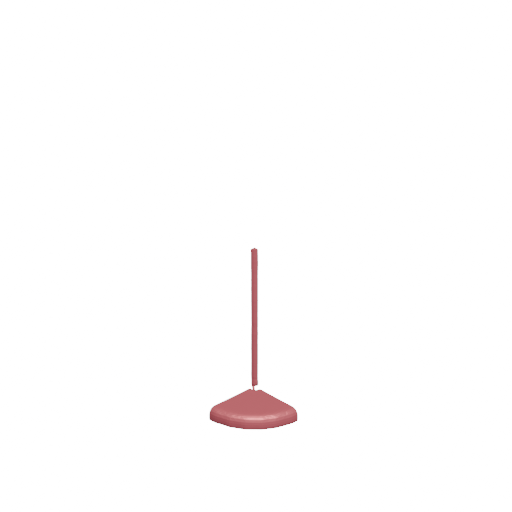}
      & \includegraphics[width=0.11\linewidth]{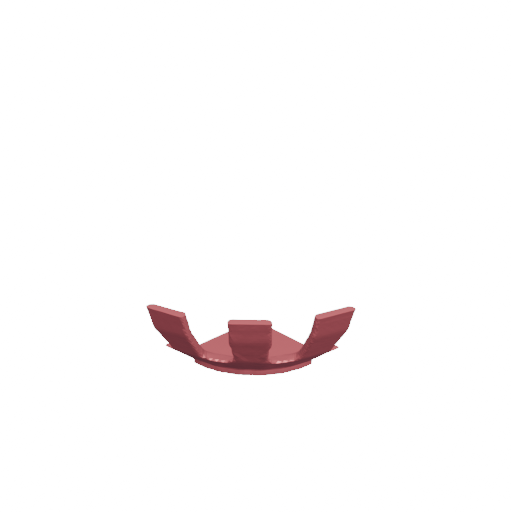}
      & \includegraphics[width=0.11\linewidth]{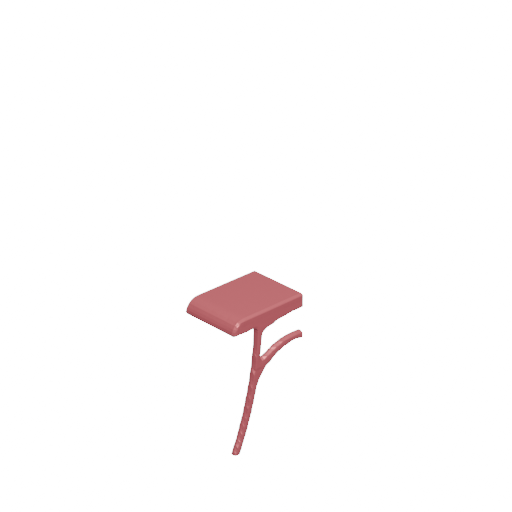}
      & \includegraphics[width=0.11\linewidth]{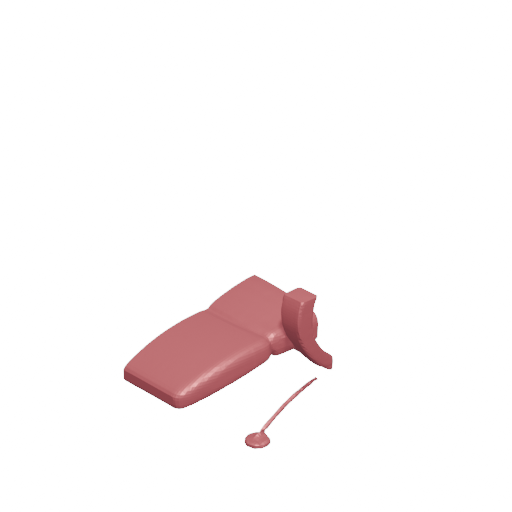}
      & \includegraphics[width=0.11\linewidth]{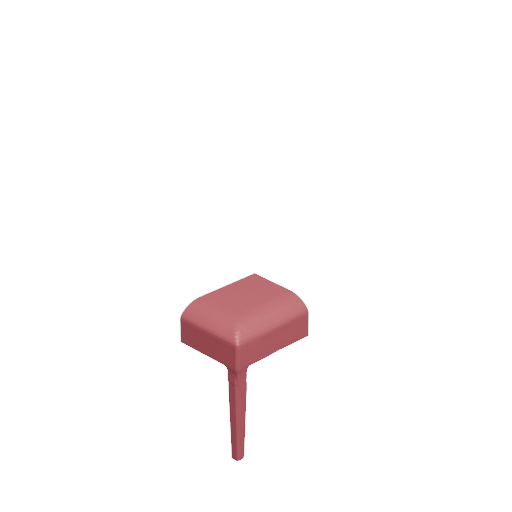}
      & \includegraphics[width=0.11\linewidth]{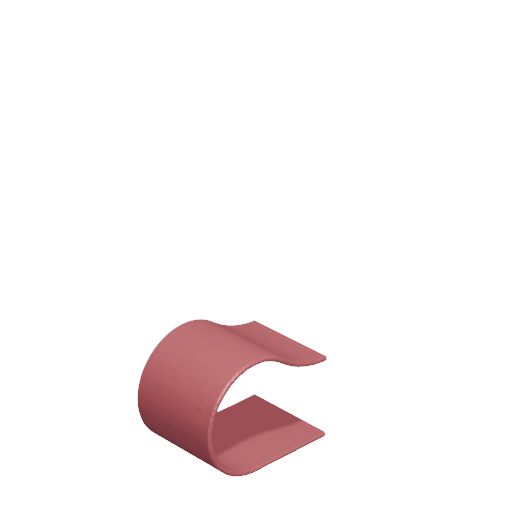}
      \\
        \rotatebox{90}{\parbox{0.11\linewidth}{\centering SLT}}
      & \includegraphics[width=0.11\linewidth]{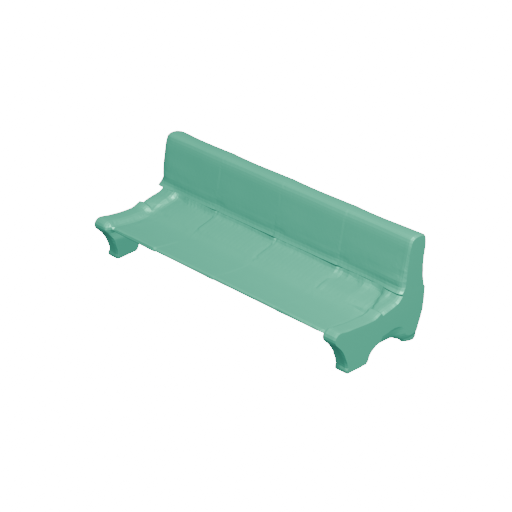}
      & \includegraphics[width=0.11\linewidth]{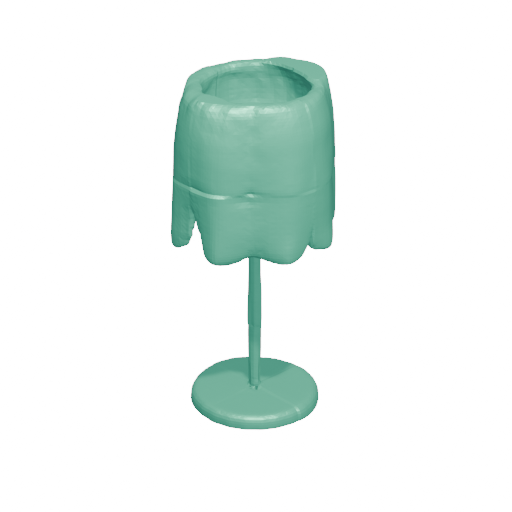}
      & \includegraphics[width=0.11\linewidth]{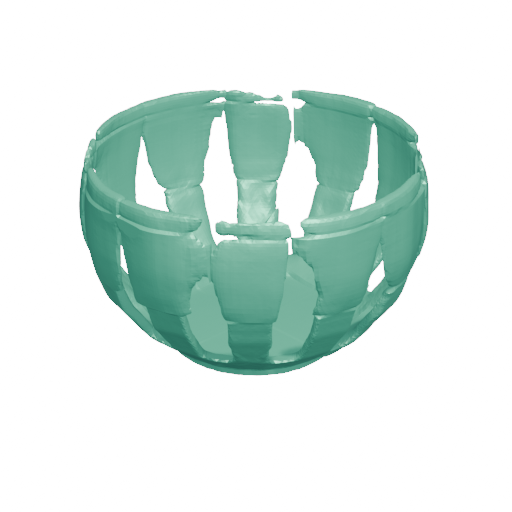}
      & \includegraphics[width=0.11\linewidth]{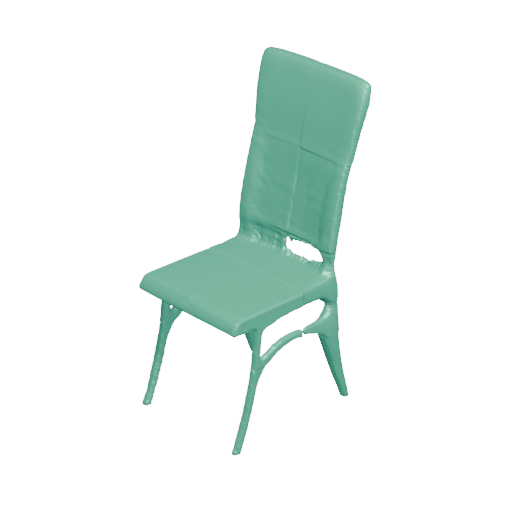}
      & \includegraphics[width=0.11\linewidth]{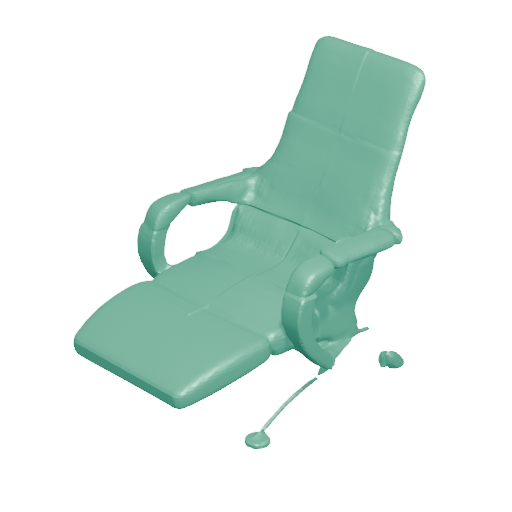}
      & \includegraphics[width=0.11\linewidth]{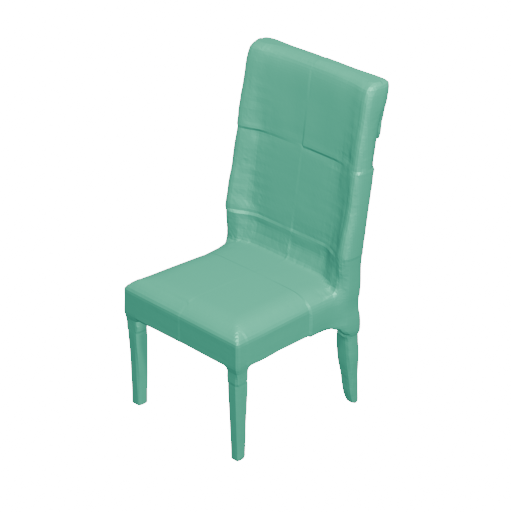}
      & \includegraphics[width=0.11\linewidth]{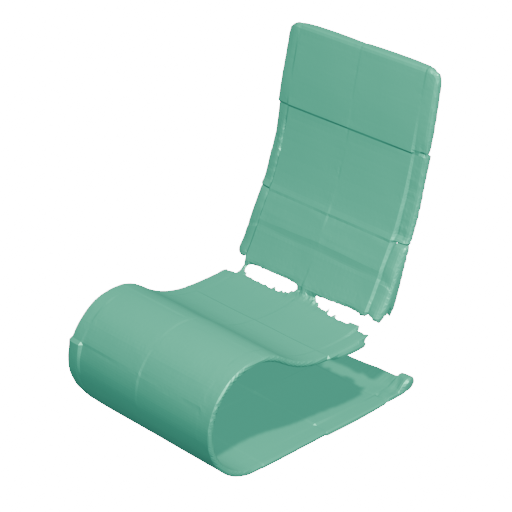}
        \\
        \rotatebox{90}{\parbox{0.11\linewidth}{\centering SLT + SLT-AD}}
      & \includegraphics[width=0.11\linewidth]{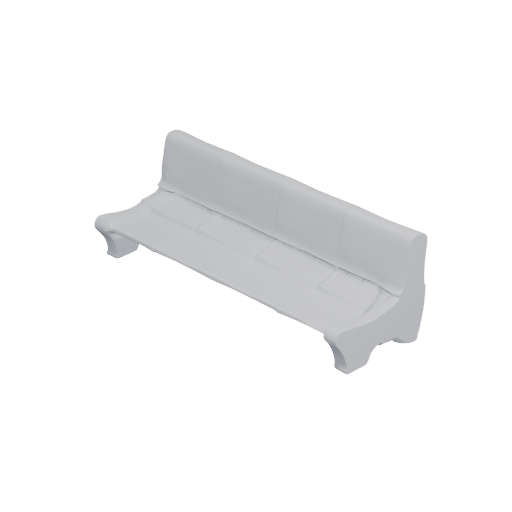}
      & \includegraphics[width=0.11\linewidth]{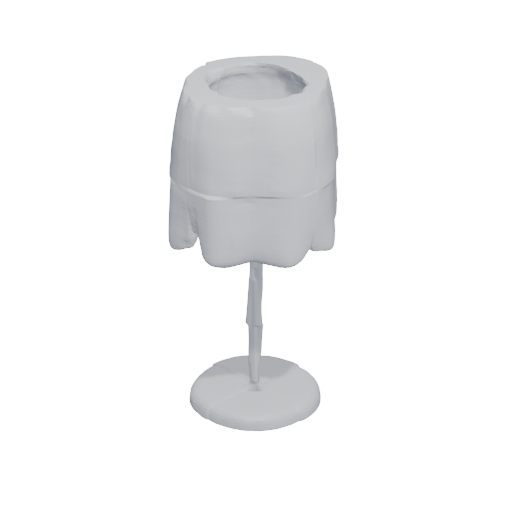}
      & \includegraphics[width=0.11\linewidth]{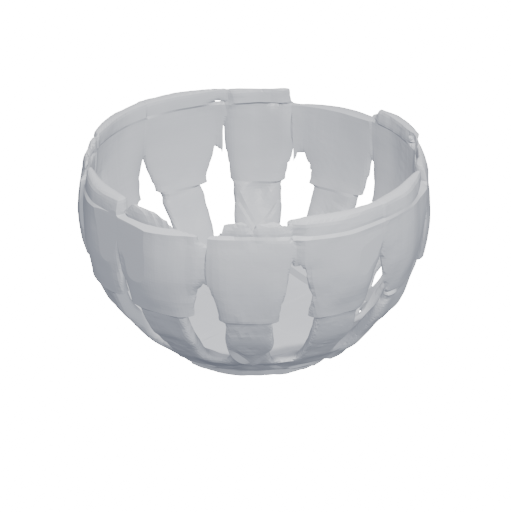}
      & \includegraphics[width=0.11\linewidth]{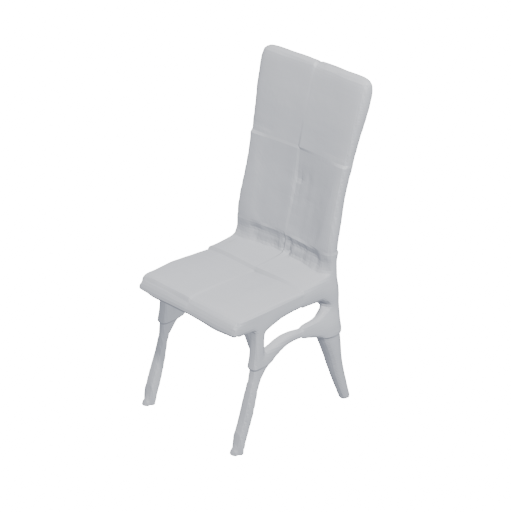}
      & \includegraphics[width=0.11\linewidth]{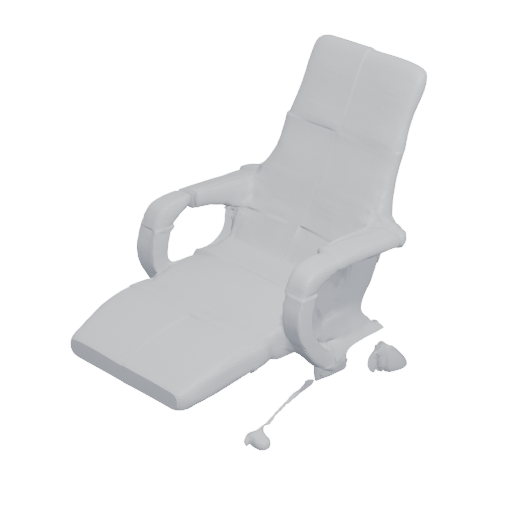}
      & \includegraphics[width=0.11\linewidth]{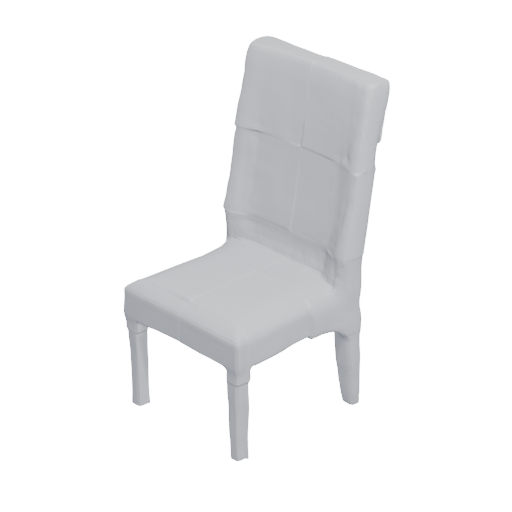}
      & \includegraphics[width=0.11\linewidth]{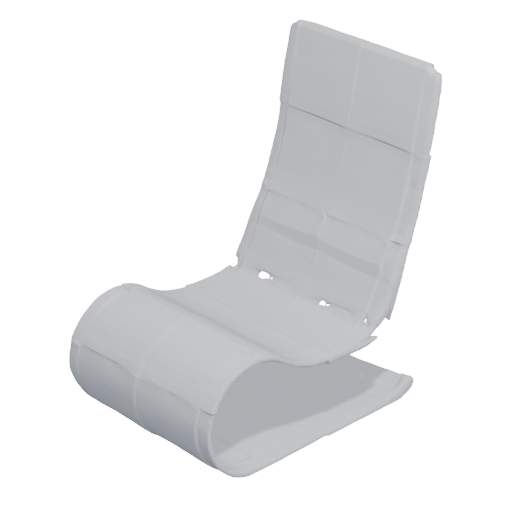}
      \\
        \rotatebox{90}{\parbox{0.11\linewidth}{\centering SN (Half)}}
      & \includegraphics[width=0.11\linewidth]{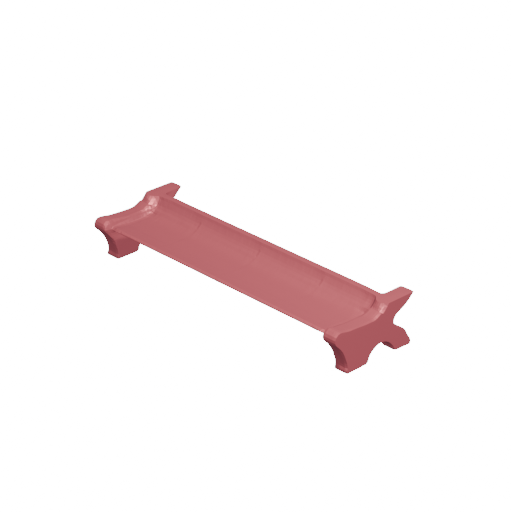}
      & \includegraphics[width=0.11\linewidth]{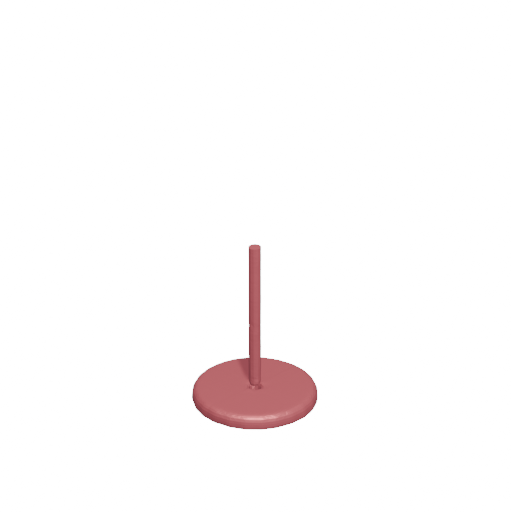}
      & \includegraphics[width=0.11\linewidth]{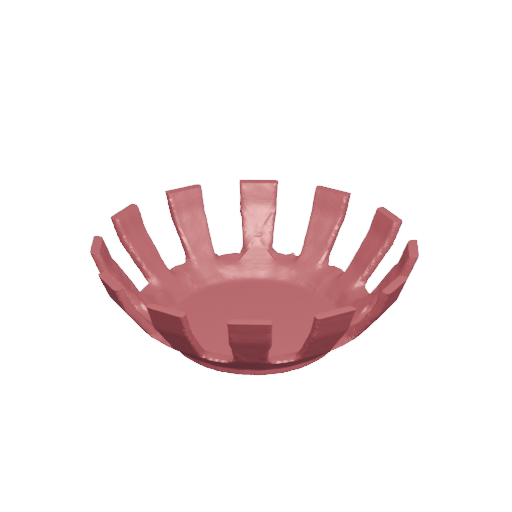}
      & \includegraphics[width=0.11\linewidth]{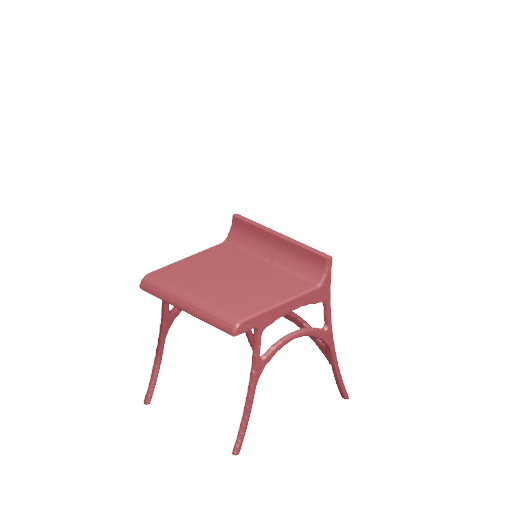}
      & \includegraphics[width=0.11\linewidth]{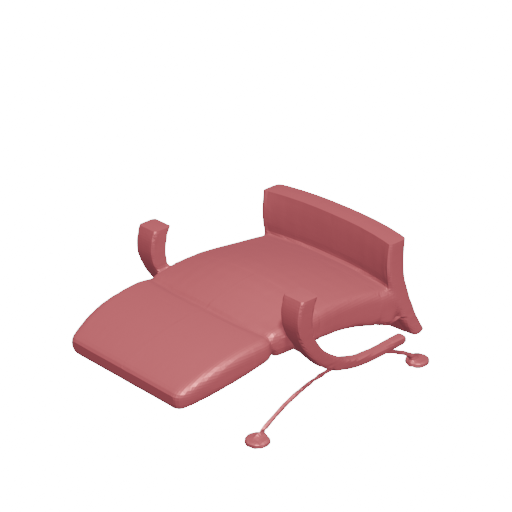}
      & \includegraphics[width=0.11\linewidth]{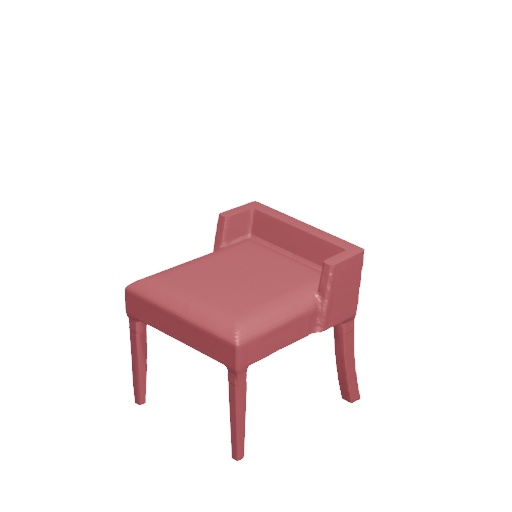}
      & \includegraphics[width=0.11\linewidth]{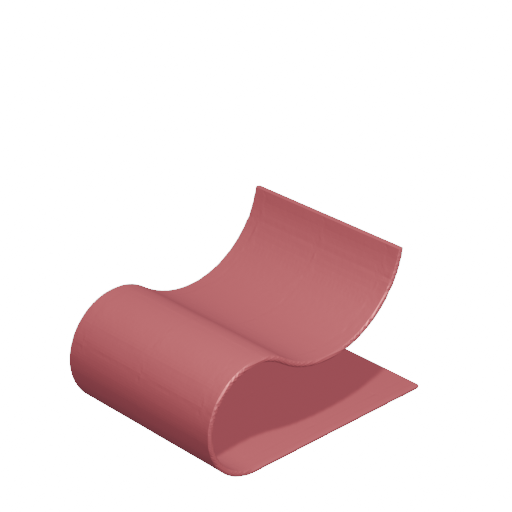}
      \\
        \rotatebox{90}{\parbox{0.11\linewidth}{\centering SLT}}
      & \includegraphics[width=0.11\linewidth]{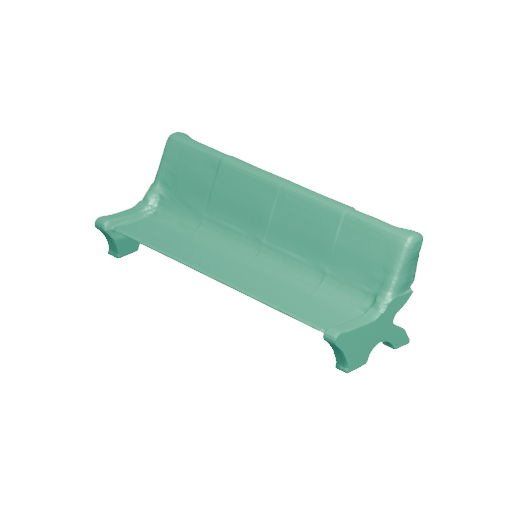}
      & \includegraphics[width=0.11\linewidth]{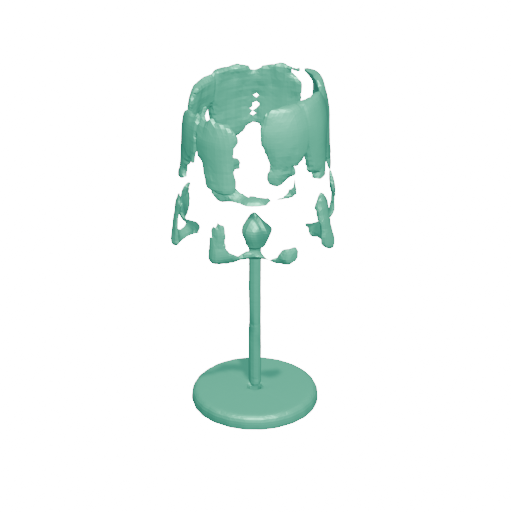}
      & \includegraphics[width=0.11\linewidth]{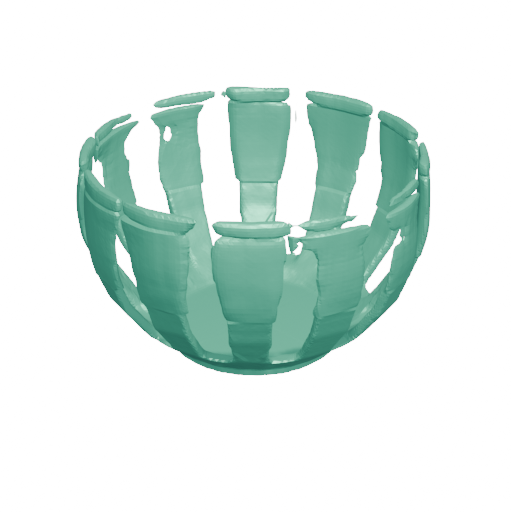}
      & \includegraphics[width=0.11\linewidth]{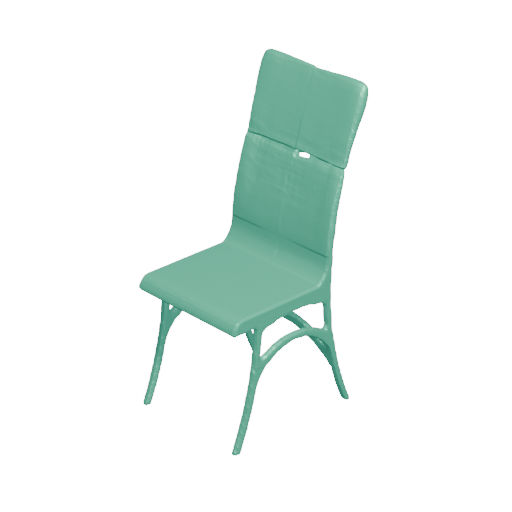}
      & \includegraphics[width=0.11\linewidth]{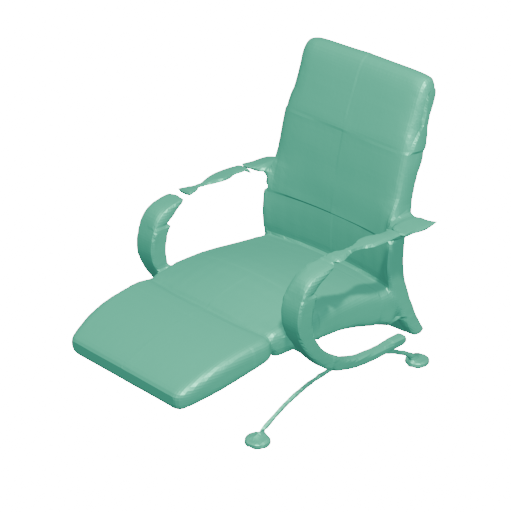}
      & \includegraphics[width=0.11\linewidth]{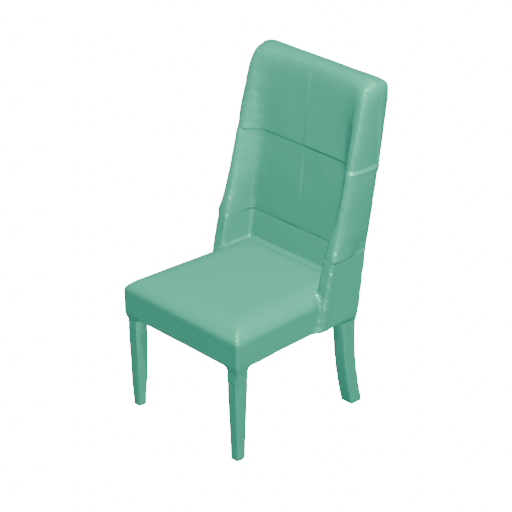}
      & \includegraphics[width=0.11\linewidth]{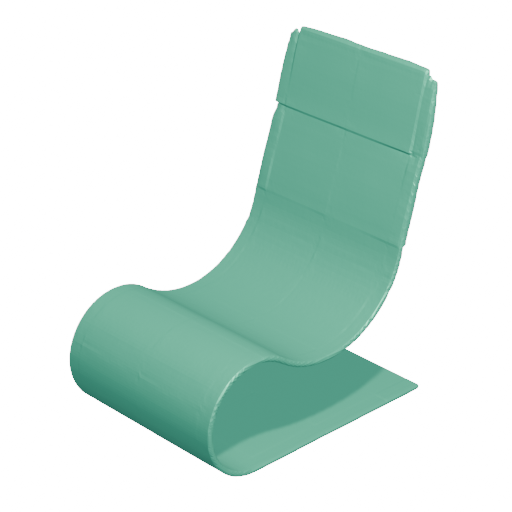}
        \\
        \rotatebox{90}{\parbox{0.11\linewidth}{\centering SLT + SLT-AD}}
      & \includegraphics[width=0.11\linewidth]{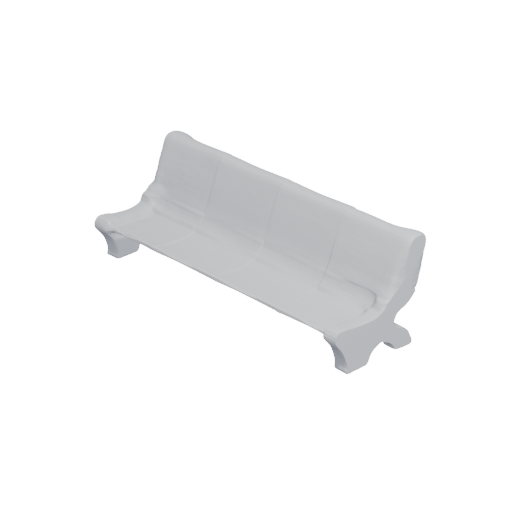}
      & \includegraphics[width=0.11\linewidth]{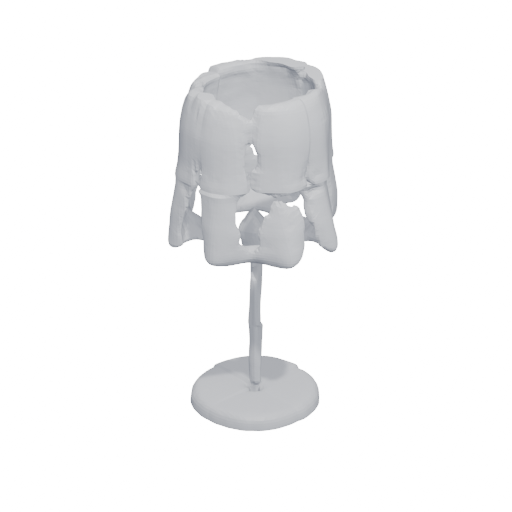}
      & \includegraphics[width=0.11\linewidth]{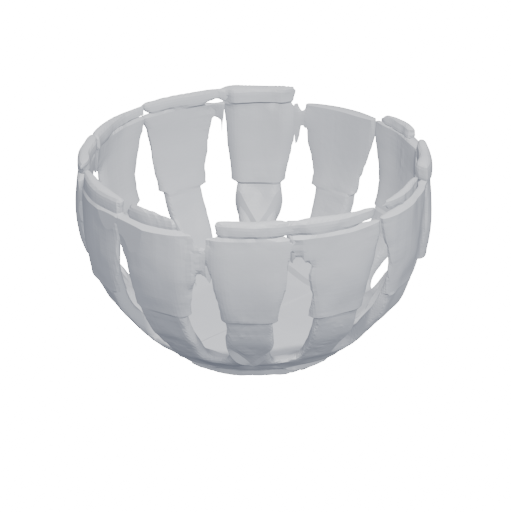}
      & \includegraphics[width=0.11\linewidth]{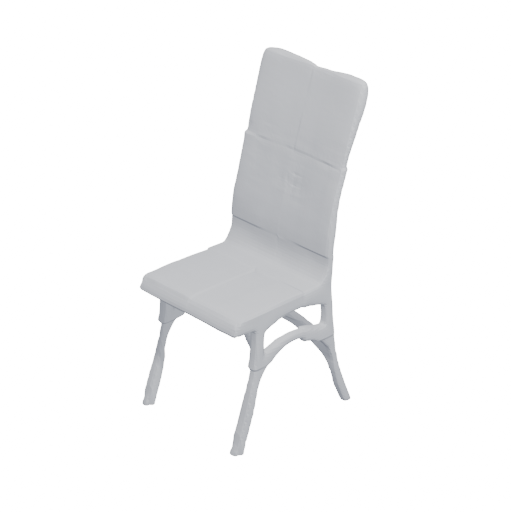}
      & \includegraphics[width=0.11\linewidth]{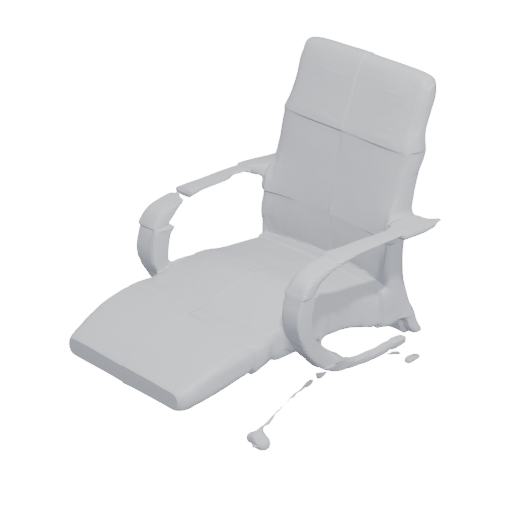}
      & \includegraphics[width=0.11\linewidth]{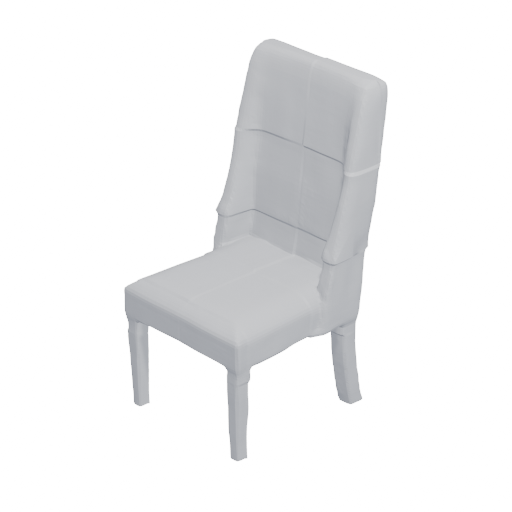}
      & \includegraphics[width=0.11\linewidth]{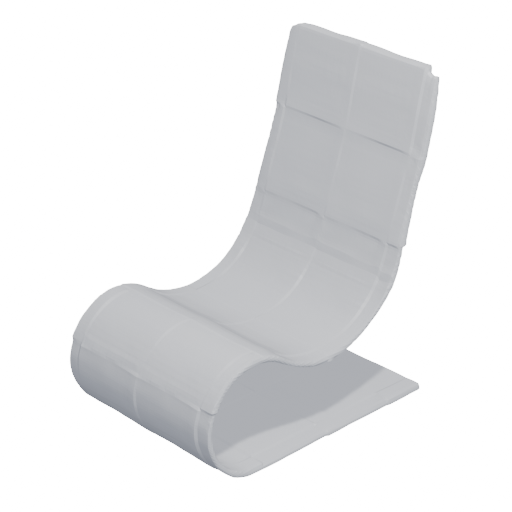}
      \\
        \rotatebox{90}{\parbox{0.11\linewidth}{\centering GT}}
      & \includegraphics[width=0.11\linewidth]{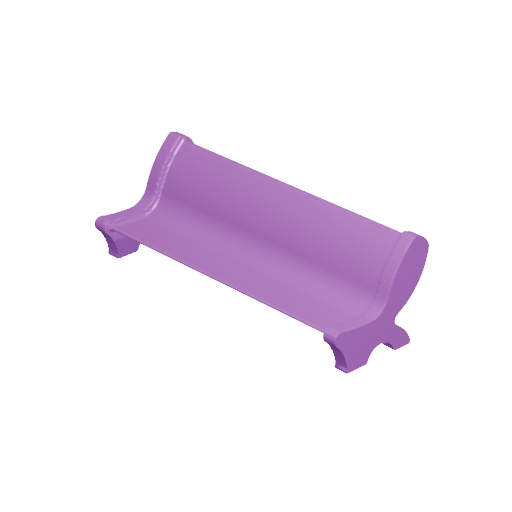}
      & \includegraphics[width=0.11\linewidth]{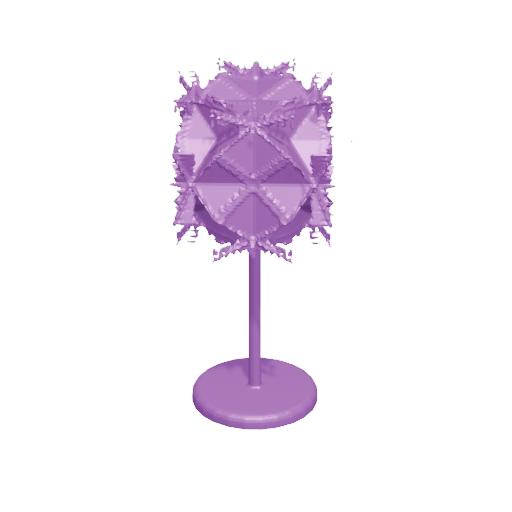}
      & \includegraphics[width=0.11\linewidth]{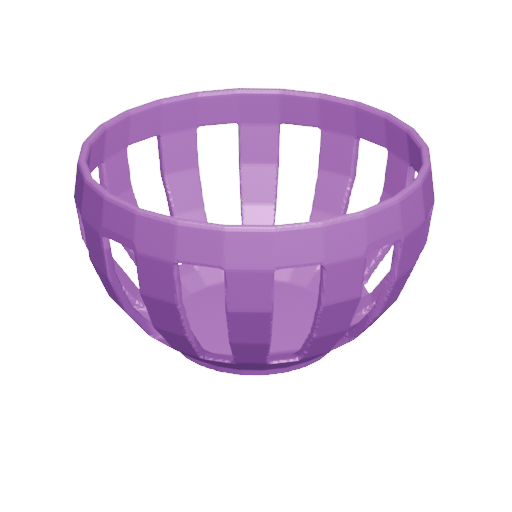}
      & \includegraphics[width=0.11\linewidth]{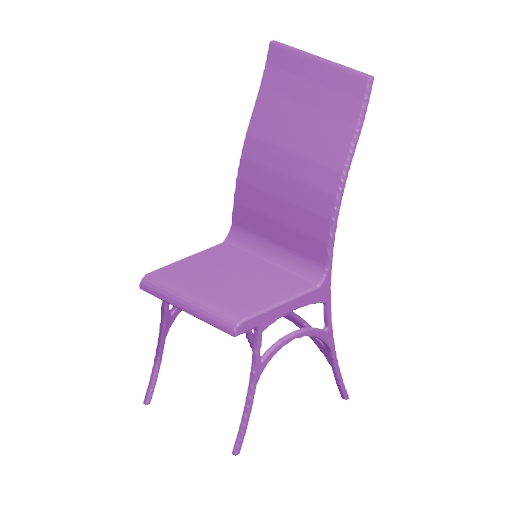}
      & \includegraphics[width=0.11\linewidth]{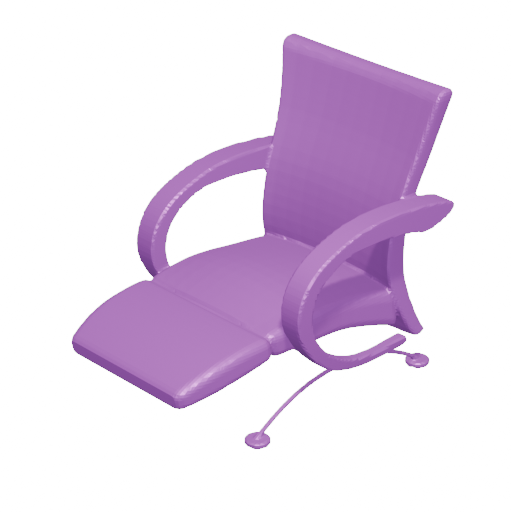}
      & \includegraphics[width=0.11\linewidth]{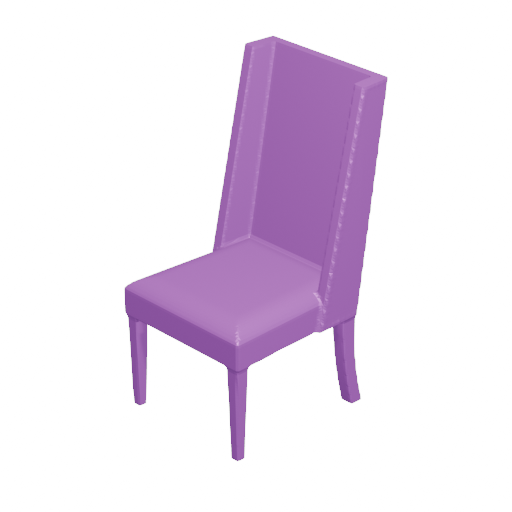}
      & \includegraphics[width=0.11\linewidth]{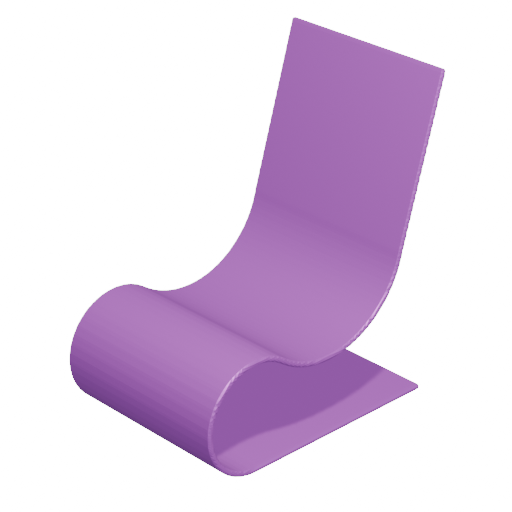}
     \\
    \end{tabular}
    \caption{Completion from halves and octants by our SLT and then refined by SLT-AD on ShapeNet.}
    \label{fig:eval:slt_refined_by_slt_ad}
\end{figure}

The quantitative results are shown in Table~\ref{tab:appendix:slt_ad} and qualitative results are shown in Figure~\ref{fig:eval:slt_refined_by_slt_ad}.
The numbers for the bottom half (Half) experiment clearly suggest, that the expected
improvement did not happen. The results for the octant (Oct) experiment are
inconclusive at best.

While the AutoDecoder technique~\citep{DeepSDF_Park_2019_CVPR, AutoDecoder_osti_80034}
produces latent codes that decode to a more accurate representation of the given input,
we believe that the latent codes generated this way are more likely to be outliers
and thus not as easily understood and utilized by the SLT-AD,
resulting in worse performance.

\begin{table*}[htp]
  \caption{Shape completion on ShapeNet~\citep{ShapeNet} with several versions of the SLT.
  SLT is the default version, SLT-AD hs been trained on auto-decoded ground-truth latent codes,
  and the "+ SLT-AD" variants feed the previous output through SLT-AD again without masking for refinement.}
  \label{tab:appendix:slt_ad}
  \centering
  \begin{tabular}{llccccccccccccccc}
    \toprule
    Model
    & Task 
    & IoU$\uparrow$ &$F_1$$\uparrow$ &CD$\downarrow$ &HD$\downarrow$  &NC$\uparrow$ &IN$\downarrow$& CMP$\uparrow$
    \\

    \midrule                     %
    SLT %
    & Half           & \textbf{0.7466} & \textbf{0.8468} & 1.0221          & \textbf{0.0765} & \textbf{0.9200} & \textbf{0.4196} & 0.9067 \\
    SLT-AD %
    & Half     & 0.7237          & 0.8344          & 1.2065          & 0.0846          & 0.9119          & 0.4767          & \textbf{0.9086} \\
    SLT + SLT-AD %
    & Half      & 0.6867          & 0.8101          & 1.0039          & 0.0811          & 0.8976          & 0.5268          & 0.8696 \\
    SLT-AD + SLT-AD %
    & Half   & 0.6780          & 0.7969          & \textbf{0.9111} & 0.0880          & 0.8923          & 0.5528          & 0.8658 \\
    
    \midrule                 %
    SLT %
    & Oct & 0.5884          & \textbf{0.7336} & \textbf{1.2467} & 0.0966          & 0.8589          & 0.6034          & \textbf{0.8404} \\
    SLT-AD %
    & Oct & \textbf{0.6127} & 0.7121          & 1.8994          & 0.1010          & \textbf{0.8694} & \textbf{0.5704} & 0.8279 \\
    SLT + SLT-AD %
    & Oct & 0.5878          & 0.6848          & 1.6883          & 0.0970          & 0.8557          & 0.6052          & 0.7729 \\
    SLT-AD + SLT-AD %
    & Oct & 0.5912          & 0.6795          & 1.4225          & \textbf{0.0940} & 0.8560          & 0.6167          & 0.7767 \\
    
    \bottomrule
  \end{tabular}
\end{table*}

\subsection{ABC completion using SLT trained on ShapeNet} \label{appendix:slt_sn_on_abc}
\begin{figure}[htb]
    \centering
    \vspace*{-1em} %
    \begin{tabular}{lc@{}c@{}c@{}c@{}c@{}c@{}c@{}c}
          \rotatebox{90}{\parbox{0.11\linewidth}{\centering ABC (Oct)}}
        & \includegraphics[width=0.11\linewidth]{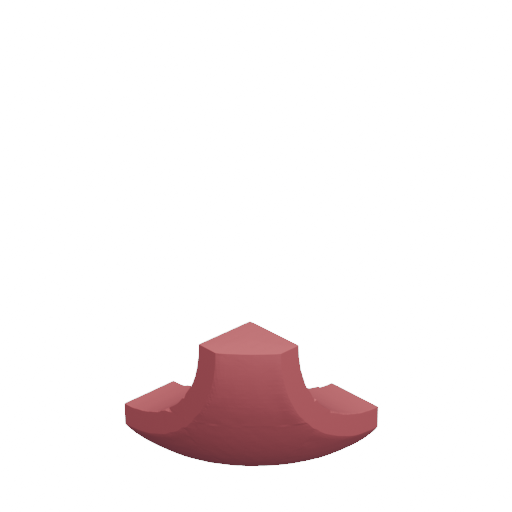}
        & \includegraphics[width=0.11\linewidth]{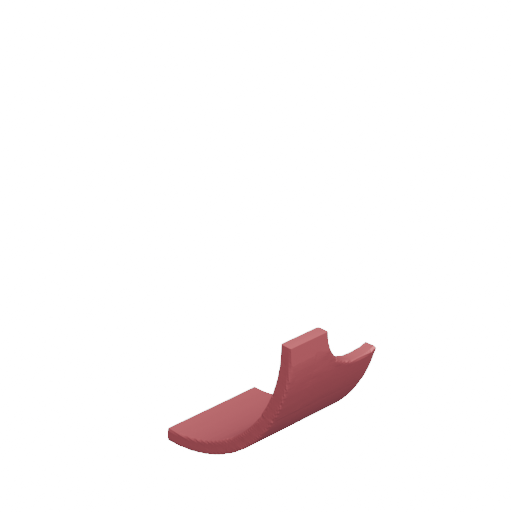}
        & \includegraphics[width=0.11\linewidth]{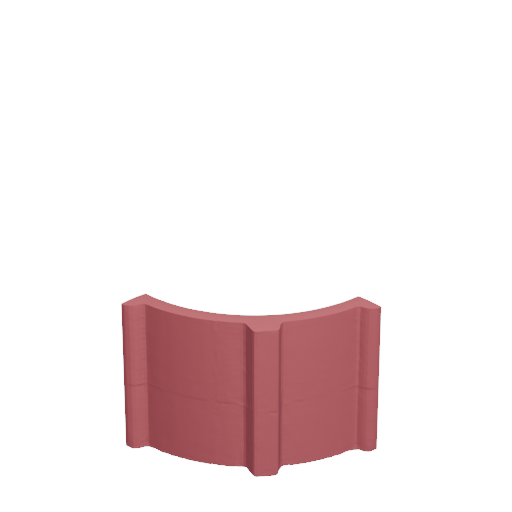}
        & \includegraphics[width=0.11\linewidth]{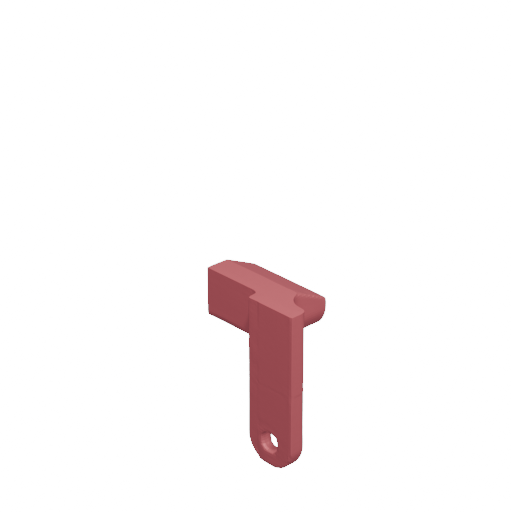}
        & \includegraphics[width=0.11\linewidth]{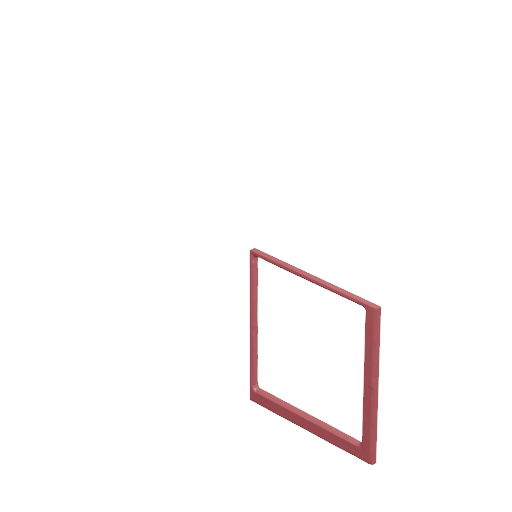}
        & \includegraphics[width=0.11\linewidth]{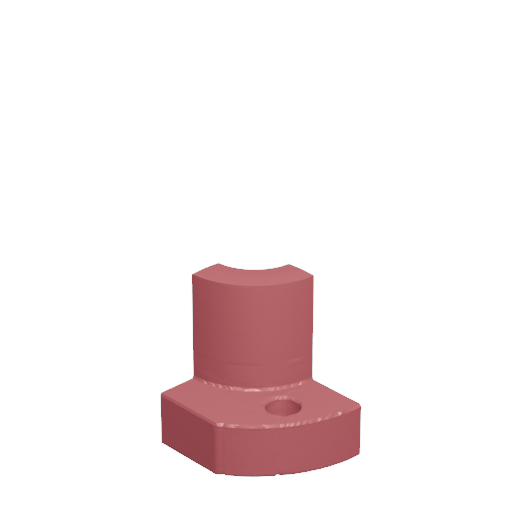}
        & \includegraphics[width=0.11\linewidth]{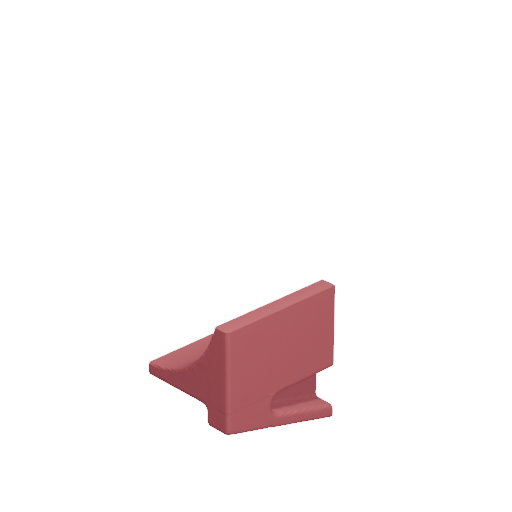}
        & \includegraphics[width=0.11\linewidth]{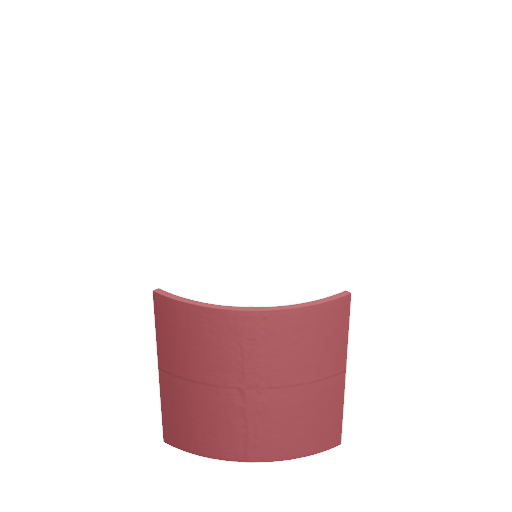}
        \\
          \rotatebox{90}{\parbox{0.11\linewidth}{\centering SLT }}
        & \includegraphics[width=0.11\linewidth]{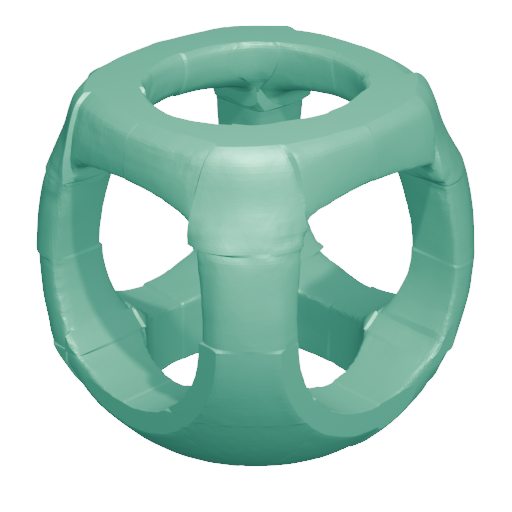}
        & \includegraphics[width=0.11\linewidth]{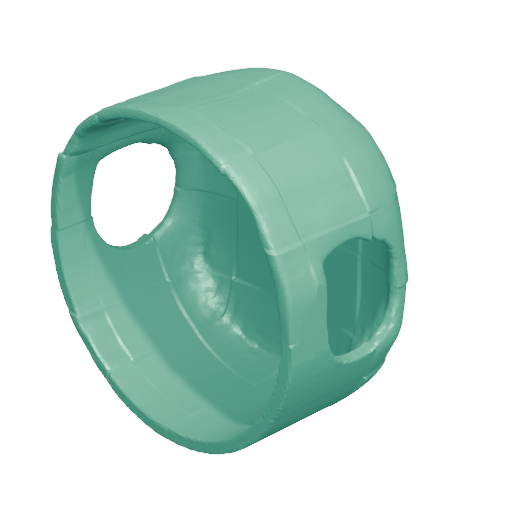}
        & \includegraphics[width=0.11\linewidth]{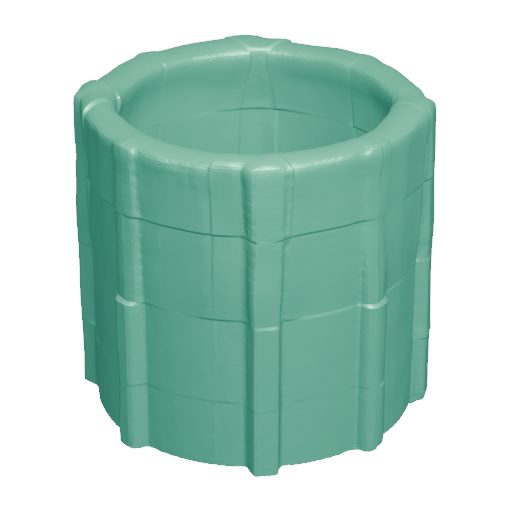}
        & \includegraphics[width=0.11\linewidth]{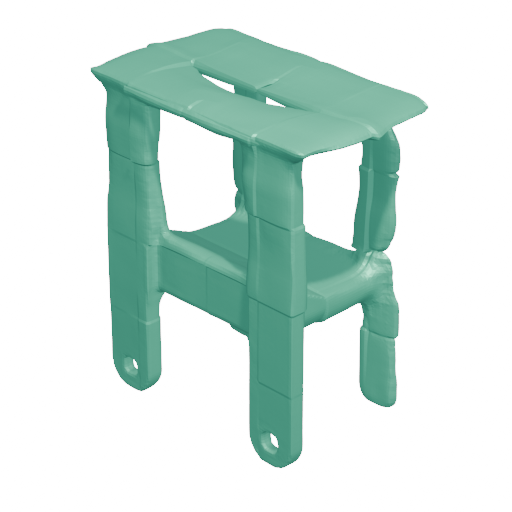}
        & \includegraphics[width=0.11\linewidth]{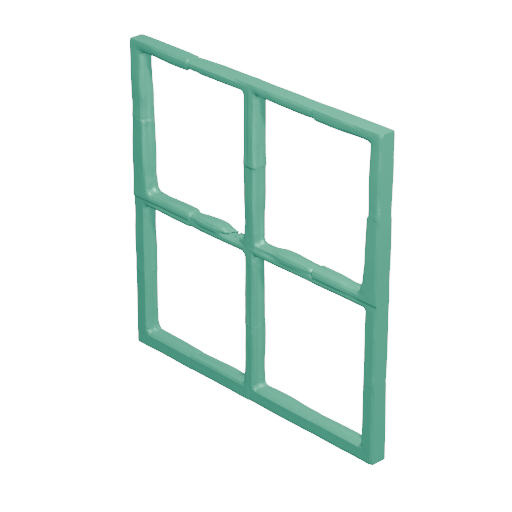}
        & \includegraphics[width=0.11\linewidth]{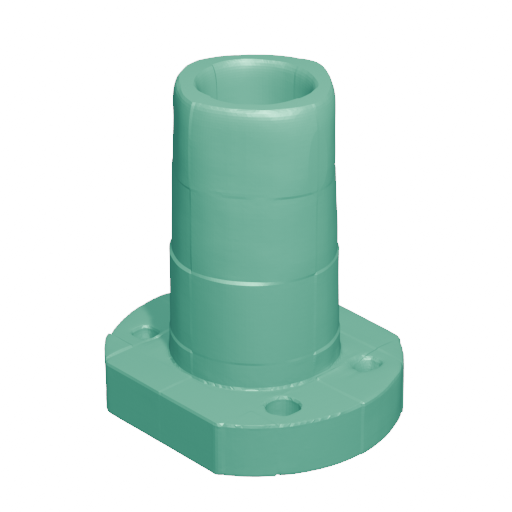}
        & \includegraphics[width=0.11\linewidth]{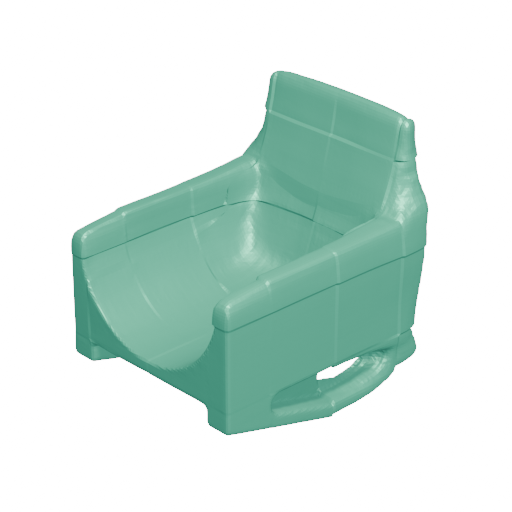}
        & \includegraphics[width=0.11\linewidth]{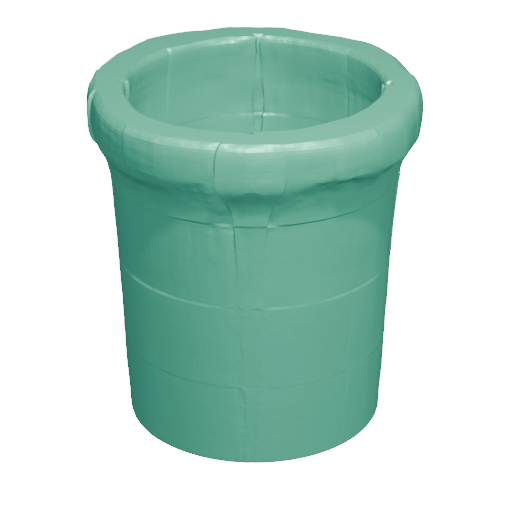}
        \\
          \rotatebox{90}{\parbox{0.11\linewidth}{\centering ABC (Half)}}
        & \includegraphics[width=0.11\linewidth]{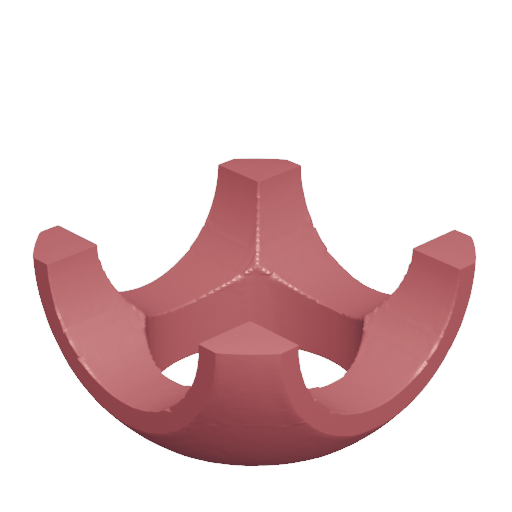}
        & \includegraphics[width=0.11\linewidth]{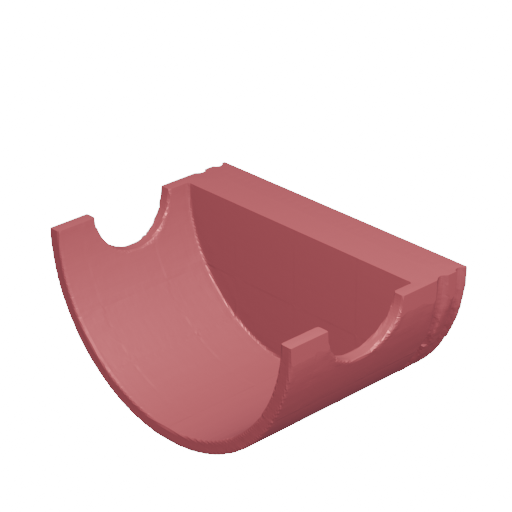}
        & \includegraphics[width=0.11\linewidth]{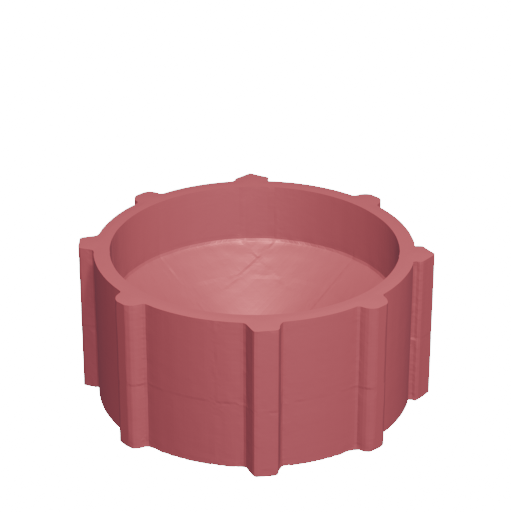}
        & \includegraphics[width=0.11\linewidth]{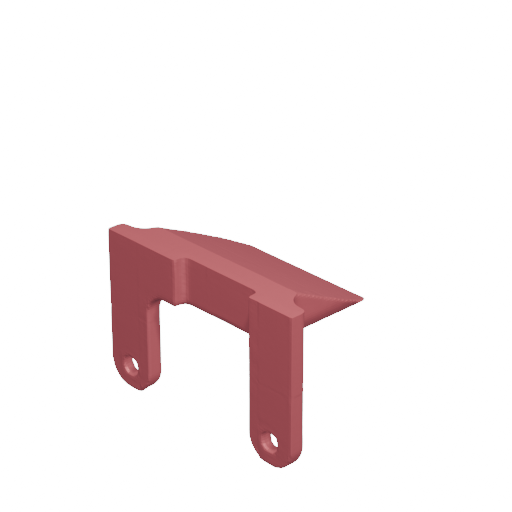}
        & \includegraphics[width=0.11\linewidth]{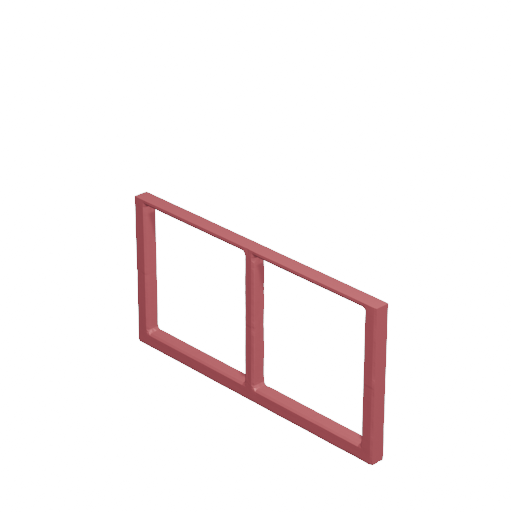}
        & \includegraphics[width=0.11\linewidth]{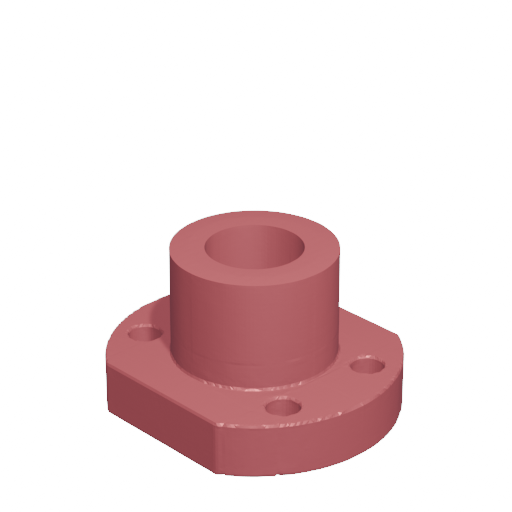}
        & \includegraphics[width=0.11\linewidth]{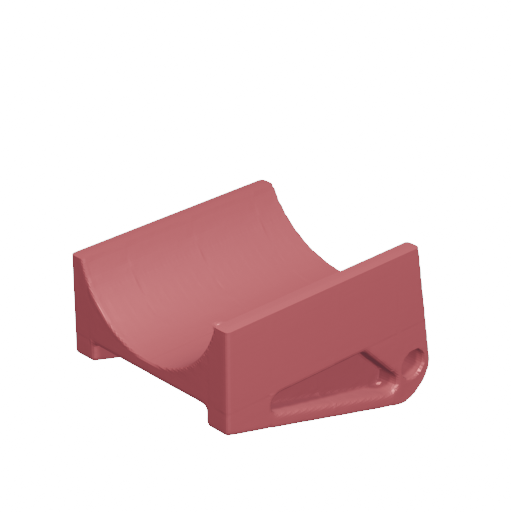}
        & \includegraphics[width=0.11\linewidth]{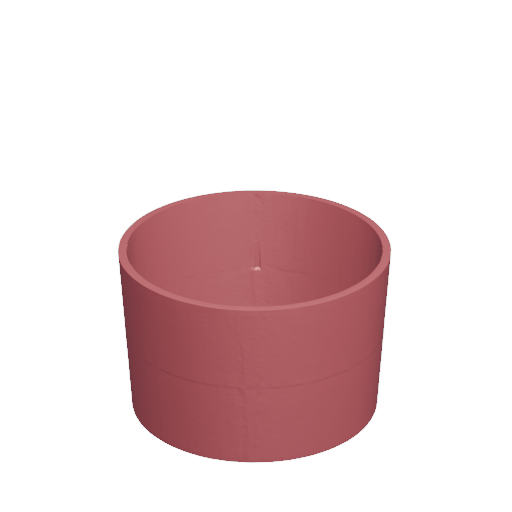}
        \\
          \rotatebox{90}{\parbox{0.11\linewidth}{\centering SLT}}
        & \includegraphics[width=0.11\linewidth]{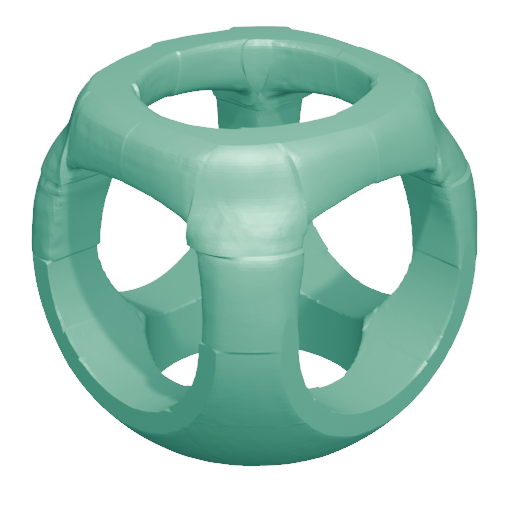}
        & \includegraphics[width=0.11\linewidth]{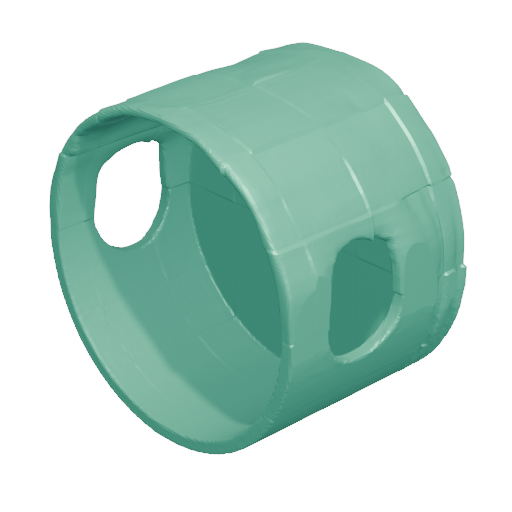}
        & \includegraphics[width=0.11\linewidth]{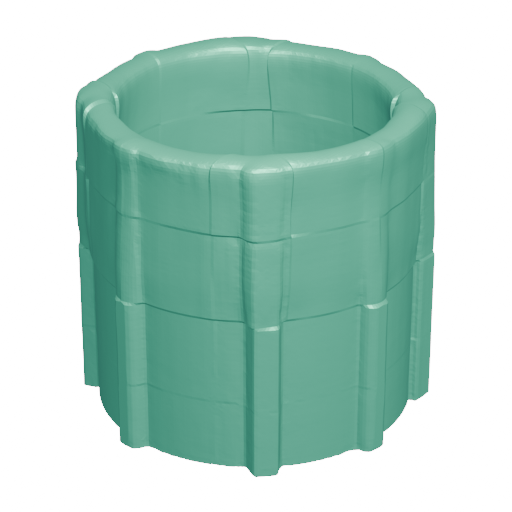}
        & \includegraphics[width=0.11\linewidth]{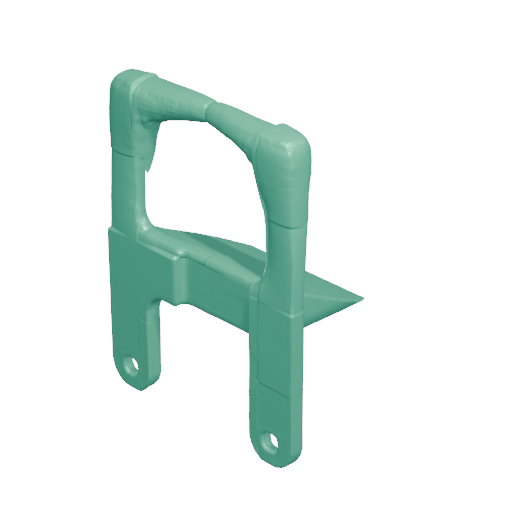}
        & \includegraphics[width=0.11\linewidth]{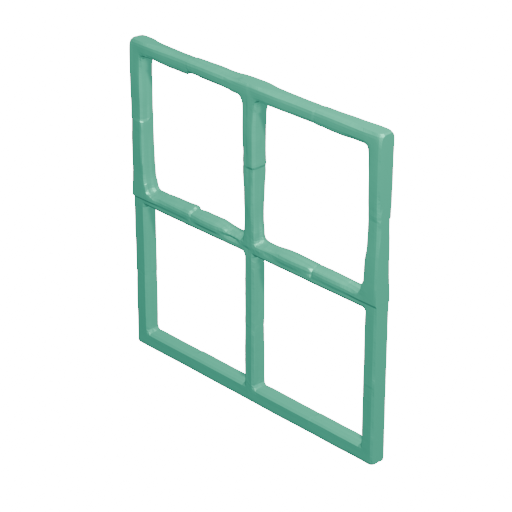}
        & \includegraphics[width=0.11\linewidth]{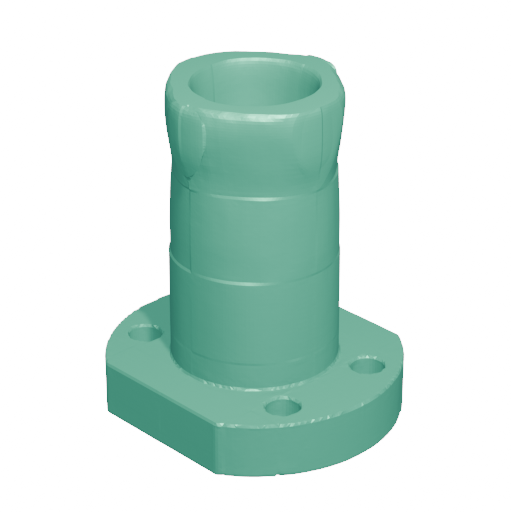}
        & \includegraphics[width=0.11\linewidth]{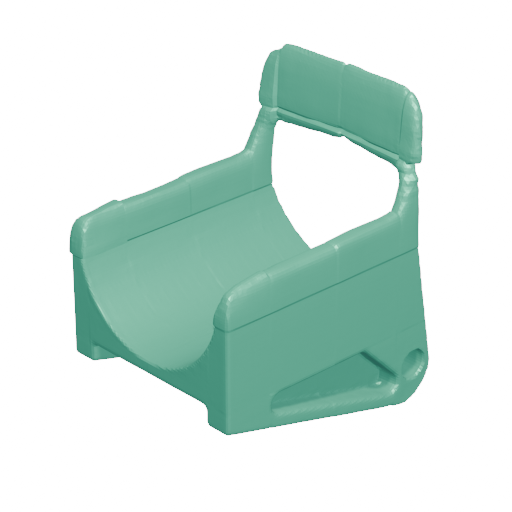}
        & \includegraphics[width=0.11\linewidth]{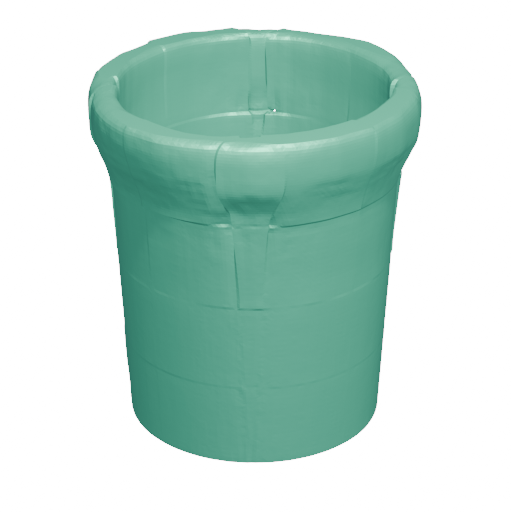}
        \\
          \rotatebox{90}{\parbox{0.11\linewidth}{\centering GT}}
        & \includegraphics[width=0.11\linewidth]{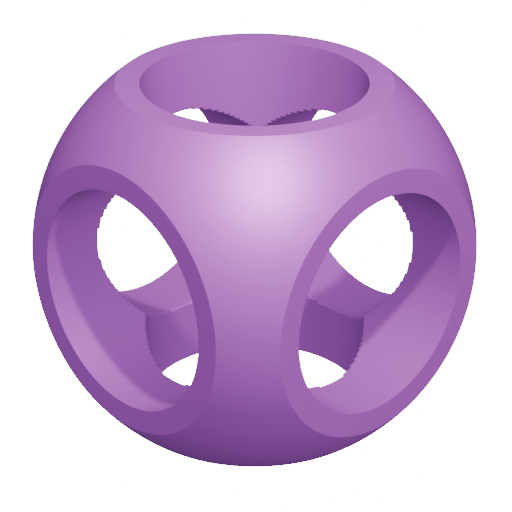}
        & \includegraphics[width=0.11\linewidth]{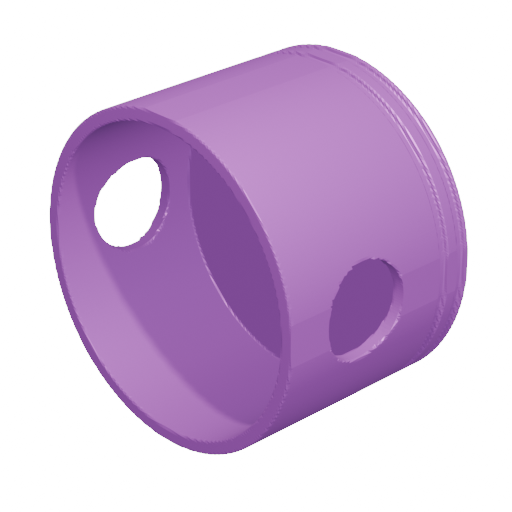}
        & \includegraphics[width=0.11\linewidth]{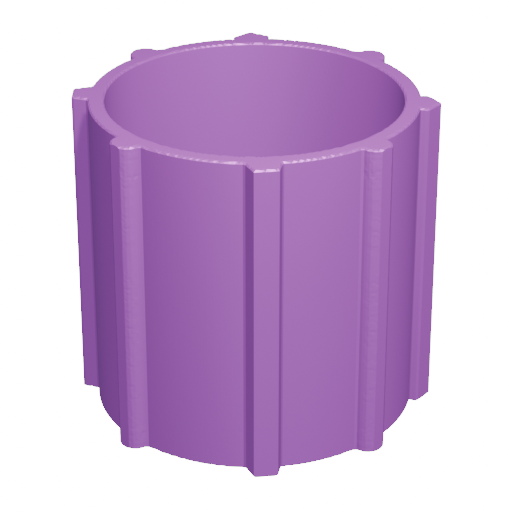}
        & \includegraphics[width=0.11\linewidth]{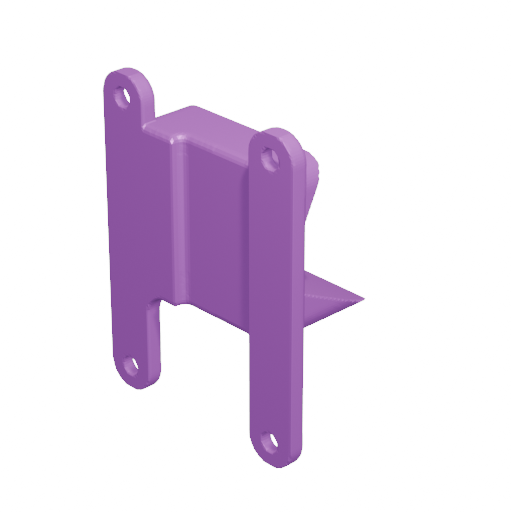}
        & \includegraphics[width=0.11\linewidth]{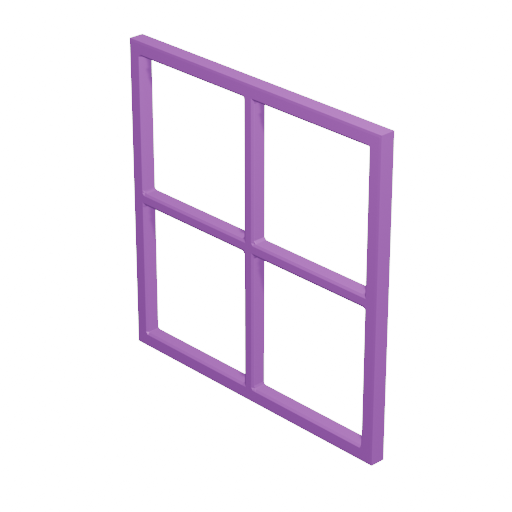}
        & \includegraphics[width=0.11\linewidth]{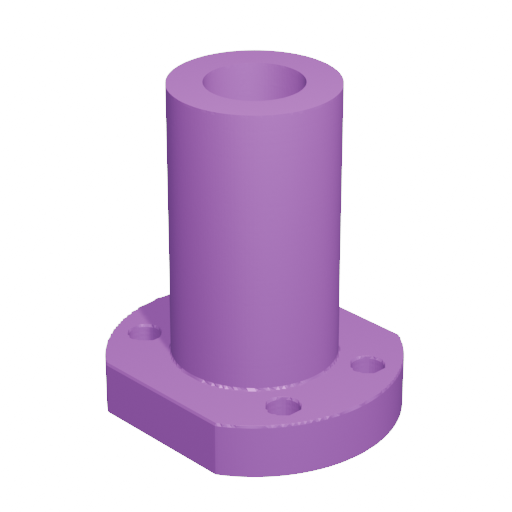}
        & \includegraphics[width=0.11\linewidth]{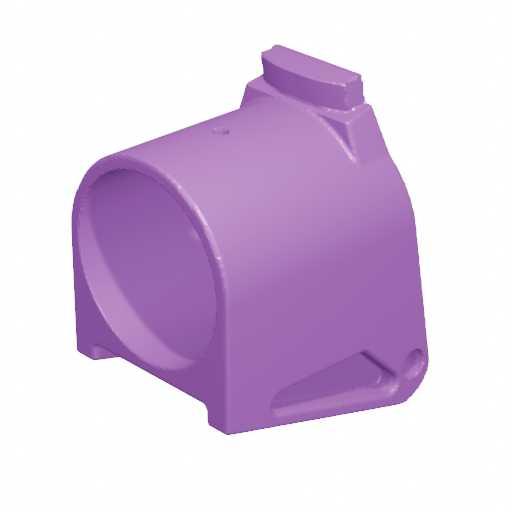}
        & \includegraphics[width=0.11\linewidth]{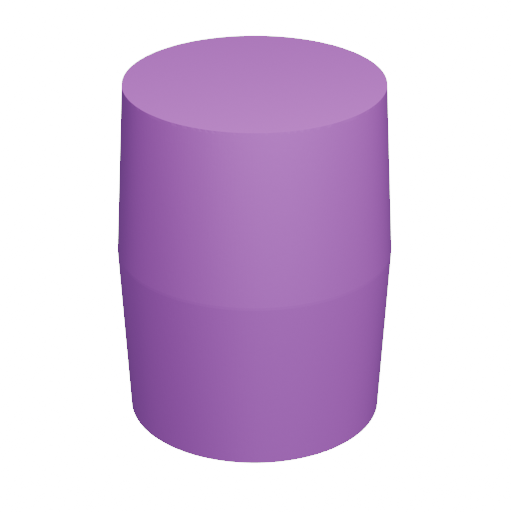}
        \\
    \end{tabular}
    \caption{Testing generalizability via completion of ABC~\citep{ABC_Koch_2019_CVPR} objects
    using the SLT trained exclusively on ShapeNet~\citep{ShapeNet}.}
    \label{fig:appendix:slt_sn_on_abc}
\end{figure}

To test the generalizability of our SLT trained on ShapeNet~\citep{ShapeNet},
we evaluate it on the ABC~\citep{ABC_Koch_2019_CVPR} dataset without fine-tuning.
We report the metrics for the tasks (Half) and (Oct) in Table~\ref{tab:appendix:eval_ABC_through_Shapenet}.
To compare to the fine-tuned SLT ABC, we duplicate its numbers
reported in Table~\ref{tab:eval:completion} here for convenience.
As expected, the reconstruction metrics of the fine-tuned model are much better. 
Figure~\ref{fig:appendix:slt_sn_on_abc} shows, however, that the STL still does a
reasonably good job at completing many of the objects from ABC~\citep{ABC_Koch_2019_CVPR}.
Instead of producing complete mechanical parts,
it often creates chairs, tables, and vases from their partial inputs
and it picks up surprisingly well on some of the inherent symmetries.
With the (Oct) input, the SLT has more freedom to complete the input into objects found in ShapeNet~\citep{ShapeNet}, such as chairs, e.g., column four.

\begin{table*}[htb]
  \caption{
  Evaluation on completion tasks (Half), (Oct) on the
  ABC~\citep{ABC_Koch_2019_CVPR} dataset using SLT and comparing to the SLT ABC results copied from Table~\ref{tab:eval:completion}.
  The results report the mean over all categories.
  Details on the metrics can be found in Appendix~\ref{appendix:metrics}.}
  \label{tab:appendix:eval_ABC_through_Shapenet}
  \centering
  \begin{tabular}{ccccccccccccccc}
    \toprule
    Model
    & Dataset
    & Task 
    & IoU$\uparrow$ &$F_1$$\uparrow$ &CD$\downarrow$ &HD$\downarrow$  &NC$\uparrow$ &IN$\downarrow$& CMP$\uparrow$
    \\
    
    \midrule               %
    SLT       & ABC & Half & 0.7478 & 0.8123 & 2.0658 & 0.1053 & 0.9134 & 0.3730 & 0.8869 \\
    SLT ABC   & ABC & Half & \textbf{0.8617} & \textbf{0.9159} & \textbf{0.8703} & \textbf{0.0575} & \textbf{0.9435} & \textbf{0.2551} & \textbf{0.9466} \\
    \midrule                 %
    SLT       & ABC & Oct  & 0.5710 & 0.6304 & 5.1872 & 0.1532 & 0.8208 & 0.5433 & 0.7234 \\
    SLT ABC   & ABC & Oct  & \textbf{0.7144} & \textbf{0.7744} & \textbf{2.9247} & \textbf{0.1077} & \textbf{0.8779} & \textbf{0.3986} & \textbf{0.8391} \\
    \bottomrule
  \end{tabular}
\end{table*}

\subsection{Completion on Randomly Masked ShapeNet} \label{appendix:random_masking}
\begin{figure}[htp]
    \centering
    \begin{tabular}{cccccccccccccccc}
          SN (R25)
        & SLT
        & SN (R50)
        & SLT
        & SN (R75)
        & SLT
        & GT
        \\
        \includegraphics[width=0.11\linewidth]{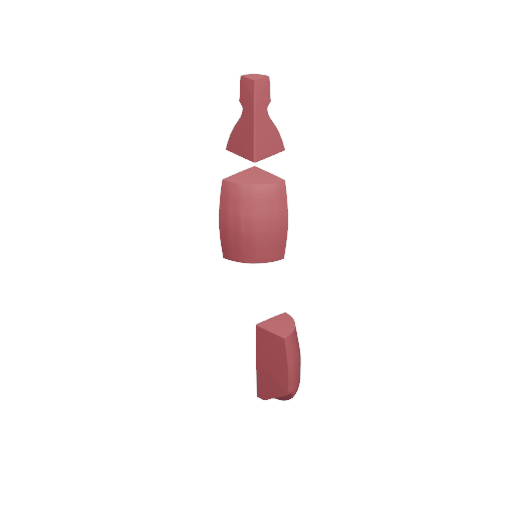}
        & \includegraphics[width=0.11\linewidth]{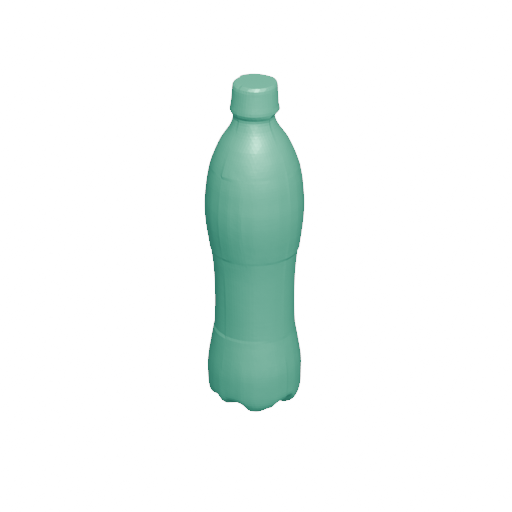}
        & \includegraphics[width=0.11\linewidth]{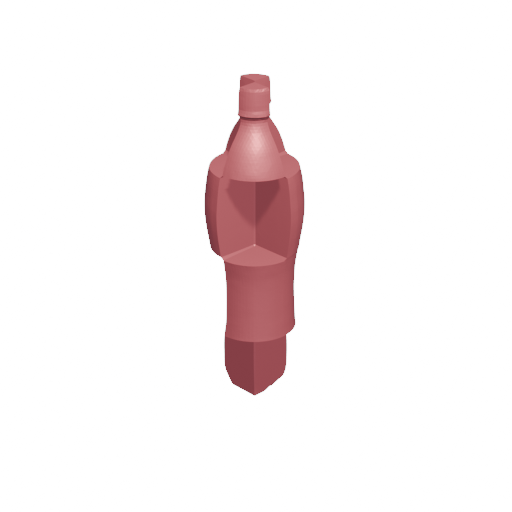}
        & \includegraphics[width=0.11\linewidth]{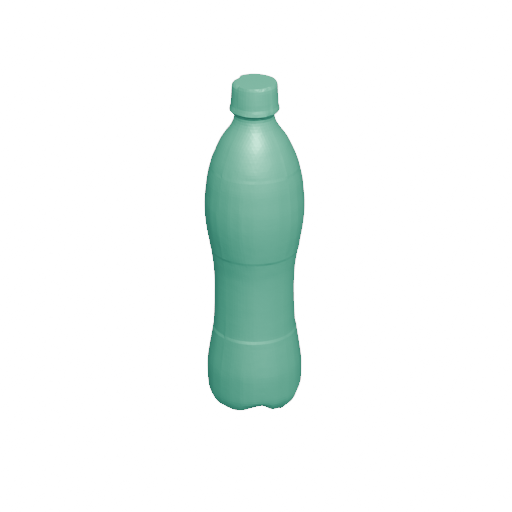}
        & \includegraphics[width=0.11\linewidth]{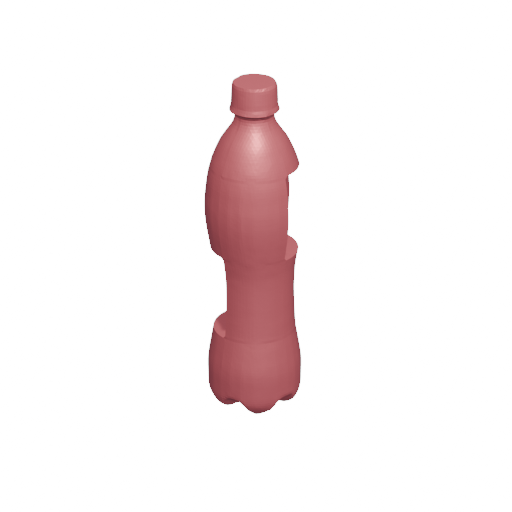}
        & \includegraphics[width=0.11\linewidth]{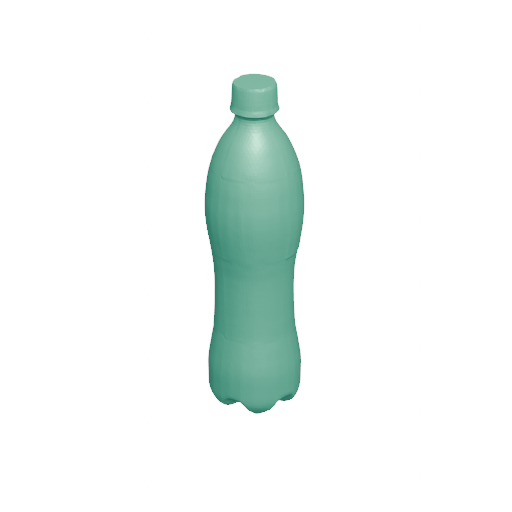}
        & \includegraphics[width=0.11\linewidth]{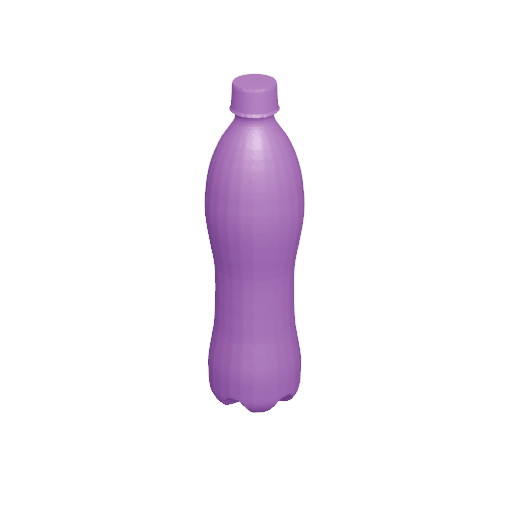}
        \\
        \includegraphics[width=0.11\linewidth]{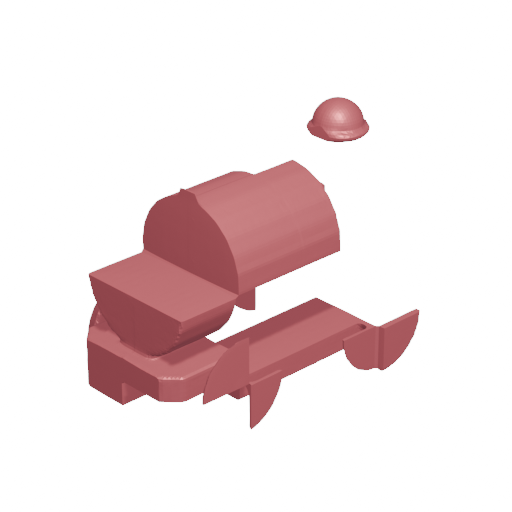}
        & \includegraphics[width=0.11\linewidth]{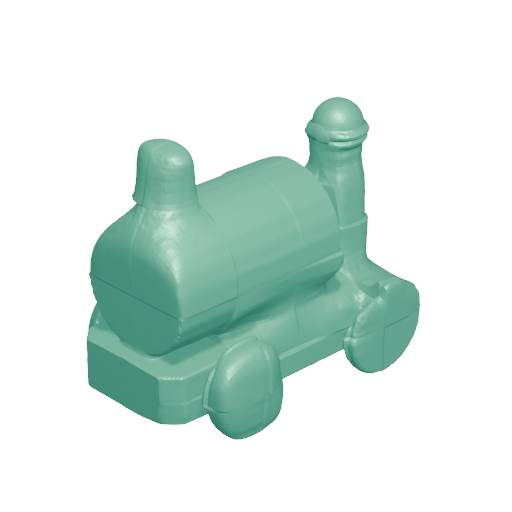}
        & \includegraphics[width=0.11\linewidth]{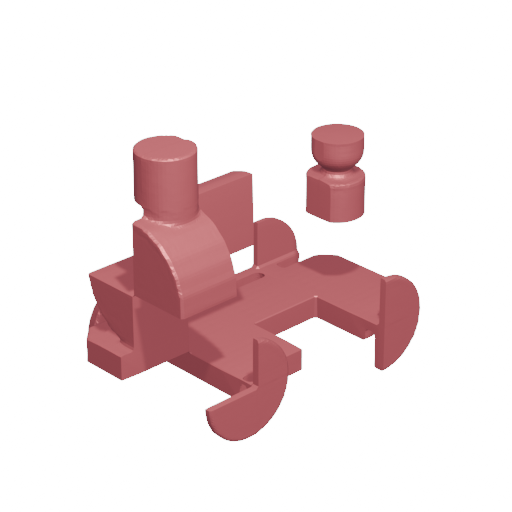}
        & \includegraphics[width=0.11\linewidth]{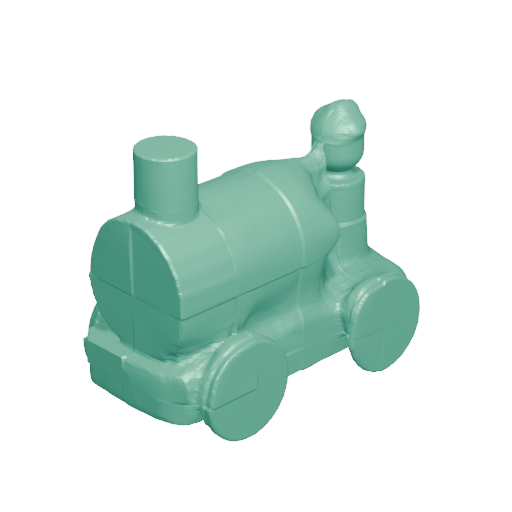}
        & \includegraphics[width=0.11\linewidth]{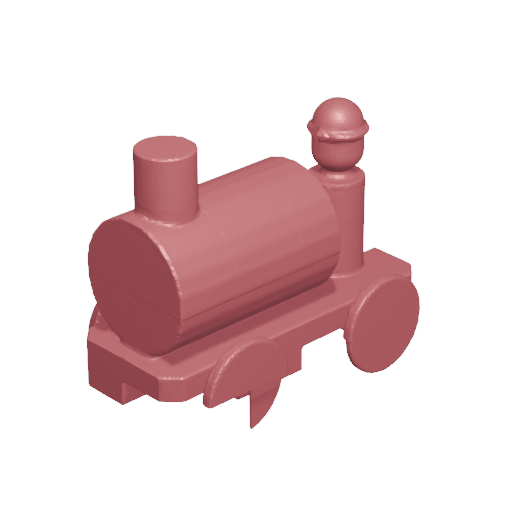}
        & \includegraphics[width=0.11\linewidth]{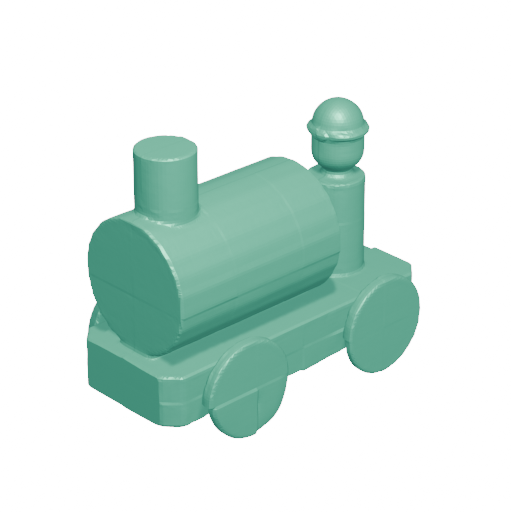}
        & \includegraphics[width=0.11\linewidth]{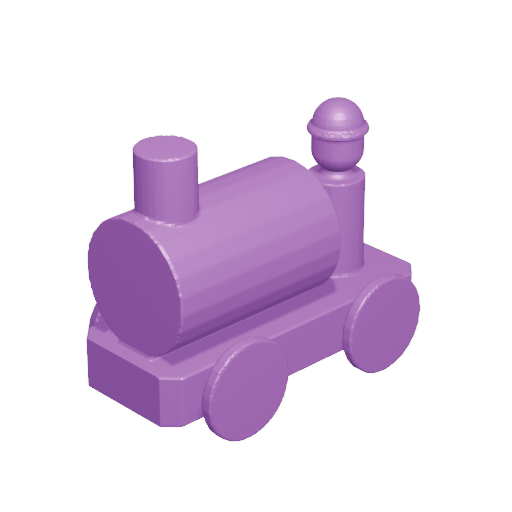}
        \\
        \includegraphics[width=0.11\linewidth]{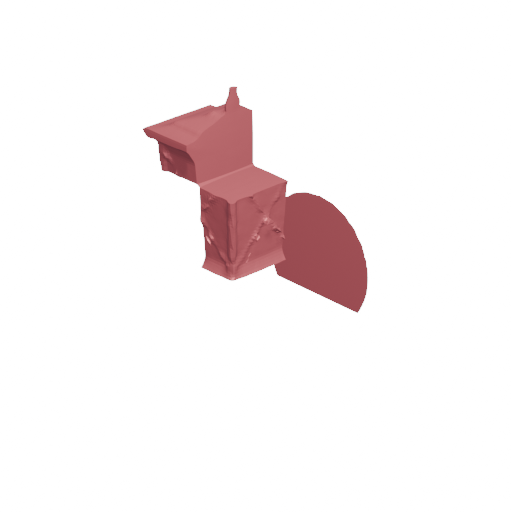}
        & \includegraphics[width=0.11\linewidth]{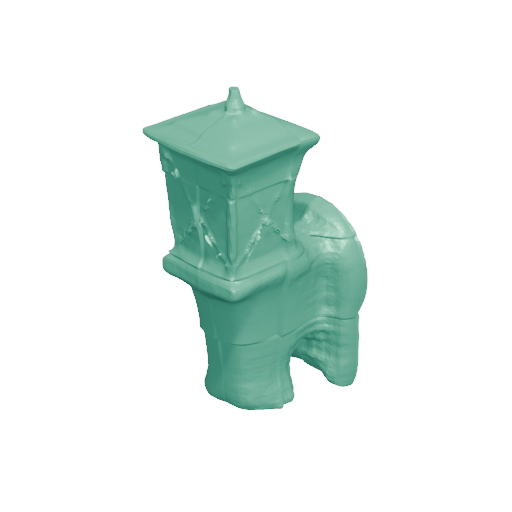}
        & \includegraphics[width=0.11\linewidth]{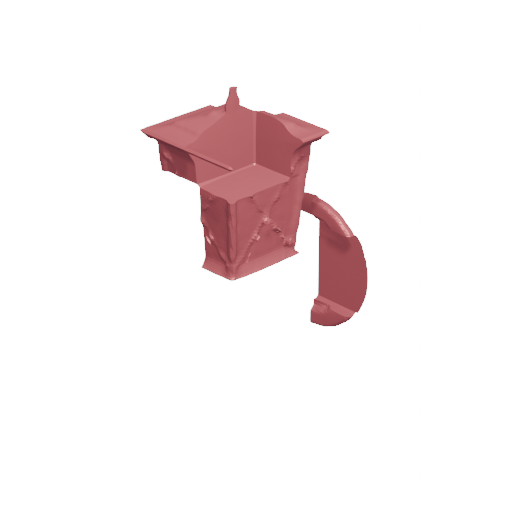}
        & \includegraphics[width=0.11\linewidth]{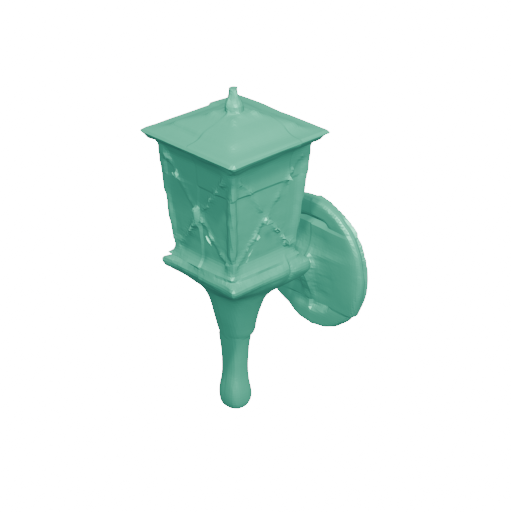}
        & \includegraphics[width=0.11\linewidth]{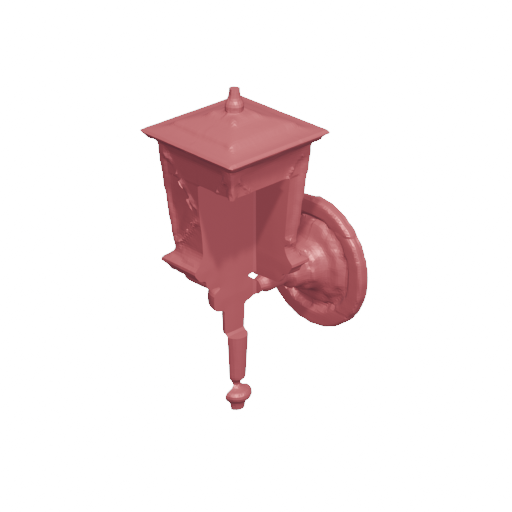}
        & \includegraphics[width=0.11\linewidth]{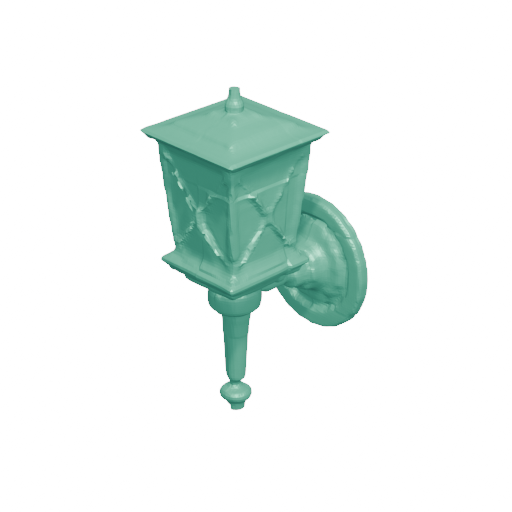}
        & \includegraphics[width=0.11\linewidth]{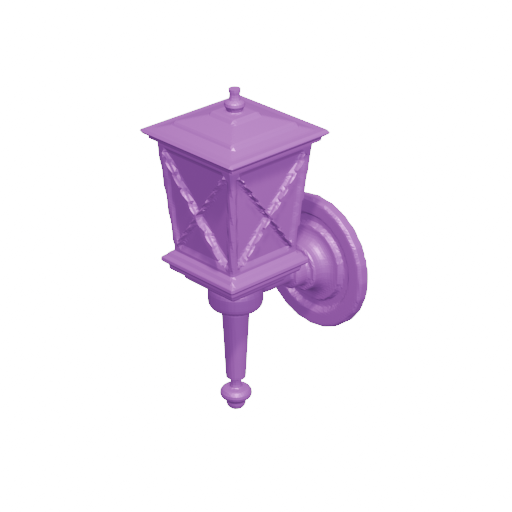}
    \end{tabular}
    \caption{Completion of randomly masked ShapeNet~\citep{ShapeNet} objects using the SLT. This is often a slightly simpler task as small holes are relatively easily filled by propagating information from neighboring patches.}
    \label{fig:eval:random_masking_ratio}
\end{figure}

One of the masking strategies during training is to randomly mask out inputs as outlined in Section~\ref{para:method:masking}. We also numerically evaluated the random masking
completion tasks (R75), (R50) and (R25) in Table~\ref{tab:eval:completion}.
Now, in Figure~\ref{fig:eval:random_masking_ratio} we show visual results on ShapeNet~\citep{ShapeNet}, for the same three completion tasks.
Even when only small parts of the input are given, e.g.\ 25\% in the (R25) task,
the information in neighboring tokens is used to produce near-perfect completions.
This is in contrast to the significantly more challenging (Half) and (Oct) tasks we chose to evaluate our method where completion needs to happen most often for non-neighbor patches.

\subsection{Latent Quality Comparison with DeepSDF and 3DShape2VecSet}%
\label{sec:appendix:latent_quality_comparison}
We compare the auto-encoding quality of our P-VAE (Section~\ref{sec:pvae})
against DeepSDF~\citep{DeepSDF_Park_2019_CVPR} in
Table~\ref{tab:appendix_pvae_vs_deepsdf} and against
3DShape2VecSet~\citep{Zhang2023_3DShape2VecSet} in
Table~\ref{tab:appendix_pvae_vs_3ds2vs}. For
3DShape2VecSet~\citep{Zhang2023_3DShape2VecSet}, we follow their evaluation code
with regards to the size of the object and with regards to the L1 Chamfer
Distance. Like in their code, we omit the
division by 2 which would be part of the original definition from the
supplemental of~\citet{Mescheder2019OccupancyNets}.
In addition to the numbers on our regular P-VAE, we also report the results when
decoding optimized latent codes $z'$ using an AutoDecoder as described in
Section~\ref{sec:appendix:slt_ad} under the label P-VAE-AD.

The results in Table~\ref{tab:appendix_pvae_vs_deepsdf} demonstrate that,
on auto encoding quality,
our P-VAE outperforms DeepSDF on all categories with and without optimized latent codes.
The numbers in Table~\ref{tab:appendix_pvae_vs_3ds2vs} suggest that
the auto-encoding performance of our P-VAE is significantly better in $F_1$ and CD$_\mathrm{L1}$
and on par with the cross-attention based auto encoding strategy from 3DShape2VecSet.

\begin{table}[ht]
    \caption{Comparing mean and median (L2) Chamfer Distance
    in a P-VAE auto-encoding quality comparison with DeepSDF~\citep{DeepSDF_Park_2019_CVPR}.} \label{tab:appendix_pvae_vs_deepsdf}
    \centering
    \begin{tabular}{c@{ }c@{ }c@{ }c@{ }c@{ }c}
        \toprule
        CD$\downarrow$ (mean)  %
                     & chair & plane & table & lamp & sofa 
        \\
        \midrule
     
        P-VAE         & \textbf{0.0759} & 0.0236 & 0.0914 & 0.415 & 0.0269 \\ 
        P-VAE-AD      & 0.1427 & \textbf{0.0177} & \textbf{0.0743} & \textbf{0.4041} & \textbf{0.0263} \\
        DeepSDF       & 0.204 & 0.143 & 0.553 & 0.832 & 0.132 \\
    
        \midrule
        CD$\downarrow$ (median)  %
                      & chair & plane & table & lamp & sofa 
        \\
        \midrule
        P-VAE         & 0.0203 & 0.0075 & 0.023 & 0.0167 & 0.0264 \\ 
        P-VAE-AD      & \textbf{0.0199} & \textbf{0.0067} & \textbf{0.0225} & \textbf{0.0138} & \textbf{0.0259} \\
        DeepSDF       & 0.072  & 0.036 & 0.068 & 0.219 & 0.088 \\
       \bottomrule
    \end{tabular}
\end{table}

\begin{table}[ht]
    \caption{P-VAE auto-encoding quality comparison with 3DShape2VecSet~\citep{Zhang2023_3DShape2VecSet}.} \label{tab:appendix_pvae_vs_3ds2vs}
    \centering
        \begin{tabular}{lcccc}
        \toprule
          Method
        & IoU$\uparrow$
        & $F_1$$\uparrow$
        & CD$_\mathrm{L1}\downarrow$
        & HD$\downarrow$
        \\
        \midrule
        P-VAE
        & 0.9483
        & \textbf{0.9931}
        & \textbf{0.0094}
        & 0.0658
        \\
        3DShape2VecSet
        & \textbf{0.965}
        & 0.970
        & 0.038
        & -
        \\
        \bottomrule
    \end{tabular}
\end{table}

\subsection{Shape Completion on 3D-EPN}
\label{sec:appendix:3d_epn_comparison}
While our model was never intended for and never trained to perform shape
completion from incomplete SDF patches, it might still be interesting to see how
it performs on such a task.
Incomplete SDF patches mean that, within the same patch, parts of the geometry are already missing.

\begin{figure}
    \centering
    \begin{tabular}{l@{}c@{}c@{}c@{}c@{}c@{}c@{}c@{}c}
          \rotatebox{90}{\parbox{0.11\linewidth}{\centering Input}}
        & \includegraphics[width=0.11\linewidth]{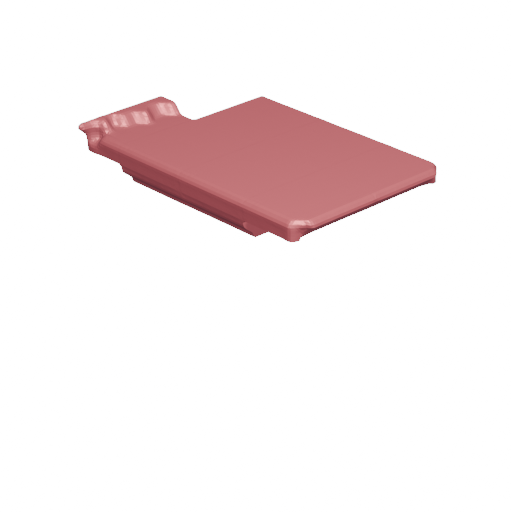}
        & \includegraphics[width=0.11\linewidth]{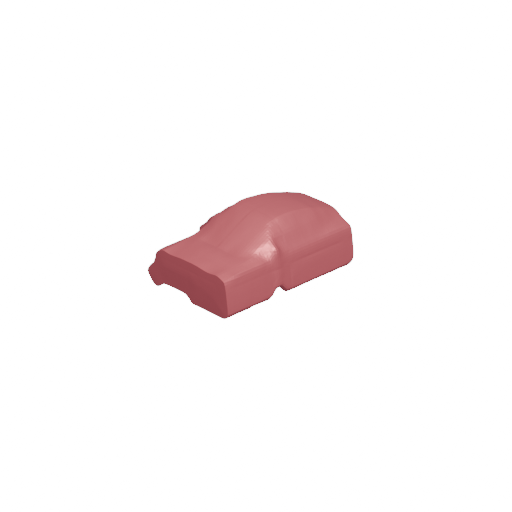}
        & \includegraphics[width=0.11\linewidth]{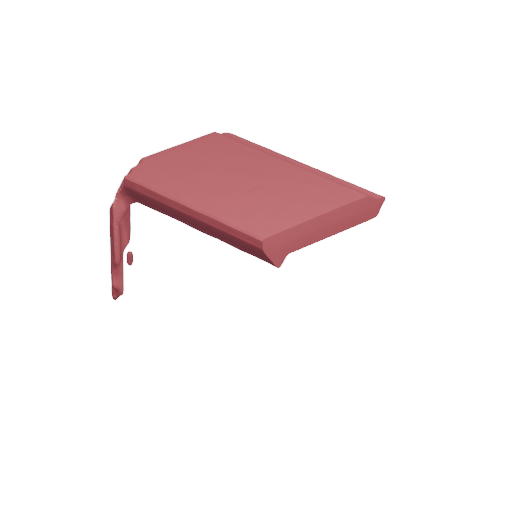}
        & \includegraphics[width=0.11\linewidth]{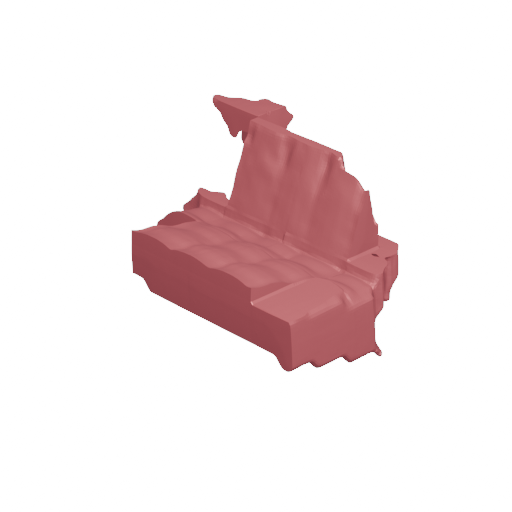}
        & \includegraphics[width=0.11\linewidth]{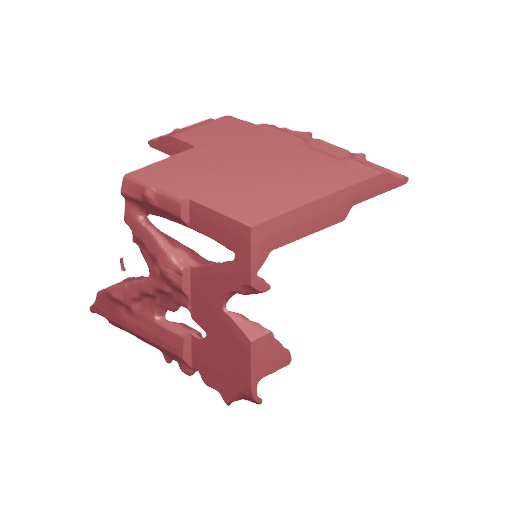}
        & \includegraphics[width=0.11\linewidth]{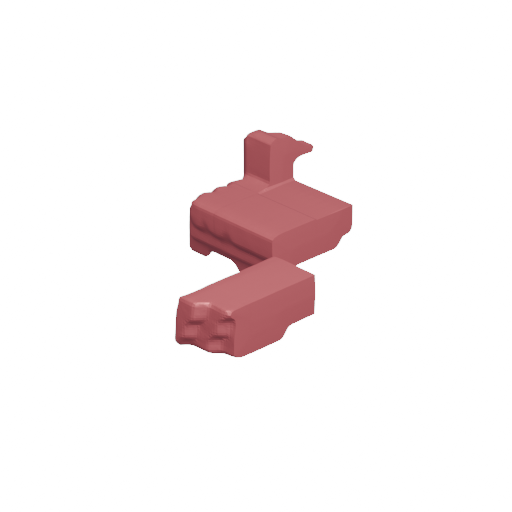}
        & \includegraphics[width=0.11\linewidth]{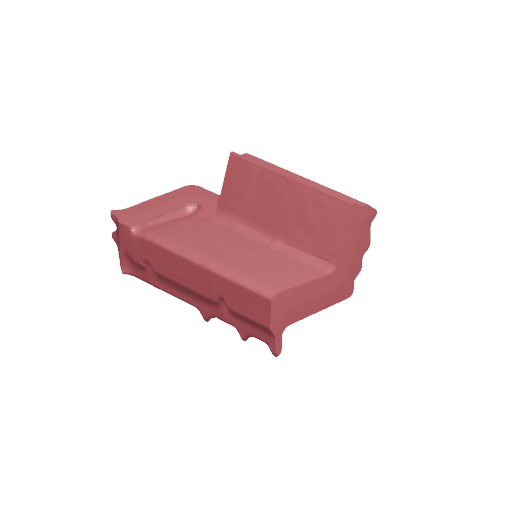}
        & \includegraphics[width=0.11\linewidth]{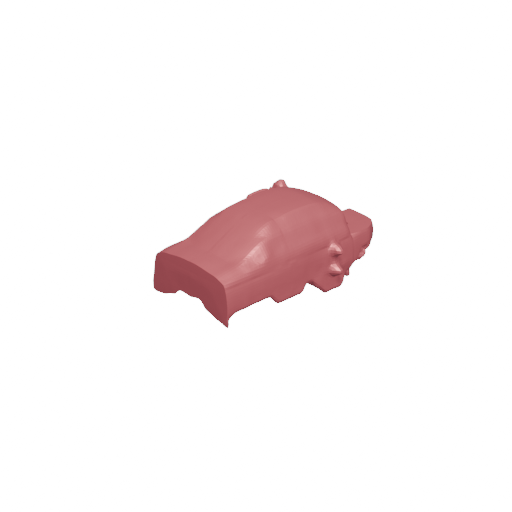}
        \\
          \rotatebox{90}{\parbox{0.11\linewidth}{\centering SLT}}
        & \includegraphics[width=0.11\linewidth]{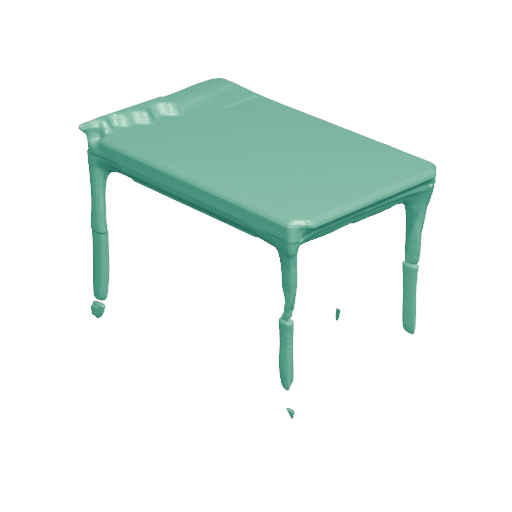}
        & \includegraphics[width=0.11\linewidth]{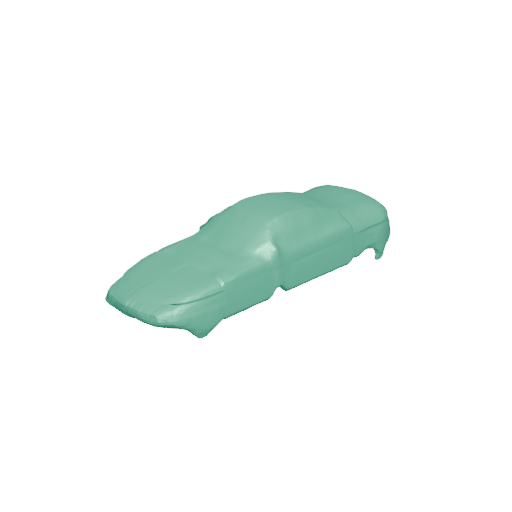}
        & \includegraphics[width=0.11\linewidth]{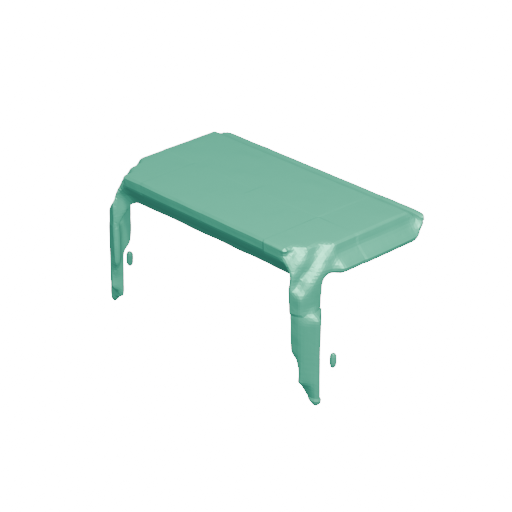}
        
        & \includegraphics[width=0.11\linewidth]{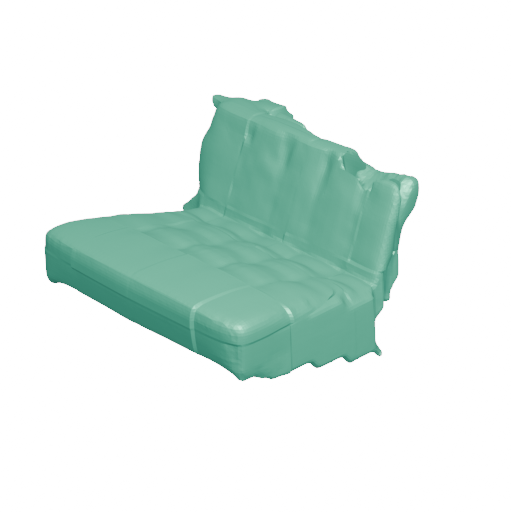}
        & \includegraphics[width=0.11\linewidth]{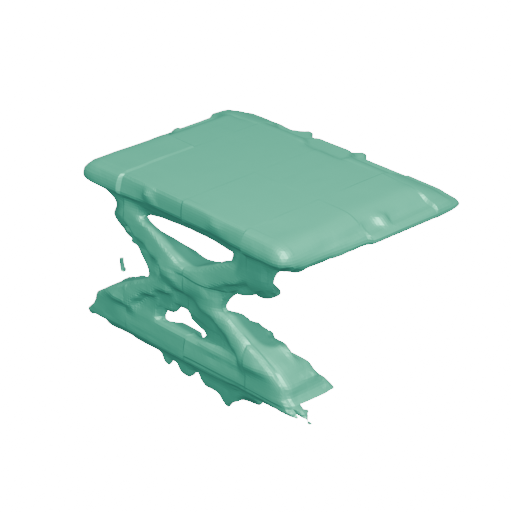}
        & \includegraphics[width=0.11\linewidth]{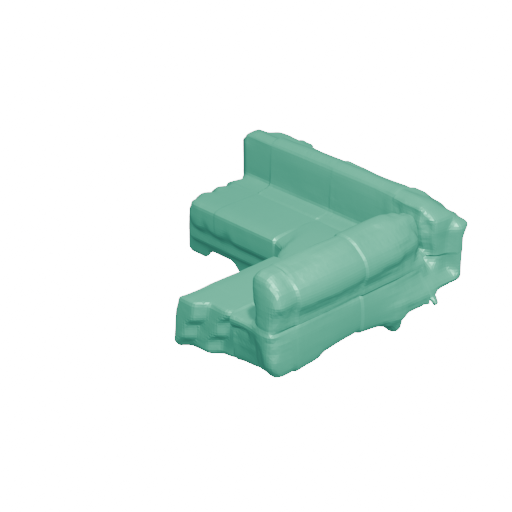}
        & \includegraphics[width=0.11\linewidth]{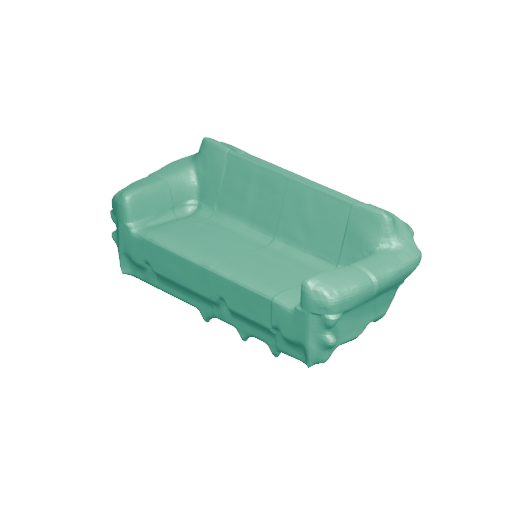}
        & \includegraphics[width=0.11\linewidth]{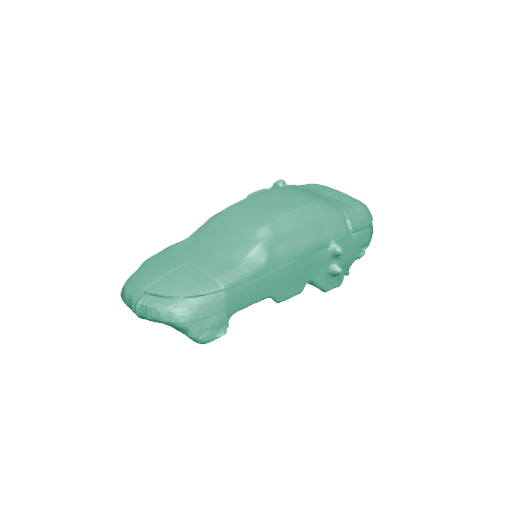}
        \\
    \end{tabular}
    \caption{Completion of partial $32^3$ SDFs from 3D-EPN~\citep{TR1_Dai_2017_CVPR} using our SLT
    which is trained only on complete patches from $128^3$ SDFs from ShapeNet~\citep{ShapeNet}.
    While not trained on this kind of input, the SLT still demonstrates its geometric understanding
    and produces reasonable shape completions.}
    \label{fig:appendix:3depn}
\end{figure}

In Figure~\ref{fig:appendix:3depn}, we show the result of trying to complete the
partial $32^3$ SDFs in the 3D-EPN~\citep{TR1_Dai_2017_CVPR} dataset with our SLT
trained on complete, high-resolution ShapeNet~\citep{ShapeNet} patches.
The partial SDFs from 3D-EPN~\citep{TR1_Dai_2017_CVPR} are generated from a
single camera view and, thus, everything behind the first visible surface is
considered inside, often producing incorrect SDF patches where outside
areas are encoded to be deep inside an object.
We upsample the low resolution SDFs to our $128^3$ resolution in order to get multiple patches. This leads to staircase artifacts which have not been seen in this way by the SLT, even for complete patches.

Our SLT assumes that every given, non-masked $32^3$ input patch contains
complete information.
Therefore, it does not attempt to repair their partially corrupt information
and -- for the most part -- just feeds the given input through to the output.
While some masked patches on the side of the objects can be filled in,
as expected, this generally leads to poor results.

While it could be possible to train or fine-tune the existing SLT architecture on
this task, we think that further research would be required to best adapt our
method to this setting.

\subsection{Comparison with AnchorFormer} \label{sec:appendix:comparison_anchorformer}
Since our method works on SDF patches rather than point clouds,
performing one-to-one comparisons with point cloud methods is somewhat problematic.

In Table~\ref{tab:appendix:comparison_anchorformer}, we show the performance of AnchorFormer~\citep{Chen2023AnchorFormer}
when evaluated on the regular PCN~\citep{PCN_8491026} test split
and when presented with bottom-half inputs from the same objects,
uniformly sampled from our preprocessed (Appendix~\ref{appendix:data_prep}) version of ShapeNet~\citep{ShapeNet}.
We removed some of the objects from the PCN~\citep{PCN_8491026} test split, mostly cars,
for which we did not have a matching mesh available to generate bottom-half inputs.

For evaluating AnchorFormer~\citep{Chen2023AnchorFormer},
we use their only published pre-trained checkpoint which was trained on PCN~\citep{PCN_8491026}.
Completion from just the bottom half can be a much harder task,
since the output on the top half is completely unconstrained.
Typically, PCN~\citep{PCN_8491026} and other partial point cloud settings provide
at least some sparse global information which guides and constrains the completion process.

As a second experiment (also shown in Table~\ref{tab:appendix:comparison_anchorformer}),
we use the same AnchorFormer~\citep{Chen2023AnchorFormer} model and checkpoint
to complete our test set of ShapeNet~\citep{ShapeNet} meshes from bottom halves,
as shown in Figure~\ref{fig:eval:completion_shape_net} and evaluated for our method in Table~\ref{tab:eval:completion}.

The numbers on the PCN~\citep{PCN_8491026} data show that AnchorFormer~\citep{Chen2023AnchorFormer} is not well suited to solve
the (Half) completion task. Both metrics drop significantly.
Furthermore, on our ShapeNet~\citep{ShapeNet} data, the (Half) task produces slightly better numbers than on the PCN~\citep{PCN_8491026} data.
However, our approach achieves better completion quality on the (Half) shape completion task.

\begin{table}[ht]
    \caption{Comparison with AnchorFormer~\citep{Chen2023AnchorFormer} on completing partial inputs from ShapeNet~\citep{ShapeNet}.
    Details on the compared configurations are explained in Section~\ref{sec:appendix:comparison_anchorformer}.
    Some qualitative results can be seen as part of Figure~\ref{fig:eval:completion_shape_net}.}
    \label{tab:appendix:comparison_anchorformer}
    \centering
    \begin{tabular}{cccccc}
        \toprule
         Method & Data Source & Split & Task & $F_1$ $\uparrow$ & CD$_\mathrm{(L2)}$ $\downarrow$ 
         \\
         \midrule
         AnchorFormer & PCN & PCN & PCN & 0.8379 & 0.1986
         \\
         AnchorFormer & ShapeNet & PCN & Half & 0.6953 & 1.4791
         \\
         \midrule
         AnchorFormer & ShapeNet & Ours & Half & 0.7164 & 4.2816
         \\
         Ours (see Table~\ref{tab:eval:completion}) & ShapeNet & Ours & Half & \textbf{0.8468} & \textbf{1.0221}
         \\
         \bottomrule
    \end{tabular}
\end{table}

%% file: sections/appendix_metrics_and_preprocessing.tex
\section{Metrics} \label{appendix:metrics}
In order to evaluate any of the following metrics,
we use Marching Cubes~\citep{marching_cube_article} to generate a mesh from our generated SDF
and then uniformly sample 1M points $X$ and $Y$ from the resulting meshes.

We use the following point-based metrics:
Hausdorff Distance (HD), to measure the largest gap between the original and the reconstructed geometry:
\begin{equation}
    \operatorname{HD}(X, Y)  = \max \left\{\max_{x \in X} \left\{ \min_{y \in Y} d(x, y) \right\}, \max_{y \in Y} \left\{ \min_{x \in X} d(x, y) \right\}\right\}.
\end{equation}
Similarly, the Unidirectional Hausdorff Distance (UHD) measures the maximum distance between objects, but here, in line with previous work~\citep{AutoSDF_Mittal_2022_CVPR},
it is measured from partial input $X$ to completion $Y$ to measure the "fidelity" of the completed output:
\begin{equation}
    \operatorname{UHD}(X, Y) = \max_{x \in X} \left\{ \min_{y \in Y} d(x, y) \right\}.
\end{equation}

We use the ($L2$) Chamfer Distance~\citep{Fan_2017_CVPR}, multiplied by 1000, to measure the accuracy of the reconstruction via the mean squared distance between the original and reconstructed geometry:
\begin{equation}
    \operatorname{CD}(X,Y) = \frac{1}{|X|} \sum_{x \in X} \min_{y \in Y}\left\{ d^2(x,y) \right\} + \frac{1}{|Y|} \sum_{y \in Y} \min_{x \in X}\left\{ d^2(y,x) \right\}.
\end{equation}
In both cases, $d(x,y)$ measures the $L2$ distance between two points $x,y \in \mathbb{R}^3$.

For consistency with prior work, these metrics are evaluated at a normalized scale where the bounding box diameter is 1.
This corresponds to the normalization of the ShapeNet~\citep{ShapeNet} dataset.

Following the evaluation of AutoSDF~\citep{AutoSDF_Mittal_2022_CVPR},
we separately measure objectionable reconstruction artifacts using F-score @1\% ($F_1$), normal consistency (NC) and inaccurate normals (IN).
The F-score @1\% ($F_1$) sets a threshold at 1\% of the side length of the reconstructed volume~\citep{What3D_Tatarchenko_2019_CVPR},
within which a neighboring point on the ground truth mesh needs to be found from a reconstruction sample for computing precision and vice-versa for recall.
From these, the $F_1$ score is computed as usual.

Completeness (CMP) measures the recall within a threshold of $1.5\%$ of the side length, based on the implementation of~\citet{MPC}.

Normal Consistency (NC) measures the mean absolute dot-product between normals on the reconstructed surface and the closest ground-truth surface point:
\begin{equation}
    NC(X,Y) = \frac{1}{|X|}\sum_{x \in X} \left\{ |n_x \cdot n_y| ~:~ y = \operatorname{\arg\min}_{y \in Y} d(x,y) \right\}
\end{equation}
The absolute value is taken to allow for flipped normals in the ground truth data.
For the normals $n_x$ and $n_y$, we use the geometric face normals of the faces that generated the sampled points $x$ and $y$.
Inaccurate Normals (IN) complements the NC measure by counting the percentage of normals which are outside of a 5-degree threshold of the normal of the closest ground-truth point.

We also measure Intersection over Union (IoU) based on the sign of the $128^3$ SDFs.

\section{Data Preparation} \label{appendix:data_prep}
In order to compute SDFs from meshes,
we translate them such that their bounding box is centered at the origin and then uniformly scale them into $[-1,1]^3$.
Then, for ShapeNetCoreV1, we remeshed the mesh in Blender via voxelization using a voxel size of $0.008$.
This was necessary to create manifold meshes from which SDFs could be computed.
Admittedly, this resulted in the loss of some geometry which does not have any volumetric counterpart.
Nevertheless, we consider this remeshed version our ground truth data for SDF-based shape completion training.
For evaluation, we excluded objects which lost all geometry in the conversion.
The meshes in the ABC dataset are of much higher quality.
Here, we only normalized the meshes to $[-1,1]^3$ and then densely sampled signed distances in a regular grid of size $128^3$.

%% file: sections/appendix_architecture_and_training.tex
\section{Architecture Details} \label{appendix:architecture}
We implemented all our models in PyTorch~\citep{pytorch} and trained using PyTorch Lightning~\citep{lightning}.
 
The P-VAE is a variational autoencoder built with 3D convolutions.
The encoder converts an SDF-patch of shape $[B, 1, 32, 32, 32]$ into a mean
$\mu$ and variance $\sigma^2$ vector of shape $[B, 8192]$.
To be precise,
the encoder returns $\log(\sigma^2)$, from which we compute $\sigma^2$. During
training, the latent representation of the SDF-patch is sampled from the normal
distribution $z\sim \mathcal{N}(\mu, \sigma^2)$. During inference we use $z=\mu$.
The architectures of the encoder and the decoder are given in Table~\ref{tab:vae_encoder} and Table~\ref{tab:vae_decoder} respectively.
The ConvBlock specified in Table~\ref{tab:conv_block} and the DecoderLayer in Table~\ref{tab:vae_decoder_layer} are reoccurring architectural structures in both the encoder and the decoder.
Note that each layer specifies its predecessor in the \emph{parent} column. Skip connections are joined by adding the output of two parents together.

\begin{table*}[htb]
    \centering
    \caption{Architecture of a \textbf{ConvBlock} element as used in the P-VAE Encoder in Table~\ref{tab:vae_encoder}}
    \begin{tabular}{r|r|c|l}
        \toprule
        Layer Name & Parent Name & Layer Parameters & Output Shape \\
        \hline
        Conv3D& - & kernel 3, stride 1, pad 1& $[B, c_{out}, d,d,d]$\\
        BatchNorm3d & Conv3D & - & $[B, c_{out}, d,d,d]$\\
        ReLU & BatchNorm3d & - & $[B, c_{out}, d,d,d]$\\
        \bottomrule
    \end{tabular}\\
    \label{tab:conv_block}
\end{table*}

\begin{table*}[htb]
    \centering
    \caption{Architecture of a \textbf{DecoderLayer} as used in Table~\ref{tab:vae_decoder}.
    If not otherwise noted, all layers use a kernel size of 3 ($[3,3,3]$), stride 1 and padding 1 ($[1,1,1]$).
    The input has shape $[B, c_{in}, d, d, d]$.}
    \begin{tabular}{r|r|c|l}
        \toprule
        Layer Name & Parent Name & Layer Parameters & Output Shape \\
        \hline
        Conv3D & layer input & - & $[B,c_{out},d,d,d]$\\
        Conv3DTrans & layer input & stride 2, output padding 2 & $[B, c_{out}, 2d, 2d, 2d]$\\
        BatchNorm3D & Conv3D & - & $[B,c_{out},d,d,d]$\\
        Upsample & BatchNorm3D & factor 2, trilinear & $[B, c_{out}, 2d, 2d, 2d]$\\
        Add & Conv3DTrans, Upsample& -  & $[B, c_{out}, 2d, 2d, 2d]$\\
        ReLU & Add & -  & $[B, c_{out}, 2d, 2d, 2d]$\\
        \bottomrule
    \end{tabular}\\
    \label{tab:vae_decoder_layer}
\end{table*}

\begin{table*}[htb]
    \centering
    \caption{Architecture of the \textbf{P-VAE Encoder}.
    If not otherwise noted, all layers use a kernel size of 3 ($[3,3,3]$) stride 1 and padding 1 ($[1,1,1]$).
    The input patch has shape $[B, 1, 32, 32, 32]$.}
    \begin{tabular}{r|r|c|l}
        \toprule
        Layer Name & Parent Name & Layer Parameters & Output Shape \\
        \hline
        ConvBlock-1& SDF patch & - & $[B, 32, 32, 32 32]$\\
        ConvBlovk-2 & ConvBlock-1 & - & $[B, 32, 32,32,32]$\\
        ConvBlovk-3 & ConvBlock-2 & - & $[B, 32, 32,32,32]$\\
        ConvBlovk-4 & ConvBlock-3 & - & $[B, 32, 32,32,32]$\\
        Add-1 & ConvBlock-1, SDF patch & broadcasting & $[B, 32, 32,32,32]$\\
        MaxPool-1 & Add-1 & stride 2 &  $[B, 32, 16,16,16]$\\
        Conv3D-1 & Add-1 & stride 2 & $[B, 32, 16,16,16]$\\
        Add-2 & MaxPool-1, Conv3D-1 & - & $[B, 32, 16,16,16]$\\
        Conv3D-2 & Add-2 & kernel 1, padding 0 & $[B, 32, 16,16,16]$ \\
        ConvBlock-5 & Add-2& - & $[B, 64, 16,16,16]$\\
        ConvBlock-6 & ConvBlock-5 & - & $[B, 64, 16,16,16]$\\
        ConvBlock-7 & ConvBlock-6 & - & $[B, 64, 16,16,16]$\\
        ConvBlock-8 & ConvBlock-7 & - & $[B, 64, 16,16,16]$\\
        Add-3 & Conv3D-1, ConvBlock-7& - & $[B, 64, 16,16,16]$\\
        MaxPool-2 & Add-3 & stride 2 &  $[B, 64, 8,8,8]$\\
        Conv3D-3 & Add-3 & stride 2 &  $[B, 64, 8,8,8]$\\
        Add-4 & MaxPool-2, Conv3D-3 & - & $[B, 64, 8,8,8]$\\
        Conv3D-4 & Add-4 & kernel 1, padding 0 & $[B, 128, 8,8,8]$ \\
        ConvBlock-9 & Add-4& - & $[B, 128, 8,8,8]$\\
        ConvBlock-10 & ConvBlock-9 & - & $[B, 128, 8,8,8]$\\
        ConvBlock-11 & ConvBlock-10 & - & $[B, 128, 8,8,8]$\\
        ConvBlock-12 & ConvBlock-11 & - & $[B, 128, 8,8,8]$\\
        Add-5 & Conv3D-4, ConvBlock-12 & - & $[B, 128, 8,8,8]$\\
        MaxPool-3 & Add-5 & stride 2 &  $[B, 128, 4,4,4]$\\
        Conv3D-5 & Add-5 & stride 2 &  $[B, 128, 4,4,4]$\\
        Add-6 & MaxPool-3, Conv3D-5 & - & $[B, 128, 4,4,4]$\\
        Conv3D-6 & Add-4 & kernel 1, padding 0 & $[B, 256, 4,4,4]$ \\
        ConvBlock-13 & Add-6 & - & $[B, 256, 4,4,4]$\\
        ConvBlock-14 & ConvBlock-13 & - & $[B, 256, 4,4,4]$\\
        ConvBlock-15 & ConvBlock-14 & - & $[B, 256, 4,4,4]$\\
        ConvBlock-16 & ConvBlock-15 & - & $[B, 256, 4,4,4]$\\
        Add-7 & Conv3D-4, ConvBlock-16 & - & $[B, 128, 8,8,8]$\\
        MaxPool-4 & Add-7 & stride 2 &  $[B, 128, 2,2,2]$\\
        Conv3D-7 & Add-7 & stride 2 &  $[B, 128, 2,2,2]$\\
        Add-8 & MaxPool-4, Conv3D-7 & - & $[B, 128, 2,2,2]$\\
        Conv3D-8 & Add-8 & kernel 1, padding 0 & $[B, 512, 2,2,2]$ \\
        ConvBlock-17 & Add-8 & - & $[B, 512, 2,2,2]$\\
        ConvBlock-18 & ConvBlock-17 & - & $[B, 512, 2,2,2]$\\
        ConvBlock-19 & ConvBlock-18 & - & $[B, 512, 2,2,2]$\\
        ConvBlock-20 & ConvBlock-19 & - & $[B, 512, 2,2,2]$\\
        Add-9 & Conv3D-8, ConvBlock-20 & - & $[B, 512, 2,2,2]$\\
        Conv3D-9 & Add-9 & kernel 1, padding 0 & $[B, 1024, 2, 2, 2]$\\
        \bottomrule
    \end{tabular}\\
    \label{tab:vae_encoder}
\end{table*}

\begin{table*}[htb]
    \centering
    \caption{Architecture of the \textbf{P-VAE Decoder}. The decoder converts a input latent code with shape 
    $[B, 8192] \rightarrow [B, 512, 2, 2, 2]$ into an SDF-patch of shape $[B,1, 32, 32, 32]$.}
    \begin{tabular}{r|r|c|l}
        \toprule
        Layer Name & Parent Name & Layer Parameters & Output Shape \\
        \hline
        DecoderLayer-1 & latent code & $c_{in}=512$, $c_{out}=512$ & $[B, 512, 4, 4, 4]$\\
        DecoderLayer-2 & DecoderLayer-1 & $c_{in}=512$, $c_{out}=256$ & $[B, 256, 8, 8, 8]$\\
        DecoderLayer-3 & DecoderLayer-2 & $c_{in}=256$, $c_{out}=256$ & $[B, 256, 16, 16, 16]$\\
        DecoderLayer-4 & DecoderLayer-3 & $c_{in}=256$, $c_{out}=128$ & $[B, 128, 32, 32, 32]$\\
        Conv3D-1 & DecoderLayer-4 & kernel 1, stride 1, padding 0 & $[B, 1, 32, 32, 32]$\\
        \bottomrule 
    \end{tabular}\\
    \label{tab:vae_decoder}
\end{table*}

\section{Training Details} \label{appendix:training_details}

\subsection{Patch-Variational Autoencoder}
We trained the P-VAE for 250 epochs with early stopping at epoch 193 with initial learning rate of 1e-4 and CosineAnnealing scheduler. 
The batch size for the training was 128 per GPU.

\subsection{AutoDecoder}
The refined codes $z'$ were optimized with Adam for 200 steps per patch. The
gaussian noise that was added on the initial $z$ from the P-VAE was sampled from a gaussian distribution
$\mathcal{N}(0, 1e^{-2})$.

\subsection{SDF Latent Transformer}
The masking ratio used for training is $0.4$.
Our learning rate was $1e^{-5}$ and we use a cosine scheduler with 1400 warm-up steps for the transformer and trained for 120k steps. The batch size was 64 per GPU.
All SLT configurations used the same training configuration and setup.